\documentclass{article} 
\usepackage{iclr2027_conference,times}

\usepackage{amsmath,amsfonts,bm}

\def\eqref#1{equation~\ref{#1}}

\def\1{\bm{1}}

\DeclareMathAlphabet{\mathsfit}{\encodingdefault}{\sfdefault}{m}{sl}
\SetMathAlphabet{\mathsfit}{bold}{\encodingdefault}{\sfdefault}{bx}{n}

\usepackage{hyperref}
\usepackage{url}

\usepackage{pifont}
\usepackage{booktabs}       
\usepackage{amsfonts}       
\usepackage{microtype}      
\usepackage{xcolor}         
\usepackage{graphicx}
\usepackage{subcaption}
\usepackage{amsmath}
\usepackage{array}
\usepackage{amsthm}
\usepackage{thmtools}
\usepackage{amssymb}
\usepackage{adjustbox}
\usepackage{wrapfig}
\usepackage[normalem]{ulem}
\usepackage{mathrsfs}
\usepackage{multirow}
\usepackage{svg}
\usepackage{thm-restate}
\usepackage{algorithm}
\usepackage{algpseudocode}
\usepackage{tabularx}
\usepackage{longtable}
\usepackage{colortbl}
\usepackage{makecell}
\usepackage{titletoc}

\usepackage[normalem]{ulem}

\title{Fast World Planner: Advantage-Aware \\Action-Chunk Policy for Online Planning}

\author{Jiangtao Kong, Yuchen Wang \& Huajie Shao  \\
William \& Mary \\
\texttt{\{jkong01,ywang142,hshao\}@wm.edu} \\
\And
Yandong Luo \& Lu Gan \\
Georgia Institute of Technology  \\
\texttt{\{yluo471,lgan\}@gatech.edu} \\
\AND
Rongchao Xu \& Guang Wang \\
Florida State University  \\
\texttt{\{rx21a,guang.wang\}@fsu.edu} \\
\And
Yanbing Mao \\
Wayne State University  \\
\texttt{hm9062@wayne.edu} \\
\And
Tianyi Zhou\\
MBZUAI  \\
\texttt{Tianyi.Zhou@mbzuai.ac.ae} \\
}

\iclrfinalcopy 
\begin{document}
\setlength{\abovedisplayskip}{3pt} 
\setlength{\belowdisplayskip}{2pt} 

\maketitle
\lhead{}
\begin{abstract}
Model Predictive Path Integral (MPPI) control is central to online planning for model-based reinforcement learning (MBRL), but its iterative refinement incurs planning latency. Reducing refinement iterations often degrades performance, motivating stronger policy guidance learned from the planner. However, existing methods typically learn from the planner through action-level alignment, which does not directly capture the complete $H$-step planner proposal. Moreover, planner imitation leaves further target improvement to online refinement. To address these limitations, we propose Fa\textbf{S}t \textbf{W}orld Pl\textbf{AN}ner (\textbf{SWAN}), an advantage-aware MPPI-based MBRL framework that explicitly represents and improves the complete \(H\)-step planner proposal to improve policy guidance for efficient online planning. For proposal representation, SWAN combines k-MPPI, a multimodal MPPI extension, with a mixture of factor analyzers (MFA) action-chunk policy to capture multiple modes and correlations across time and action dimensions. For proposal optimization, we introduce Advantage-Guided Proposal Optimization (APO), which reuses MPPI evaluations to construct an improved target using advantage estimates and trains the policy toward this target while preserving planner alignment. Experiments across 31 high-dimensional continuous-control tasks from DMControl, HumanoidBench, and MyoSuite show that SWAN substantially improves control performance and sample efficiency while using 50\% fewer MPPI refinement iterations than state-of-the-art planning-based MBRL methods. Finally, SWAN demonstrates zero-shot sim-to-real transfer on a physical Unitree Go2, achieving stable velocity-tracking locomotion with high-frequency control.
(see our demo at~\url{https://swanfastworldplanner.github.io/}).

\end{abstract} 
\section{Introduction}\label{sec:intro}
\vspace{-0.1in}
Recent model-based reinforcement learning (MBRL) methods such as TD-MPC2~\citep{hansen2024td} and BMPC~\citep{wang2025bootstrapped}
combine learned world models and policies with Model Predictive Path
Integral (MPPI) control, a sampling-based model predictive control (MPC) algorithm, to optimize action sequences online 
~\citep{williams2017model,Hansen2022tdmpc,hansen2024td,
wang2025bootstrapped,zhan2026bootstrap}.
In these MPPI-based methods, the policy guides candidate generation and MPPI improves action sequences through iterative refinement.
However, repeated trajectory evaluation introduces planning latency, 
while reducing refinement iterations leaves less opportunity 
to correct poor candidates
~\citep{williams2017model,pinneri2021sample}. 
Prior work~\citep{anthony2017thinking,silver2017mastering} on expert iteration shows that policies learned
from a planner can improve subsequent search.
Thus, we seek better policy guidance through 
learning from the planner to improve control performance 
with fewer MPPI refinement iterations
~\citep{sacks2023learning}.

\begin{wrapfigure}[18]{R}{0.5\columnwidth}
  \centering
  \captionsetup{font=small,skip=4pt}
  \captionsetup[subfigure]{font=footnotesize,skip=2pt}
  \begin{subfigure}[t]{\linewidth}
    \centering
    \includegraphics[width=\linewidth]{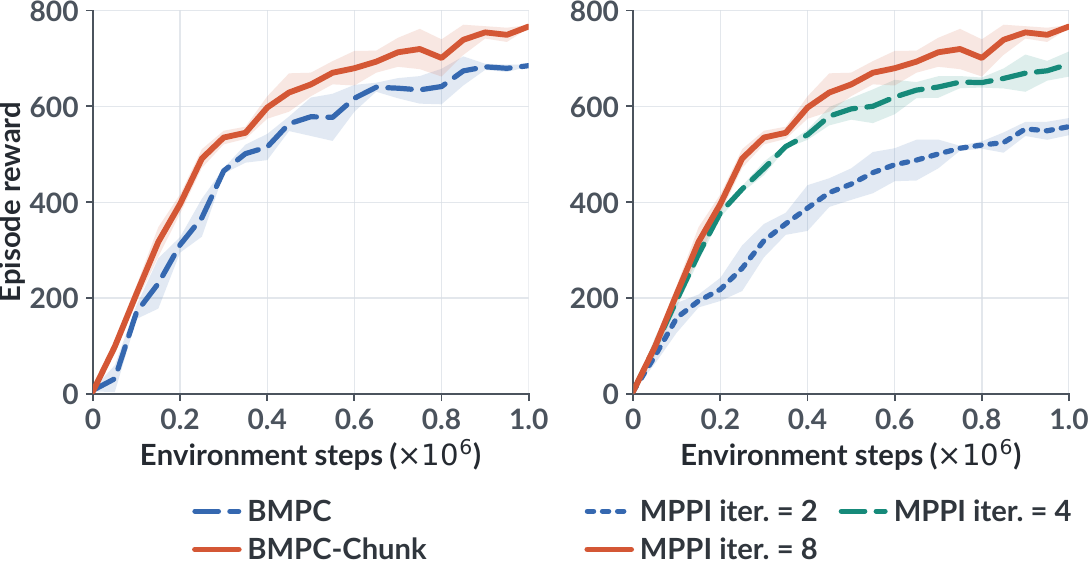}
  \end{subfigure}
  \caption{Motivation on \textit{Dog-Run} task. 
  BMPC-Chunk uses our action-chunk proposal representation
  with planner imitation.
  \textbf{\textit{Left:}} Comparison between BMPC and BMPC-Chunk under 8 MPPI refinement iterations. 
  \textbf{\textit{Right:}} Comparison of BMPC-Chunk across 2, 4, and 8 MPPI refinement iterations. 
  Candidate trajectory counts per iteration are fixed across settings.
  Curves show mean and 95\% CIs over 3 seeds.}
  \label{fig:motivation}
\end{wrapfigure}

A central obstacle to improving policy guidance is
policy-planner mismatch between learned policies and
planner-derived action distributions.
Specifically, in prior critic-guided methods, the critic
learns from planner-generated data but bootstraps with
actions from the learned policy, while the resulting
value estimates guide policy updates and provide terminal
values for MPPI.
Thus, policy-planner mismatch
can lead to inaccurate value estimates,
affecting both policy improvement and planning
~\citep{wang2025bootstrapped,serra-gomez2026a}.
To reduce the mismatch, recent MPPI-based methods
either learn the policy through planner imitation~\citep{wang2025bootstrapped}
or retain critic-guided updates with planner-derived
regularization
~\citep{zhan2026bootstrap,serra-gomez2026a}.
However, two gaps remain in representing and improving the
complete planner proposal. First, the \textit{proposal representation gap}. Existing alignment objectives operate on individual actions or the
planner's first-action distribution, rather than the complete
\(H\)-step action-chunk distribution.
Consequently, the learned policy does not directly learn the temporal
correlations and multiple modes represented by the weighted action
chunks.
To examine the gap, we extend BMPC with our action-chunk
proposal representation while retaining planner imitation,
yielding BMPC-Chunk.
As shown in Fig.~\ref{fig:motivation} (left), BMPC-Chunk
outperforms BMPC under the same MPPI refinement budget on \textit{Dog-Run},
\textbf{motivating} direct representation of the complete planner proposal.
Second, the \textit{proposal optimization gap}. 
Even with an action-chunk representation, direct critic-guided policy updates remain susceptible to policy-planner mismatch.
To reduce the mismatch, planner-derived regularization encourages
planner alignment in critic-guided updates, but the trade-off
with value maximization can leave residual mismatch.
In contrast, planner imitation directly uses the current planner
proposal as the learning target, leaving further target
improvement to MPPI refinement.
As shown in Fig.~\ref{fig:motivation} (right), 
BMPC-Chunk
remains sensitive to the number of refinement iterations,
\textbf{motivating} additional proposal improvement in
policy learning while preserving planner alignment.


Overcoming both gaps poses two technical \textbf{challenges}: 
i) efficiently represent the multimodal action-chunk planner proposal, and
ii) optimize the proposal using advantage estimates 
while preserving planner alignment during policy learning. 
To address both challenges, we introduce 
Fa\textbf{S}t \textbf{W}orld Pl\textbf{AN}ner (SWAN), 
an advantage-aware MPPI-based MBRL framework that 
improves policy guidance through
explicit representation and improvement of the complete
\(H\)-step planner proposal. For proposal representation, SWAN proposes k-MPPI, a multimodal extension of MPPI, with a \textbf{M}ixture of \textbf{F}actor \textbf{A}nalyzers (MFA) action-chunk policy.
k-MPPI maintains multiple search modes through standard MPPI refinement within each mode, while the MFA policy models
correlations across time and action dimensions and supports direct action-chunk sampling.
For proposal optimization, SWAN introduces \textbf{A}dvantage-Guided \textbf{P}roposal
\textbf{O}ptimization (APO). APO uses advantage estimates from MPPI trajectory evaluations to construct an improved target under a KL constraint relative to the current planner proposal. We then train the policy toward the improved target under
a separate constraint that preserves planner alignment. Advantage information therefore enters policy learning
through target construction rather than direct critic gradients through policy actions.
Experiments on 31 high-dimensional continuous-control tasks
across DMControl~\citep{tunyasuvunakool2020},
HumanoidBench~\citep{sferrazza2024humanoidbench}, and
MyoSuite~\citep{caggiano2022myosuite} show that SWAN improves
control performance and sample efficiency with \(50\%\)
fewer MPPI refinement iterations than the MPPI-based baselines.
We further confirm the gains in Unitree G1 and Go2
simulations~\citep{zakka2026mjlab}, then demonstrate
zero-shot sim-to-real transfer to a physical Go2 with
stable velocity tracking at an average \(308.16\,\mathrm{Hz}\).

Our contributions are as follows. 
\textbf{1).}
We propose SWAN, an advantage-aware MPPI-based MBRL
framework that improves online planning efficiency through
better policy guidance.
\textbf{2).} We propose a novel approach that
jointly introduces k-MPPI and an MFA policy to explicitly represent
multiple proposal modes and correlations
across time and action dimensions.
\textbf{3).} We introduce APO to improve the planner proposal using
advantage estimates while preserving planner alignment. Theoretically,
we prove nonnegative expected \(H\)-step advantage of the
exact APO target over the current planner proposal and 
derive a projection-aware sufficient condition for
improvement over ideal planner imitation.
\textbf{4).} 
We validate SWAN's control performance, sample efficiency,
and online planning efficiency through extensive benchmark
comparisons, controlled ablations, and robot locomotion
simulations, and demonstrate zero-shot sim-to-real transfer
to a physical Go2.
\section{Related Work}\label{sec:relatedwork}
\vspace{-0.1in}
\noindent \textbf{Model-based Reinforcement Learning (MBRL).} 
Recent MBRL learns world models to predict the consequences of
actions for policy learning or planning~\citep{m2023model}.
Based on how the learned world model is used, existing methods can be grouped into two categories:
1) imagination-based methods optimize policies from model-generated trajectories~\citep{hafner2019dream,hafner2020mastering,hafner2023mastering,janner2019trust}, 2) planning-based methods optimize action sequences online~\citep{hafner2019learning,sikchi2022learning,Hansen2022tdmpc,hansen2024td}.
LOOP~\citep{sikchi2022learning} learns dynamics in the environment state space and uses an actor-derived prior to constrain trajectory optimization and reduce actor divergence. 
For planning in learned latent world models,
TD-MPC and TD-MPC2 use MPPI as an online planner with learned policies guiding candidate generation~\citep{Hansen2022tdmpc,hansen2024td}. To improve policy learning, subsequent MPPI-based methods reduce the mismatch between learned policies and planner-derived action distributions. BMPC~\citep{wang2025bootstrapped} learns a policy through imitation of the planner's first-action distribution. BOOM~\citep{zhan2026bootstrap} combines critic-guided policy
updates with soft value-weighted alignment to planner actions
stored in replay. PO-MPC~\citep{serra-gomez2026a} learns an adaptive prior from the planner's first-action distribution to regularize policy optimization. 
However, these methods incorporate planner information through one-step action policy learning.
In contrast, SWAN \textit{directly} represents the complete \(H\)-step
planner proposal and uses advantage estimates from MPPI trajectory
evaluations to construct an improved target while
preserving planner alignment.

\noindent \textbf{Multimodal Action-Chunk Policies.}
Recent robot policies use diffusion and flow matching to \textit{implicitly} model high-dimensional, multimodal action-chunk distributions~\citep{chi2025diffusion,chisari2024learning,jiang2025streaming}. D-MPC~\citep{zhou2025diffusion} also learns diffusion action proposals from offline trajectories for online MPC. Furthermore, for MPPI-based MBRL, nonlinear dynamics and non-convex control objectives can produce separated high-value regions in action-chunk space~\citep{lambert2021stein,honda2024stein}. Return-weighted MPPI candidates can therefore contain multimodal structure, motivating diffusion or flow matching for learning the planner proposal. However, online proposal learning requires fast sampling and adaptation to changing planner proposals. Iterative generation introduces control latency~\citep{wang2025onestep}, while direct policy-gradient backpropagation through diffusion sampling can be \textit{computationally expensive and unstable}~\citep{ma2025efficient}. We thus combine k-MPPI with an MFA action-chunk policy to \textit{explicitly} model the planner proposal, supporting direct sampling and capturing correlations across time and action dimensions.

\noindent \textbf{Advantage-Guided Policy Learning.}
Prior model-free RL methods optimize policies using estimated advantages~\citep{schulman2015trust,schulman2016gae,schulman2017proximal,abdolmaleki2018maximum}. Subsequent work incorporates estimated advantages into policy learning by weighting actions from historical, replay, or offline data~\citep{wang2018exponentially,peng2019advantage,kostrikov2022implicit}. 
Recent SPO~\citep{macfarlane2024spo} uses advantage-weighted SMC search to construct first-action targets with exact environment models for policy fitting.
In this paper, we propose APO, which applies advantage estimates from MPPI trajectory evaluations
to improve the current complete planner proposal,
and then trains the policy toward the improved target
under a separate planner-alignment constraint.
\vspace{-0.05in}
\section{Preliminaries}\label{sec:prelim}
\vspace{-0.1in}
\noindent\textbf{Problem Formulation.} 
Reinforcement learning (RL) problems can be considered as
an infinite-horizon discounted Markov decision
process (MDP)
\(\mathcal{M}=(\mathcal{S},\mathcal{A},\mathcal{P},\mathcal{R},\gamma)\),
where \(\mathcal{S}\subseteq\mathbb{R}^{d_s}\) and
\(\mathcal{A}\subseteq\mathbb{R}^{d_a}\) denote the state and continuous
action spaces, \(\mathcal{P}\) is the transition kernel,
\(\mathcal{R}\) is the reward function, and
\(\gamma\in[0,1)\) is the discount factor.
At each discrete time step \(t\), the agent observes
\(s_t\), samples action \(a_t\) from
policy \(\pi(\cdot\mid s_t)\), receives
\(r_t=\mathcal{R}(s_t,a_t)\), and transitions to
\(s_{t+1}\) according to \(\mathcal{P}\).
The objective is to learn a policy \(\pi\) that maximizes
\(J(\pi)=\mathbb{E}_{\pi,\mathcal{P}}
[\sum_{t=0}^{\infty}\gamma^t r_t]\). 
\textbf{Table~\ref{tab:notation} in Appendix~\ref{app:notation} summarizes the notations used in this paper.}

\noindent\textbf{MPPI-Based MBRL.}
MPPI-based MBRL combines learned world models and policies
with MPPI to optimize action sequences online
~\citep{wang2025bootstrapped}.
Following TD-MPC2 and BMPC~\citep{hansen2024td,wang2025bootstrapped},
SWAN uses a latent world model comprising an encoder \(E_\psi\),
a dynamics model \(F_\psi\), a reward model \(R_\psi\), and
a value model \(V_\phi\) for terminal bootstrapping.
The encoder gives \(z_t=E_\psi(s_t)\in\mathbb R^{d_z}\),
from which MPPI evaluates candidate action chunks.


At control step \(t\), MPPI performs \(M\) refinement iterations over
\(H\)-step action chunks
\(\tau=a_{t:t+H-1}\in\mathbb{R}^{D}\), where \(D=Hd_a\).
In prior methods, iteration \(m=1,\ldots,M\) forms a batch
\(\mathcal T^{(m)}\) of candidate chunks by combining samples
from a diagonal Gaussian with chunks generated by rolling out
the one-step action policy \(\pi_{\theta_a}(\cdot\mid z_t)\)
in the latent world model.
For each candidate, MPPI rolls out
\(\widehat z_{h+1}=F_\psi(\widehat z_h,a_{t+h})\) from
\(\widehat z_0=z_t\) and combines predicted rewards with
the terminal value to estimate the \(H\)-step return
\begin{equation}
\widehat G_H(z_t,\tau)
=
\sum_{h=0}^{H-1}
\gamma^h R_\psi(\widehat z_h,a_{t+h})
+
\gamma^H V_\phi(\widehat z_H).
\label{eq:h_step_return}
\end{equation}
Based on the estimated returns, MPPI assigns each candidate a normalized
weight
\(w^{(m)}(\tau)
=
\operatorname{softmax}_{\mathcal T^{(m)}}
\bigl(
\widehat G_H(z_t,\tau)/\eta_{\mathrm{MPPI}}
\bigr)\),
where \(\eta_{\mathrm{MPPI}}>0\) is the temperature and normalization
is over \(\mathcal T^{(m)}\).
To continue refinement, MPPI projects the weighted empirical
distribution onto a diagonal Gaussian by matching the weighted
mean and elementwise variance
\begin{equation}
\mu^{(m+1)}
=
\sum_{\tau\in\mathcal T^{(m)}}w^{(m)}(\tau)\tau,
\qquad
(\sigma^{(m+1)})^2
=
\sum_{\tau\in\mathcal T^{(m)}}w^{(m)}(\tau)
\left(\tau-\mu^{(m+1)}\right)^2,
\label{eq:mppi_update}
\end{equation}
where \(\mu^{(m)},\sigma^{(m)}\in\mathbb R^D\) are the
mean and element-wise standard deviation, and vector squares
are element-wise.
After \(M\) iterations, the final weighted candidates before
Gaussian projection define the empirical planner proposal
\(\pi_P(\tau\mid z_t)=w^{(M)}(\tau)\) for
\(\tau\in\mathcal T^{(M)}\), with first-action marginal
\(\pi_{P,0}(a_t\mid z_t)\).
Only the first planned action is executed before replanning.

\noindent\textbf{Policy-Planner Mismatch.}
In critic-guided MPPI-based methods, the critic learns from
planner-generated transitions in \(\mathcal B\), with actions
selected using \(\pi_{P,0}\), but bootstraps with actions
from \(\pi_{\theta_a}\).
Since \(\pi_{\theta_a}\) need not match \(\pi_{P,0}\),
the terminal value estimates used by MPPI may not reflect
the return under subsequent replanning
~\citep{wang2025bootstrapped,serra-gomez2026a}.
To reduce the mismatch, BMPC~\citep{wang2025bootstrapped}
learns \(\pi_{\theta_a}\) through planner imitation,
minimizing the Kullback--Leibler (KL) divergence
\(\mathrm{KL}(\pi_{P,0}^{\mathrm G}\|\pi_{\theta_a})\),
where \(\pi_{P,0}^{\mathrm G}\) is the Gaussian approximation
to \(\pi_{P,0}\).
Alternatively, BOOM~\citep{zhan2026bootstrap} combines
critic-guided policy updates with soft value-weighted alignment
to planner actions in replay, while
PO-MPC~\citep{serra-gomez2026a} regularizes policy optimization
toward an adaptive prior distilled from the planner's
first-action distribution.
\textit{We omit the common conditioning on \(z_t\) in following KL expressions for simplicity.
}

\section{Proposed SWAN}
\label{sec:method}
\vspace{-0.1in}
\begin{figure}[tbp]
    \vspace{-0.1in}
    \centering
    \captionsetup{font=small,skip=4pt}
    \includegraphics[width=\linewidth]{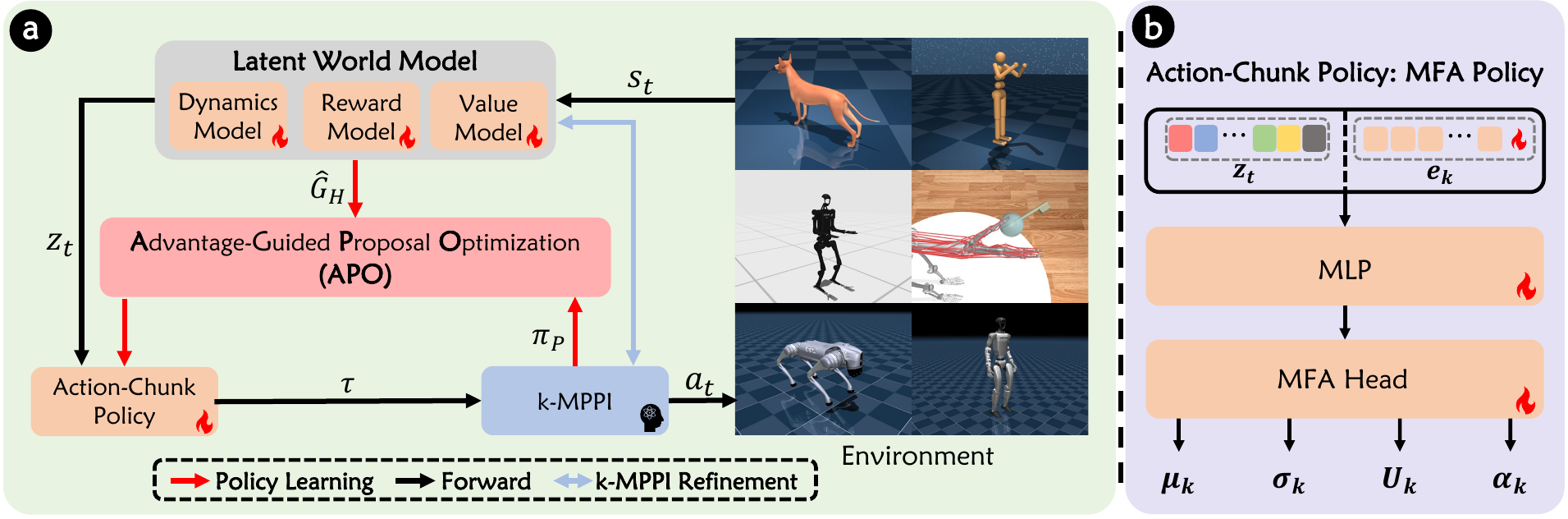}
    \caption{
    \textbf{Overview of the Proposed SWAN.}
    (a). The action-chunk policy initializes k-MPPI, which
    refines the proposal using predicted rewards and terminal
    value from the latent world model.
    APO constructs an improved target using advantage
    estimates and trains the policy toward the target while
    preserving planner alignment, providing initialization
    for subsequent planning.
    (b).
    The MFA policy models a multimodal distribution with correlations
    across time and action dimensions, enabling direct sampling
    of complete action chunks.}
    \label{fig:overview}
    \vspace{-0.2in}
\end{figure}
\noindent\textbf{Motivation}. 
Although the methods such as BMPC and PO-MPC
described in Sec.~\ref{sec:prelim} reduce
policy-planner mismatch, two gaps remain in learning
the complete planner proposal \(\pi_P\) as follows.

\noindent
{\textit{1) Proposal representation gap}}. Existing alignment objectives operate at the action
level, whereas the complete planner proposal factorizes as
\begin{equation}
\pi_P(\tau\mid z_t)
=
\pi_{P,0}(a_t\mid z_t)
\pi_P(a_{t+1:t+H-1}\mid a_t,z_t).
\label{eq:proposal_factorization}
\end{equation}
Matching \(\pi_{P,0}\) therefore does not determine the
conditional distribution of the remaining actions.

\noindent
{\textit{2) Proposal optimization gap}}. 
Even with complete action chunks, critic-guided updates remain
susceptible to mismatch since the temporal difference (TD)
bootstrapping  follows
\(\pi_{\theta_a}\), which need not match \(\pi_{P,0}\).
To reduce the mismatch, planner-derived regularization balances
planner alignment against value maximization.
In contrast, planner imitation uses
\(\pi_{P,0}^{\mathrm G}\) as the learning target.
However, extending imitation to complete chunks still uses
the current \(\pi_P\) as the learning target, leaving further
target improvement to MPPI refinement.


\noindent\textbf{Proposed Method.} 
To address both gaps, we introduce Fa\textbf{S}t \textbf{W}orld Pl\textbf{AN}ner (SWAN), an advantage-aware
MPPI-based MBRL framework shown in Fig.~\ref{fig:overview}. The proposed method combines k-MPPI and an MFA action-chunk policy for complete proposal representation with APO to improve the learning target while preserving planner
alignment, thereby improving policy guidance for efficient online planning. We detail the two novel components below.

\subsection{Multimodal Action-Chunk Proposal Representation}
\label{sec:mfa}
\vspace{-0.05in}
In this section, we introduce a novel approach to explicitly
representing the complete \(H\)-step planner proposal. Motivated by the diagonal Gaussian projection in
Eq.~\ref{eq:mppi_update}, which cannot capture multiple modes or
correlations across time and action dimensions, we propose
k-MPPI to preserve multiple search modes and the MFA
action-chunk policy to model the planner proposal. 

\noindent\textbf{k-MPPI.} 
k-MPPI initializes \(K\) local Gaussian search distributions from the corresponding policy components and maintains a global Gaussian for exploration. For each refinement iteration, both local and global Gaussians are updated from their respective elites using Eq.~\ref{eq:mppi_update} under a fixed total sampling budget.
After \(M\) refinement iterations, the local and global search
results form a proposal with \(K\) modes, each represented by
a batch \(\mathcal E_k\) of \(N\) elite action chunks.
To obtain candidate weights and mode probabilities, we concatenate
the batches into \(\mathcal E\) and obtain
\begin{equation}
w^P(\tau)
=
\operatorname{softmax}_{\mathcal E}
\!\left(
\widehat G_H(z_t,\tau)/\eta_{\mathrm{MPPI}}
\right),
\qquad
\alpha_k^P
=
\sum_{\tau\in\mathcal E_k}w^P(\tau),
\label{eq:kmppi_weights}
\end{equation}
where \(w^P(\tau)\) is the normalized weight of candidate \(\tau\),
\(\alpha_k^P\) is the probability of mode \(k\),
and superscript \(P\) denotes planner quantities.
The retained chunks and their weights define the empirical planner
proposal \(\pi_P\), with branch labels providing the mode
assignments for policy learning.
Appendix~\ref{app:kmppi} provides the complete k-MPPI algorithm.

\noindent\textbf{MFA action-chunk policy (Fig.~\ref{fig:overview} (b)).}
In order to learn the proposal constructed by k-MPPI, we associate each retained mode with a Gaussian component, following the Gaussian
projection in Eq.~\ref{eq:mppi_update}. We therefore parameterize \(\pi_\theta\) as a mixture of
factor analyzers (MFA)~\citep{ghahramani1996algorithm},
using a low-rank covariance term and diagonal residual
variances to \textit{capture correlations across time and action
dimensions} without parameterizing full covariance matrices.
The policy network with parameters \(\theta\) uses \(z_t\)
and learnable mode embeddings \(e_k\) to predict the MFA
parameters
\(\{(\alpha_k^\theta,\mu_k^\theta,U_k^\theta,
\sigma_k^\theta)\}_{k=1}^{K}\).
Omitting their dependence on \(z_t\), the policy is
\begin{equation}
\begin{aligned}
\pi_\theta(\tau\mid z_t)
&=
\sum_{k=1}^{K}\alpha_k^\theta
\mathcal N\!\left(
\tau\mid\mu_k^\theta,\Sigma_k^\theta
\right),\qquad
\Sigma_k^\theta
&=
U_k^\theta(U_k^\theta)^\top
+
\operatorname{Diag}\!\left((\sigma_k^\theta)^2\right),
\end{aligned}
\label{eq:mfa_policy}
\end{equation}
where \(\alpha_k^\theta\geq0\) is the component probability
with \(\sum_k\alpha_k^\theta=1\),
\(\mu_k^\theta\in\mathbb R^D\) is the mean,
\(U_k^\theta\in\mathbb R^{D\times r}\) is the factor loading
matrix with rank parameter \(r\ll D\), and
\(\sigma_k^\theta\in\mathbb R^D\) contains residual standard
deviations.
We select \(k\) with probability \(\alpha_k^\theta\)
and sample an action chunk without policy rollouts.

To support alignment with the action-chunk policy, we project
the empirical planner proposal \(\pi_P\) onto the same MFA
family, extending the Gaussian projection in
Eq.~\ref{eq:mppi_update}.
The projection preserves the mode probabilities \(\alpha_k^P\)
and produces the parameters
\(\{(\alpha_k^P,\mu_k^P,U_k^P,\sigma_k^P)\}_{k=1}^{K}\),
which define the continuous MFA approximation \(\widetilde\pi_P\),
\begin{equation}
\begin{aligned}
\widetilde\pi_P(\tau\mid z_t)
&=
\sum_{k=1}^{K}\alpha_k^P
\mathcal N\!\left(
\tau\mid\mu_k^P,\Sigma_k^P
\right),\qquad
\Sigma_k^P
&=
U_k^P(U_k^P)^\top
+
\operatorname{Diag}\!\left((\sigma_k^P)^2\right).
\end{aligned}
\label{eq:planner_mfa_approximation}
\end{equation}
We also detail the projection and sampling
procedures in Appendix~\ref{app:mfa_projection} and~\ref{app:mfa_sampling}.

\subsection{Advantage-Guided Proposal Optimization}
\label{sec:apo}
\vspace{-0.05in}
After establishing the complete proposal representation in Sec.~\ref{sec:mfa}, we
introduce APO to improve the planner proposal with advantage
estimates while preserving planner alignment.

\noindent\textbf{Advantage-guided proposal improvement.}
To improve the current planner proposal \(\pi_P\), we seek
an updated proposal \(\pi\) that maximizes
\(\mathbb E_{\tau\sim\pi}[\widehat G_H(z_t,\tau)]\).
Since improvement is measured relative to \(\pi_P\), we
express the expected return gain as
\(\mathbb E_{\tau\sim\pi}[A(\tau)]\), where
\(A(\tau)=\widehat G_H(z_t,\tau)
-\mathbb E_{\tau'\sim\pi_P}[\widehat G_H(z_t,\tau')]\).

However, advantage maximization alone cannot constrain
how far \(\pi\) departs from \(\pi_P\).
We therefore impose a KL trust region on the proposal update
at a given latent state \(z_t\) and solve
\begin{equation}
\max_{\pi}\quad
\mathbb E_{\tau\sim\pi}[\bar A(\tau)]
\quad
\text{s.t.}\quad
\mathrm{KL}(\pi\|\pi_P)\leq\epsilon_{\mathrm{MD}},
\label{eq:apo_improvement}
\end{equation}
where \(\epsilon_{\mathrm{MD}}\) is the KL radius and
\(\bar A(\tau)=A(\tau)/(s_A\sum_{h=0}^{H-1}\gamma^h)\)
is the advantage normalized to account for reward scale,
with \(s_A\) denoting the weighted standard deviation of
the horizon-normalized advantage under the current \(\pi_P\).
Then we derive the proposal update by introducing a Lagrange
multiplier \(\lambda\geq0\) for the KL constraint and forming
\begin{equation}
\mathcal L(\pi,\lambda)
=
\mathbb E_{\tau\sim\pi}[\bar A(\tau)]
-
\lambda
\left(
\mathrm{KL}(\pi\|\pi_P)
-
\epsilon_{\mathrm{MD}}
\right).
\label{eq:apo_lagrangian}
\end{equation}
For fixed \(\lambda>0\), removing the constant
\(\lambda\epsilon_{\mathrm{MD}}\) leaves a KL-regularized
advantage objective, corresponding to a mirror-descent update
from \(\pi_P\)~\citep{beck2003mirror}.
Thus, maximizing the Lagrangian over normalized distributions
yields the optimal target \(\pi^*\),
\begin{equation}
\pi^*(\tau\mid z_t)
=
\frac{
\pi_P(\tau\mid z_t)
\exp\!\left(\bar A(\tau)/\lambda\right)
}{
\mathbb E_{\tau'\sim\pi_P}
\left[
\exp\!\left(\bar A(\tau')/\lambda\right)
\right]
}.
\label{eq:apo_target}
\end{equation}
The target reuses MPPI evaluations to favor advantageous
action chunks without additional model rollouts.
With \(\lambda>0\) fixed in implementation, \(\pi^*\)
also solves Eq.~\ref{eq:apo_improvement} with the induced
KL radius
\(\epsilon_{\mathrm{MD}}=\mathrm{KL}(\pi^*\|\pi_P)\), as shown in Appendix~\ref{app:apo_derivation}.

\noindent\textbf{Planner alignment.}
After obtaining the optimal target \(\pi^*\), we learn
\(\pi_\theta\) to incorporate the proposal improvement into
subsequent planning.
For policy learning, we construct the MFA approximation
\(\widetilde\pi^*\) of \(\pi^*\) using the projection in
Sec.~\ref{sec:mfa}.
However, the trust region in Eq.~\ref{eq:apo_improvement}
constrains \(\pi^*\) rather than \(\pi_\theta\).
We thus introduce a separate forward-KL trust region
around \(\widetilde\pi_P\) to preserve planner alignment
and optimize the policy parameters over replay states \(s_t\),
\begin{equation}
\begin{aligned}
\min_\theta\quad&
\mathbb E_{s_t\sim\mathcal B}
\left[
\mathrm{KL}(\widetilde\pi^*\|\pi_\theta)
\right]\quad
\text{s.t.}\quad&
\mathbb E_{s_t\sim\mathcal B}
\left[
\mathrm{KL}(\widetilde\pi_P\|\pi_\theta)
\right]
\leq\epsilon,
\end{aligned}
\label{eq:apo_alignment}
\end{equation}
where \(\epsilon\) denotes the budget for the average alignment KL in the constrained objective.
The MFA KL terms retain branch labels, as defined in
Appendix~\ref{app:mfa_kl}.
To optimize the constrained policy objective, we introduce
a Lagrange multiplier \(\beta\geq0\) for the alignment
constraint and form
\begin{equation}
\begin{aligned}
\mathcal L_{\mathrm{APO}}(\theta,\beta)
={}&
\mathbb E_{s_t\sim\mathcal B}
\left[
\mathrm{KL}(\widetilde\pi^*\|\pi_\theta)
\right]
+
\beta
\left(
\mathbb E_{s_t\sim\mathcal B}
\left[
\mathrm{KL}(\widetilde\pi_P\|\pi_\theta)
\right]
-\epsilon
\right).
\end{aligned}
\label{eq:apo_loss}
\end{equation}
Rather than fixing \(\beta\), we alternate policy minimization
with projected dual ascent,
adapting the penalty to constraint violation.
Since the unnormalized dual update depends on the absolute KL scale,
we use an exponential moving average (EMA) of the alignment KL
as the current budget and normalize the update accordingly
\begin{equation}
\beta
\leftarrow
\left[
\beta+\eta_\beta
\left(
\frac{
\mathbb E_{s_t\sim\mathcal B}
\left[
\mathrm{KL}(\widetilde\pi_P\|\pi_\theta)
\right]
}{
\epsilon
}
-1
\right)
\right]_+,
\quad
\epsilon
=
\left[
\mathbb E_{s_t\sim\mathcal B}
\left[
\mathrm{KL}(\widetilde\pi_P\|\pi_\theta)
\right]
\right]_{\mathrm{EMA}},
\label{eq:apo_beta}
\end{equation}
where \(\eta_\beta\) is the dual step size,
\([\cdot]_{\mathrm{EMA}}\) denotes the exponential moving
average, and \([\cdot]_+\) projects onto nonnegative values.
The current budget \(\epsilon\) is held fixed during each
policy and dual update.
Further, Appendix~\ref{app:unified_view} presents a
\textbf{unified view} that derives the core policy learning
objectives of APO and prior methods from a common formulation,
clarifying their connections and differences.

\subsection{Theoretical Analysis}
\label{sec:theory}
\vspace{-0.05in}
We further theoretically analyze proposal improvement at \(z_t\) with the latent world model held fixed and define
\(\widehat J_H(\pi)= \mathbb E_{\tau\sim\pi}[\widehat G_H(z_t,\tau)]\).

\begin{restatable}[Exact target improvement]{theorem}{apotargetgain}
\label{thm:apo_target_gain}
For \(\lambda>0\), the optimal target in
Eq.~\ref{eq:apo_target} satisfies
\begin{equation}
\widehat J_H(\pi^*)-\widehat J_H(\pi_P)
=
\mathbb E_{\tau\sim\pi^*}[A(\tau)]
\geq0.
\label{eq:apo_exact_gain}
\end{equation}
\end{restatable}

Transferring the target gain to \(\pi_\theta\) requires
MFA projection and policy learning.
We thus bound the resulting improvement over ideal
planner imitation, defined by
\(\pi_{\mathrm{IL}}=\widetilde\pi_P\).

\begin{restatable}[Projection-aware policy improvement]
{theorem}{apopolicygain}
\label{thm:apo_policy_gain}
Assume the expected returns and KL divergence below are
finite and there exists a finite constant \(A_{\max}>0\)
such that
\(\lvert\widehat G_H(z_t,\tau)
-\widehat J_H(\widetilde\pi_P)\rvert\leq A_{\max}\)
for every \(\tau\) in the union of the supports of
\(\pi_\theta\), \(\widetilde\pi^*\), and
\(\widetilde\pi_P\).
Then, for any \(\beta\geq0\),
\begin{equation}
\widehat J_H(\pi_\theta)-\widehat J_H(\pi_{\mathrm{IL}})
\geq
\frac{
\widehat J_H(\widetilde\pi^*)
-\widehat J_H(\widetilde\pi_P)
}{1+\beta}
-
A_{\max}
\sqrt{
\frac{
2\left[
\mathrm{KL}(\widetilde\pi^*\|\pi_\theta)
+\beta\,\mathrm{KL}(\widetilde\pi_P\|\pi_\theta)
\right]
}{1+\beta}
}.
\label{eq:apo_policy_bound}
\end{equation}
\end{restatable}

Together, the above theorems establish nonnegative target improvement
for APO and a sufficient condition for improvement over ideal
planner imitation after MFA projection and policy learning.
We also provide the proofs of
both theorems in Appendix~\ref{app:apo_proofs}.


\section{Experiments}\label{sec:experiment}
\vspace{-0.1in}

We first evaluate SWAN in terms of \textit{(i)} control performance, sample efficiency, and
online planning efficiency,
\textit{(ii)} the benefit of complete \(H\)-step planner
proposal representation with k-MPPI and MFA under matched
planning budgets, and
\textit{(iii)} whether APO improves control performance
over planner imitation under reduced MPPI refinement budgets.
We further evaluate SWAN on Unitree G1 and
Go2 in simulation, then deploy it on a physical Go2 robot for 
real-world validation.

\noindent\textbf{Benchmarks.}
We evaluate 31 tasks comprising 7 from DMControl~\citep{tunyasuvunakool2020},
14 from HumanoidBench~\citep{sferrazza2024humanoidbench}, and
10 from MyoSuite~\citep{caggiano2022myosuite}, spanning locomotion
and manipulation with maximum action dimensions of \(38,61,39\),
respectively.
Additional Unitree G1 and Go2 tasks~\citep{zakka2026mjlab} have
\(29\) and \(12\) action dimensions.
Appendix~\ref{app:task_details} lists all tasks and descriptions.

\noindent\textbf{Baselines.}
We compare against planning-based BMPC~\citep{wang2025bootstrapped},
BOOM~\citep{zhan2026bootstrap}, and PO-MPC~\citep{serra-gomez2026a},
imagination-based DreamerV3~\citep{hafner2023mastering}, and
model-free FPO~\citep{mcallister2026flow} and
SAC~\citep{haarnoja2018soft}.
For action-chunk comparison, \textit{BMPC-DP} replaces BMPC's
one-step policy with a diffusion policy~\citep{chi2025diffusion},
retaining MPPI and learning the complete \(H\)-step planner proposal 
through planner imitation.


\noindent\textbf{Implementation Details.}
We match environment steps across methods.
In the \textbf{full-budget} comparison, MPPI-based baselines use
\(M=8\) refinement iterations, while SWAN
uses \(M=4\).
In the \textbf{half-budget} comparison, all MPPI-based methods
use \(M=4\).
We provide detailed implementation settings and hyperparameters in Appendix~\ref{sec:appendix_training_setting}.

\subsection{Main Results}\label{subsec:main_result}
\begin{figure}[htb]
    \vspace{-0.15in}
    \centering
    \captionsetup{font=small,skip=4pt}
    \includegraphics[width=\linewidth]{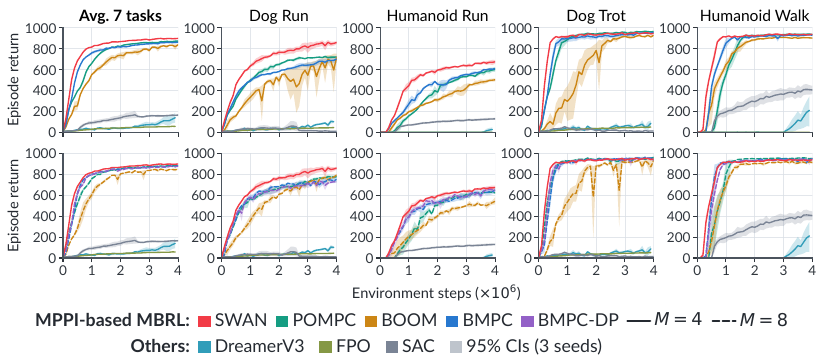}
    \caption{\textbf{DMControl locomotion suite.} 
    SWAN uses \(M=4\) MPPI refinement iterations,
    while MPPI-based baselines use \(M=4\) and \(M=8\) in the
    top (half-budget) and bottom (full-budget) rows, respectively.
    Curves and shading show means and 95\% CIs over
    $3$ random seeds.
    }
    \label{fig:benchmark_dmc}
    \vspace{-0.15in}
\end{figure}

\begin{figure}[htb]
    \vspace{-0.15in}
    \centering
    \captionsetup{font=small,skip=4pt}
    \includegraphics[width=\linewidth]{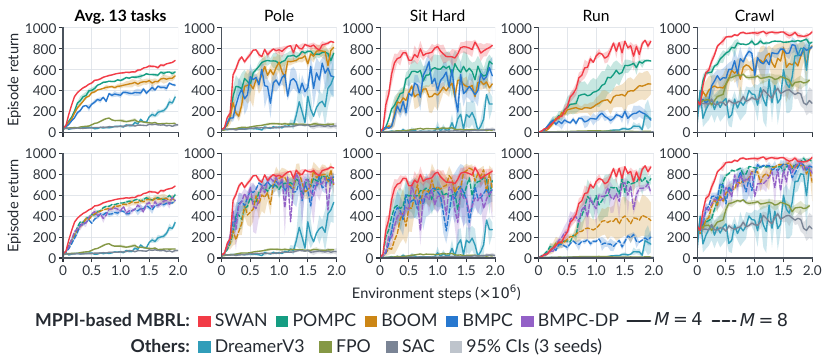}
    \caption{\textbf{HumanoidBench locomotion suite.} 
    SWAN uses \(M=4\) MPPI refinement iterations,
    while MPPI-based baselines use \(M=4\) and \(M=8\) in the
    top (half-budget) and bottom (full-budget) rows, respectively.
    Curves and shading show means and 95\% CIs over
    $3$ random seeds. The suite average excludes Reach.}
    \label{fig:benchmark_humanoidbench}
    \vspace{-0.1in}
\end{figure}

\begin{figure}[htb]
    \vspace{-0.05in}
    \centering
    \captionsetup{font=small,skip=4pt}
    \includegraphics[width=\linewidth]{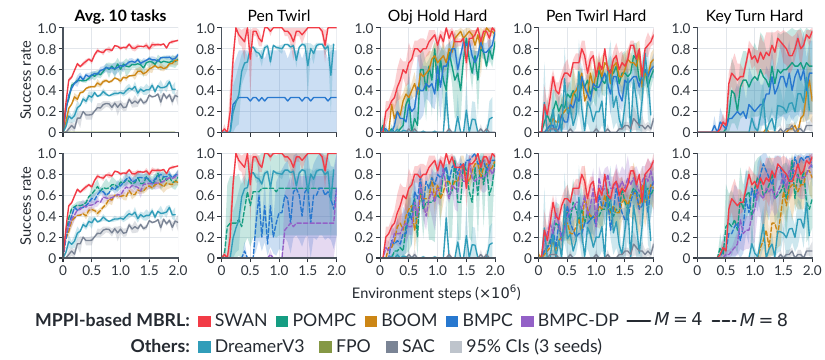}
    \caption{\textbf{MyoSuite manipulation suite.} 
    SWAN uses \(M=4\) MPPI refinement iterations,
    while MPPI-based baselines use \(M=4\) and \(M=8\) in the
    top (half-budget) and bottom (full-budget) rows, respectively.
    Curves and shading show means and 95\% CIs over
    $3$ random seeds.}
    \label{fig:benchmark_myosuite}
    \vspace{-0.25in}
\end{figure}

\vspace{-0.05in}
\noindent\textbf{Control performance and sample efficiency.}
As shown in Figs.~\ref{fig:benchmark_dmc},
\ref{fig:benchmark_humanoidbench}, and
\ref{fig:benchmark_myosuite}, we compare SWAN's control
performance and sample efficiency with baselines under
full- and half-budget settings.
Each figure plots means over environment steps for the suite
average and four representative tasks, with shaded 95\%
confidence intervals (CIs) over $3$ random seeds.
We provide all task curves in Appendix~\ref{app:all_task_curves}.

In the full-budget comparison, SWAN with \(M=4\) refinement
iterations outperforms MPPI-based baselines with \(M=8\)
in suite-average performance.
Following threshold-based evaluations
~\citep{wang2025bootstrapped,lange2023improving}, we measure
sample efficiency by \(T_{80}\), the environment steps until
the mean evaluation curve first reaches \(80\%\) of SWAN's
final performance, a common threshold across methods for
each task or suite average.
On DMControl, SWAN achieves \(898.8\) average return versus
PO-MPC's \(883.7\), with \(T_{80}=0.60\)M versus \(1.00\)M,
a \(40\%\) reduction.
In particular, SWAN exceeds BOOM on \textit{Dog-Run}
and BMPC on \textit{Humanoid-Run} by \(67.2\) and \(25.8\)
in episode return, with \(53.6\%\) and \(33.3\%\) lower
\(T_{80}\), respectively.
On HumanoidBench, SWAN reaches \(688.0\) versus PO-MPC's \(604.0\),
while reducing \(T_{80}\) from \(1.25\)M to \(0.90\)M by \(28\%\).
Against the same PO-MPC baseline, SWAN improves returns
on \textit{Run} and \textit{Sit Hard} by \(112.2\) and
\(92.0\), with \(37.9\%\) and \(62.5\%\) lower \(T_{80}\),
respectively.
The gains extend to MyoSuite, where SWAN attains \(0.88\) average
success rate compared with BMPC's \(0.81\), 
with \(T_{80}=0.55\)M versus \(1.05\)M, a \(47.6\%\) reduction.
In particular, on \textit{Pen Twirl}, SWAN reaches \(1.0\) success and
\(T_{80}=0.25\)M, whereas the MPPI-based baselines reach at most
\(0.667\) and do not reach the threshold during evaluation.
Additionally, the half-budget setting (\(M=4\)) lowers
BMPC, BOOM, and PO-MPC's performance,
reflecting continued dependence on MPPI refinement.
SWAN's average return gains over PO-MPC widen to \(25.3\)
and \(110.5\) on DMControl and HumanoidBench, with
\(50.0\%\) and \(37.9\%\) lower \(T_{80}\), respectively.
On MyoSuite, SWAN's average success rate exceeds BMPC's
by \(0.137\), with \(63.3\%\) lower \(T_{80}\).
All final performance and \(T_{80}\) results are in Appendix~\ref{app:sample_efficiency}.

\noindent\textbf{Online planning efficiency.}
We next assess whether the control gains are accompanied by
\begin{wraptable}[11]{r}{0.5\textwidth}
    \vspace{-0.1in}
    \centering
    \small
    \setlength{\tabcolsep}{3pt}
    \captionsetup{font=small,skip=4pt}
    \caption{
    Evaluation efficiency averaged over 7 DMControl tasks over
    10 episodes with planning horizon \(H=3\).
    Relative time is \(T_{\mathrm{method}}/T_{\mathrm{SWAN}}\).
    }
    \label{tab:dmc_planning_latency}
    \begin{tabular*}{\linewidth}{@{\extracolsep{\fill}}lccc@{}}
        \toprule
        Method &
        \makecell{Trainable\\Params. (M) $\downarrow$} &
        \makecell{Eval.\\Time (s) $\downarrow$} &
        \makecell{Rel.\\Time $\downarrow$} \\
        \midrule
        BMPC    & \textbf{3.29} & 27.90  & $1.54\times$ \\
        BOOM    & 5.11 & 54.26  & $2.99\times$ \\
        PO-MPC  & 8.64 & 33.17  & $1.83\times$ \\
        BMPC-DP & 10.56 & 234.38 & $12.93\times$ \\
        \midrule
        SWAN (ours) &
        3.67 &
        \textbf{18.12} &
        $\mathbf{1.00}\boldsymbol{\times}$ \\
        \bottomrule
    \end{tabular*}
\end{wraptable}
lower end-to-end wall-clock evaluation cost,
averaging evaluation times
over seven DMControl tasks with \(10\) episodes per task
and planning horizon \(H=3\) on a single NVIDIA RTX A6000.
As shown in Tab.~\ref{tab:dmc_planning_latency}, SWAN takes
\(18.12\) seconds, \(35.1\%\) less than BMPC, while BOOM,
PO-MPC, and BMPC-DP take \(2.99\times\), \(1.83\times\),
and \(12.93\times\) as long, respectively.
Moreover, according to Fig.~\ref{fig:benchmark_dmc} and Tab.~\ref{tab:dmc_planning_latency}, SWAN achieves stronger 
control performance than BMPC at lower evaluation
cost, with \(3.67\)M trainable parameters compared with \(3.29\)M.

\noindent\textbf{Findings from main results.}
We further compare SWAN, BMPC-DP and BMPC
in Figs.~\ref{fig:benchmark_dmc},
\ref{fig:benchmark_humanoidbench},
\ref{fig:benchmark_myosuite}.
BMPC-DP learns complete action chunks with Diffusion Policy,
yet remains close to BMPC in suite-average performance,
despite much higher evaluation cost.
As discussed in Sec.~\ref{sec:relatedwork}, online proposal
learning requires fast sampling and adaptation to changing
planner proposals.
Iterative denoising adds cost, while the limited gains may
reflect difficulty adapting an implicit representation online.
In contrast, SWAN uses k-MPPI and MFA to explicitly model
proposal modes and correlations across time and action
dimensions.
APO further improves the learning target using advantage
estimates while preserving planner alignment in policy learning.
Thus, results support combining explicit proposal
representation with advantage-guided proposal optimization
for efficient online planning.
\begin{wrapfigure}[17]{r}{0.5\columnwidth}
    \vspace{-0.15in}
    \centering
    \captionsetup{font=small,skip=4pt}
    \captionsetup[subfigure]{font=footnotesize,skip=2pt}
    \begin{subfigure}[t]{0.49\linewidth}
        \centering
        \includegraphics[width=\linewidth]{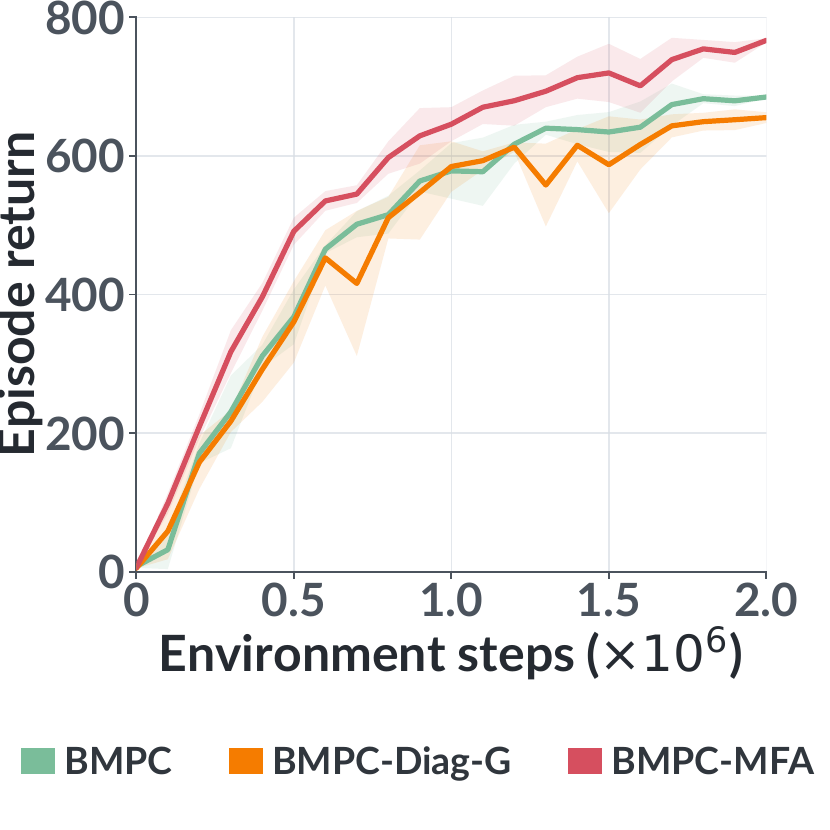}
        \caption{Proposal representation.}
        \label{fig:ablation_representation}
    \end{subfigure}\hfill
    \begin{subfigure}[t]{0.49\linewidth}
        \centering
        \includegraphics[width=\linewidth]{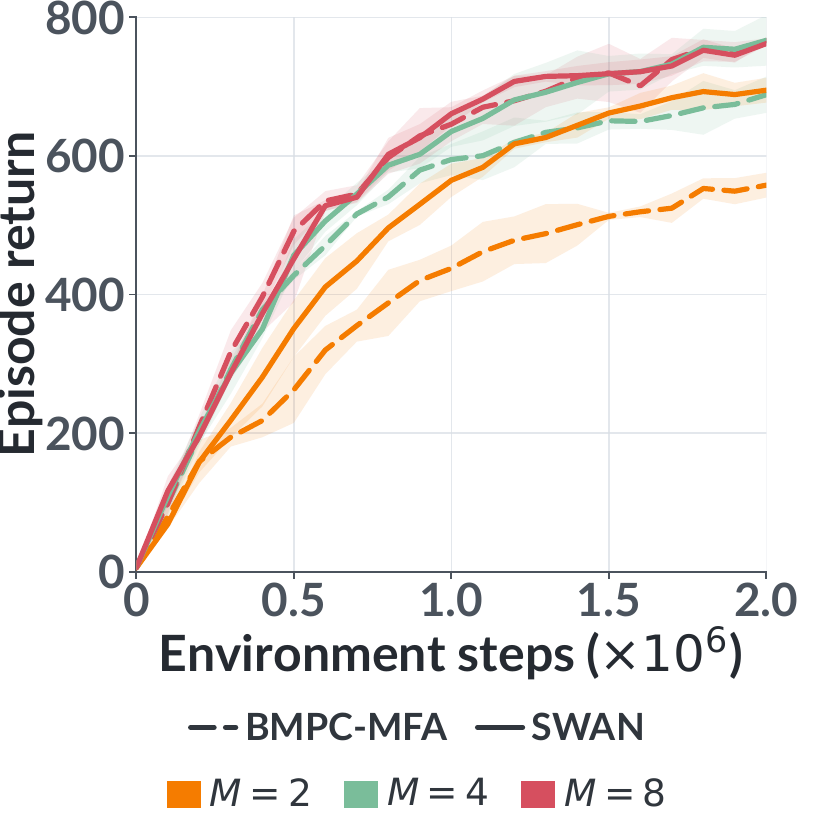}
        \caption{Effect of APO.}
        \label{fig:ablation_iterations}
    \end{subfigure}
    \caption{
    \textbf{Ablations on \textit{Dog-Run}.} \textbf{(a)} Comparison of proposal
    representations in BMPC, BMPC-Diag-G, and BMPC-MFA, all using planner imitation under matched planning budgets. \textbf{(b)} Comparing SWAN (solid) with BMPC-MFA (dashed) under \(M=2,4,8\) MPPI refinement iterations. Curves and shading show means and 95\% CIs over $3$ random seeds.}
    \label{fig:ablation}
\end{wrapfigure}
\noindent\textbf{Effect of action-chunk proposal representation.}
We compare proposal representations on \textit{Dog-Run}
under matched training and planning budgets (\(M=8\)),
retaining BMPC's planner imitation.
BMPC uses a one-step policy, whereas \textit{BMPC-Diag-G}
models complete \(H\)-step chunks with a single diagonal
Gaussian.
\textit{BMPC-MFA}, denoted BMPC-Chunk in
Fig.~\ref{fig:motivation}, instead combines k-MPPI and MFA.
As shown in Fig.~\ref{fig:ablation_representation},
BMPC-MFA's gain over BMPC supports our complete proposal
representation, while the gain over BMPC-Diag-G highlights
the benefits of explicitly modeling multiple modes and
correlations across time and action dimensions.

\noindent\textbf{Effect of APO.} 
With the k-MPPI and MFA policy, we then isolate
APO by comparing SWAN with BMPC-MFA, which retains planner
imitation, under \(M=2,4,8\) MPPI refinement iterations.
Fig.~\ref{fig:ablation_iterations} shows larger gains
from APO at \(M=2\) and \(M=4\), while the gap narrows
at \(M=8\).
Moreover, SWAN with \(M=4\) approaches its performance
with \(M=8\), whereas BMPC-MFA benefits more from
the additional refinement.
The comparison therefore shows that policy learning
with APO reduces dependence on MPPI refinement while
maintaining strong control performance.


\vspace{-0.02in}
\subsection{Sim-to-Real Transfer}
\label{subsec:real_world}
\vspace{-0.08in}
We further train and evaluate SWAN on simulated Unitree G1
and Go2 velocity-tracking tasks~\citep{zakka2026mjlab},
where it outperforms BMPC in final return and sample efficiency
with half as many MPPI refinement iterations per task.
\textbf{We then demonstrate zero-shot sim-to-real transfer
to a physical Go2.}
The robot tracks forward, backward, and lateral velocity
commands, turns in both directions, and follows figure-eight
trajectories.
Motion capture measurements over selected figure-eight and
straight-line segments give planar velocity mean absolute
errors (MAEs) of \(0.126\) and \(0.149\,\mathrm{m/s}\),
with yaw angular velocity MAEs of \(0.112\) and
\(0.068\,\mathrm{rad/s}\), respectively.
We further assess locomotion stability with external pushes
during motion, observing stable locomotion without falling.
With only \(M=3\) MPPI refinement iterations per control step,
the controller operates at average \(308.16\,\mathrm{Hz}\) on NVIDIA GeForce RTX 5090.
We detail simulation and
real-world results in Appendix~\ref{app:unitree_results}.
Overall, our SWAN achieve reliable zero-shot sim-to-real transfer ability in stable high-frequency locomotion control.
Demonstrations are available at
~\url{https://swanfastworldplanner.github.io/}
\vspace{-0.05in}

\section{Conclusion and Limitation}\label{sec:conclusion}
\vspace{-0.1in}
We proposed SWAN, an advantage-aware MPPI-based MBRL framework that improves online planning efficiency through better policy guidance.
SWAN combines k-MPPI with an MFA action-chunk policy to capture multimodal proposal structure and introduces APO, which uses advantage estimates to construct an improved target from the planner proposal while preserving planner alignment during policy learning. Extensive experiments across multiple continuous-control benchmarks demonstrated strong control performance and sample efficiency while requiring fewer MPPI refinement iterations. We further demonstrated SWAN's zero-shot sim-to-real transfer
to a physical Unitree Go2 for locomotion.
One limitation is SWAN's current focus on state-based, task-specific continuous-control settings. Future work will extend SWAN to multimodal observations and multi-task learning, enabling shared proposal learning across diverse tasks and sensory modalities.

\bibliography{iclr2027_conference}

@article{m2023model,
  title={Model-based reinforcement learning: A survey},
  author={M. Moerland, Thomas and Broekens, Joost and Plaat, Aske and M. Jonker, Catholijn},
  journal={Foundations and Trends in Machine Learning},
  volume={16},
  number={1},
  pages={1--118},
  year={2023},
  publisher={Emerald Publishing Limited}
}

@article{hafner2019dream,
  title={Dream to control: Learning behaviors by latent imagination},
  author={Hafner, Danijar and Lillicrap, Timothy and Ba, Jimmy and Norouzi, Mohammad},
  journal={arXiv preprint arXiv:1912.01603},
  year={2019}
}

@article{hafner2020mastering,
  title={Mastering atari with discrete world models},
  author={Hafner, Danijar and Lillicrap, Timothy and Norouzi, Mohammad and Ba, Jimmy},
  journal={arXiv preprint arXiv:2010.02193},
  year={2020}
}

@article{hafner2023mastering,
  title={Mastering diverse domains through world models},
  author={Hafner, Danijar and Pasukonis, Jurgis and Ba, Jimmy and Lillicrap, Timothy},
  journal={arXiv preprint arXiv:2301.04104},
  year={2023}
}

@article{janner2019trust,
  title={When to trust your model: Model-based policy optimization},
  author={Janner, Michael and Fu, Justin and Zhang, Marvin and Levine, Sergey},
  journal={Advances in neural information processing systems},
  volume={32},
  year={2019}
}

@inproceedings{hafner2019learning,
  title={Learning latent dynamics for planning from pixels},
  author={Hafner, Danijar and Lillicrap, Timothy and Fischer, Ian and Villegas, Ruben and Ha, David and Lee, Honglak and Davidson, James},
  booktitle={International conference on machine learning},
  pages={2555--2565},
  year={2019},
  organization={PMLR}
}

@inproceedings{hansen2024td,
  title={Td-mpc2: Scalable, robust world models for continuous control},
  author={Hansen, Nick and Su, Hao and Wang, Xiaolong},
  booktitle={International Conference on Learning Representations},
  volume={2024},
  pages={47376--47405},
  year={2024}
}

@inproceedings{wang2025bootstrapped,
  title={Bootstrapped model predictive control},
  author={Wang, Yuhang and Guo, Hanwei and Wang, Sizhe and Qian, Long and Lan, Xuguang},
  booktitle={International conference on learning representations},
  volume={2025},
  pages={54241--54259},
  year={2025}
}

@article{zhan2026bootstrap,
  title={Bootstrap off-policy with world model},
  author={Zhan, Guojian and Wang, Likun and Zhang, Xiangteng and Gao, Jiaxin and Tomizuka, Masayoshi and Li, Shengbo Eben},
  journal={Advances in Neural Information Processing Systems},
  volume={38},
  pages={134093--134121},
  year={2026}
}

@inproceedings{
  serra-gomez2026a,
  title={A {KL}-regularization framework for learning to plan with adaptive priors},author={Alvaro Serra-Gomez and Daniel Jarne Ornia and Dhruva Tirumala and Thomas M. Moerland},
  booktitle={Forty-third International Conference on Machine Learning},
  year={2026},
  url={https://openreview.net/forum?id=zO8vzSGgTn}
}

@article{williams2017model,
  title={Model predictive path integral control: From theory to parallel computation},
  author={Williams, Grady and Aldrich, Andrew and Theodorou, Evangelos A},
  journal={Journal of Guidance, Control, and Dynamics},
  volume={40},
  number={2},
  pages={344--357},
  year={2017},
  publisher={American Institute of Aeronautics and Astronautics}
}

@inproceedings{lambert2021stein,
  title={Stein Variational Model Predictive Control},
  author={Lambert, Alexander and Ramos, Fabio and Boots, Byron and Fox, Dieter and Fishman, Adam},
  booktitle={Proceedings of the 2020 Conference on Robot Learning},
  pages={1278--1297},
  year={2021},
  volume={155},
  series={Proceedings of Machine Learning Research},
  publisher={PMLR}
}

@inproceedings{honda2024stein,
  title={Stein variational guided model predictive path integral control: Proposal and experiments with fast maneuvering vehicles},
  author={Honda, Kohei and Akai, Naoki and Suzuki, Kosuke and Aoki, Mizuho and Hosogaya, Hirotaka and Okuda, Hiroyuki and Suzuki, Tatsuya},
  booktitle={2024 IEEE International Conference on Robotics and Automation (ICRA)},
  pages={7020--7026},
  year={2024},
  organization={IEEE}
}

@article{chi2025diffusion,
  title={Diffusion policy: Visuomotor policy learning via action diffusion},
  author={Chi, Cheng and Xu, Zhenjia and Feng, Siyuan and Cousineau, Eric and Du, Yilun and Burchfiel, Benjamin and Tedrake, Russ and Song, Shuran},
  journal={The International Journal of Robotics Research},
  volume={44},
  number={10-11},
  pages={1684--1704},
  year={2025},
  publisher={Sage Publications Sage UK: London, England}
}

@inproceedings{
  chisari2024learning,
  title={Learning Robotic Manipulation Policies from Point Clouds with Conditional Flow Matching},
  author={Eugenio Chisari and Nick Heppert and Max Argus and Tim Welschehold and Thomas Brox and Abhinav Valada},
  booktitle={8th Annual Conference on Robot Learning},
  year={2024},
  url={https://openreview.net/forum?id=vtEn8NJWlz}
}

@inproceedings{
  jiang2025streaming,
  title={Streaming Flow Policy: Simplifying diffusion/flow-matching policies by treating action trajectories as flow trajectories},
  author={Sunshine Jiang and Xiaolin Fang and Nicholas Roy and Tom{\'a}s Lozano-P{\'e}rez and Leslie Pack Kaelbling and Siddharth Ancha},
  booktitle={9th Annual Conference on Robot Learning},
  year={2025},
  url={https://openreview.net/forum?id=jnpILGz9gQ}
}

@inproceedings{wang2025onestep,
  title={One-Step Diffusion Policy: Fast Visuomotor Policies via Diffusion Distillation},
  author={Zhendong Wang and Max Li and Ajay Mandlekar and Zhenjia Xu and Jiaojiao Fan and Yashraj Narang and Linxi Fan and Yuke Zhu and Yogesh Balaji and Mingyuan Zhou and Ming-Yu Liu and Yu Zeng},
  booktitle={Forty-second International Conference on Machine Learning},
  year={2025},
  url={https://openreview.net/forum?id=E2VsqgKNlr}
}

@inproceedings{ma2025efficient,
  title={Efficient Online Reinforcement Learning for Diffusion Policy},
  author={Haitong Ma and Tianyi Chen and Kai Wang and Na Li and Bo Dai},
  booktitle={Forty-second International Conference on Machine Learning},
  year={2025},
  url={https://openreview.net/forum?id=6Anv3KB9lz}
}

@inproceedings{schulman2015trust,
  title={Trust region policy optimization},
  author={Schulman, John and Levine, Sergey and Abbeel, Pieter and Jordan, Michael and Moritz, Philipp},
  booktitle={International conference on machine learning},
  pages={1889--1897},
  year={2015},
  organization={Pmlr}
}

@inproceedings{schulman2016gae,
  title={High-Dimensional Continuous Control Using Generalized Advantage Estimation},
  author={Schulman, John and Moritz, Philipp and Levine, Sergey and Jordan, Michael I. and Abbeel, Pieter},
  booktitle={International Conference on Learning Representations},
  year={2016}
}

@article{schulman2017proximal,
  title={Proximal policy optimization algorithms},
  author={Schulman, John and Wolski, Filip and Dhariwal, Prafulla and Radford, Alec and Klimov, Oleg},
  journal={arXiv preprint arXiv:1707.06347},
  year={2017}
}

@article{wang2018exponentially,
  title={Exponentially weighted imitation learning for batched historical data},
  author={Wang, Qing and Xiong, Jiechao and Han, Lei and Liu, Han and Zhang, Tong and others},
  journal={Advances in Neural Information Processing Systems},
  volume={31},
  year={2018}
}

@article{peng2019advantage,
  title={Advantage-weighted regression: Simple and scalable off-policy reinforcement learning},
  author={Peng, Xue Bin and Kumar, Aviral and Zhang, Grace and Levine, Sergey},
  journal={arXiv preprint arXiv:1910.00177},
  year={2019}
}

@inproceedings{kostrikov2022implicit,
  title={Offline Reinforcement Learning with Implicit Q-Learning},
  author={Kostrikov, Ilya and Nair, Ashvin and Levine, Sergey},
  booktitle={International Conference on Learning Representations},
  year={2022}
}

@inproceedings{sikchi2022learning,
  title={Learning off-policy with online planning},
  author={Sikchi, Harshit and Zhou, Wenxuan and Held, David},
  booktitle={Conference on Robot Learning},
  pages={1622--1633},
  year={2022},
  organization={PMLR}
}

@inproceedings{Hansen2022tdmpc,
  title={Temporal Difference Learning for Model Predictive Control},
  author={Nicklas Hansen and Xiaolong Wang and Hao Su},
  booktitle={ICML},
  year={2022}
}

@inproceedings{pinneri2021sample,
  title={Sample-efficient cross-entropy method for real-time planning},
  author={Pinneri, Cristina and Sawant, Shambhuraj and Blaes, Sebastian and Achterhold, Jan and Stueckler, Joerg and Rolinek, Michal and Martius, Georg},
  booktitle={Conference on Robot Learning},
  pages={1049--1065},
  year={2021},
  organization={PMLR}
}

@article{zhou2025diffusion,
  title={Diffusion Model Predictive Control},
  author={Guangyao Zhou and Sivaramakrishnan Swaminathan and Rajkumar Vasudeva Raju and J Swaroop Guntupalli and Wolfgang Lehrach and Joseph Ortiz and Antoine Dedieu and Miguel Lazaro-Gredilla and Kevin Patrick Murphy},
  journal={Transactions on Machine Learning Research},
  issn={2835-8856},
  year={2025}
}

@article{anthony2017thinking,
  title={Thinking fast and slow with deep learning and tree search},
  author={Anthony, Thomas and Tian, Zheng and Barber, David},
  journal={Advances in neural information processing systems},
  volume={30},
  year={2017}
}

@article{silver2017mastering,
  title={Mastering the game of go without human knowledge},
  author={Silver, David and Schrittwieser, Julian and Simonyan, Karen and Antonoglou, Ioannis and Huang, Aja and Guez, Arthur and Hubert, Thomas and Baker, Lucas and Lai, Matthew and Bolton, Adrian and others},
  journal={nature},
  volume={550},
  number={7676},
  pages={354--359},
  year={2017},
  publisher={Nature Publishing Group UK London}
}

@inproceedings{sacks2023learning,
  title={Learning sampling distributions for model predictive control},
  author={Sacks, Jacob and Boots, Byron},
  booktitle={Conference on Robot Learning},
  pages={1733--1742},
  year={2023},
  organization={PMLR}
}

@article{tunyasuvunakool2020,
         title = {dm\_control: Software and tasks for continuous control},
         journal = {Software Impacts},
         volume = {6},
         pages = {100022},
         year = {2020},
         issn = {2665-9638},
         doi = {https://doi.org/10.1016/j.simpa.2020.100022},
         url = {https://www.sciencedirect.com/science/article/pii/S2665963820300099},
         author = {Saran Tunyasuvunakool and Alistair Muldal and Yotam Doron and
                   Siqi Liu and Steven Bohez and Josh Merel and Tom Erez and
                   Timothy Lillicrap and Nicolas Heess and Yuval Tassa},
}

@inproceedings{caggiano2022myosuite,
  title     = {MyoSuite: A Contact-rich Simulation Suite for Musculoskeletal Motor Control},
  author    = {Caggiano, Vittorio and Wang, Huawei and Durandau, Guillaume and Sartori, Massimo and Kumar, Vikash},
  booktitle = {Proceedings of the 4th Annual Learning for Dynamics and Control Conference},
  series    = {Proceedings of Machine Learning Research},
  volume    = {168},
  pages     = {492--507},
  year      = {2022}
}

@article{zakka2026mjlab,
  title={mjlab: A lightweight framework for gpu-accelerated robot learning},
  author={Zakka, Kevin and Liao, Qiayuan and Yi, Brent and Lay, Louis Le and Sreenath, Koushil and Abbeel, Pieter},
  journal={arXiv preprint arXiv:2601.22074},
  year={2026}
}

@article{sferrazza2024humanoidbench,
  title={Humanoidbench: Simulated humanoid benchmark for whole-body locomotion and manipulation},
  author={Sferrazza, Carmelo and Huang, Dun-Ming and Lin, Xingyu and Lee, Youngwoon and Abbeel, Pieter},
  journal={arXiv preprint arXiv:2403.10506},
  year={2024}
}

@article{haarnoja2018soft,
  title={Soft actor-critic algorithms and applications},
  author={Haarnoja, Tuomas and Zhou, Aurick and Hartikainen, Kristian and Tucker, George and Ha, Sehoon and Tan, Jie and Kumar, Vikash and Zhu, Henry and Gupta, Abhishek and Abbeel, Pieter and others},
  journal={arXiv preprint arXiv:1812.05905},
  year={2018}
}

@inproceedings{mcallister2026flow,
  title={Flow matching policy gradients},
  author={McAllister, David and Ge, Songwei and Yi, Brent and Kim, Chung Min and Weber, Ethan and Choi, Hongsuk and Feng, Haiwen and Kanazawa, Angjoo},
  booktitle={International Conference on Learning Representations},
  volume={2026},
  pages={36352--36372},
  year={2026}
}

@techreport{ghahramani1996algorithm,
  title={The EM algorithm for mixtures of factor analyzers},
  author={Ghahramani, Zoubin and Hinton, Geoffrey E and others},
  year={1996},
  institution={Technical Report CRG-TR-96-1, University of Toronto}
}

@inproceedings{abdolmaleki2018maximum,
  title={Maximum a Posteriori Policy Optimisation},
  author={Abbas Abdolmaleki and Jost Tobias Springenberg and Yuval Tassa and Remi Munos and Nicolas Heess and Martin Riedmiller},
  booktitle={International Conference on Learning Representations},
  year={2018}
}

@inproceedings{lange2023improving,
  title={Improving reinforcement learning efficiency with auxiliary tasks in non-visual environments: A comparison},
  author={Lange, Moritz and Krystiniak, Noah and Engelhardt, Raphael C and Konen, Wolfgang and Wiskott, Laurenz},
  booktitle={International Conference on Machine Learning, Optimization, and Data Science},
  pages={177--191},
  year={2023},
  organization={Springer}
}

@inproceedings{
amigo2025first,
title={First Order Model-Based {RL} through Decoupled Backpropagation},
author={Joseph Amigo and Rooholla Khorrambakht and Elliot Chane-Sane and Nicolas Mansard and Ludovic Righetti},
booktitle={9th Annual Conference on Robot Learning},
year={2025}
}

@article{beck2003mirror,
  title={Mirror descent and nonlinear projected subgradient methods for convex optimization},
  author={Beck, Amir and Teboulle, Marc},
  journal={Operations Research Letters},
  volume={31},
  number={3},
  pages={167--175},
  year={2003},
  publisher={Elsevier}
}

@article{macfarlane2024spo,
  title={Spo: Sequential monte carlo policy optimisation},
  author={Macfarlane, Matthew V and Toledo, Edan and Byrne, Donal and Duckworth, Paul and Laterre, Alexandre},
  journal={Advances in Neural Information Processing Systems},
  volume={37},
  pages={1019--1057},
  year={2024}
}
\bibliographystyle{iclr2027_conference}

\clearpage
\appendix
\section{Notation}
\label{app:notation}

Table~\ref{tab:notation} summarizes the symbols used in the main text.

\begingroup
\small
\setlength{\tabcolsep}{5pt}
\renewcommand{\arraystretch}{1.08}
\begin{longtable}{@{}p{0.28\linewidth}p{\dimexpr0.72\linewidth-2\tabcolsep\relax}@{}}
\caption{Symbols used in the main text.}
\label{tab:notation}\\
\toprule
Symbol & Meaning \\
\midrule
\endfirsthead
\multicolumn{2}{l}{\textit{Table~\ref{tab:notation} continued}}\\
\toprule
Symbol & Meaning \\
\midrule
\endhead
\midrule
\multicolumn{2}{r}{\textit{Continued on the next page}}\\
\endfoot
\bottomrule
\endlastfoot
\multicolumn{2}{@{}l}{\textbf{Environment and value learning}}\\*
\(\mathcal M\) & Discounted Markov decision process, \(\mathcal M=(\mathcal S,\mathcal A,\mathcal P,\mathcal R,\gamma)\).\\
\(\mathcal S\) & Environment state space.\\
\(\mathcal A\) & Continuous action space.\\
\(\mathcal P\) & Environment transition kernel.\\
\(\mathcal R\) & Environment reward function.\\
\(\gamma\) & Discount factor.\\
\(t\) & Environment time step.\\
\(s_t\) & Environment state at time \(t\).\\
\(a_t\) & Action at time \(t\).\\
\(r_t\) & Reward at time \(t\).\\
\(d_s,d_a,d_z\) & Dimensions of the environment state, action, and latent state, respectively.\\
\(z_t\) & Encoded latent state, \(z_t=E_\psi(s_t)\).\\
\(E_\psi\) & Encoder mapping an environment state to a latent state.\\
\(F_\psi\) & Dynamics model predicting the next latent state.\\
\(R_\psi\) & Reward model predicting the immediate reward.\\
\(V_\phi\) & Value model approximating the state-value function and providing terminal value estimates.\\
\(\mathcal B\) & Replay buffer.\\
\(\pi_{\theta_a}\) & Learned one-step action policy used in prior methods.\\
\(\pi\) & Policy in the RL objective or proposal distribution optimized by APO, according to context.\\
\(J(\pi)\) & Expected infinite-horizon discounted environment return.\\

\midrule
\multicolumn{2}{@{}l}{\textbf{Planning and planner proposals}}\\*
\(H,h\) & Planning horizon and latent-rollout step index, respectively.\\
\(D\) & Action-chunk dimension, \(D=Hd_a\).\\
\(\tau\) & Complete \(H\)-step action chunk, \(\tau=a_{t:t+H-1}\).\\
\(\widehat z_h\) & Predicted latent state at rollout step \(h\).\\
\(M,m\) & Number of MPPI refinement iterations and iteration index, respectively.\\
\(\mathcal T^{(m)}\) & Candidate action-chunk batch at refinement iteration \(m\).\\
\(\mu^{(m)}\) & Mean of the Gaussian search distribution at iteration \(m\).\\
\(\sigma^{(m)}\) & Standard deviations of the Gaussian search distribution at iteration \(m\).\\
\(w^{(m)}(\tau)\) & Normalized weight of candidate \(\tau\) at iteration \(m\).\\
\(\eta_{\mathrm{MPPI}}\) & Temperature for MPPI candidate weighting.\\
\(\widehat G_H(z_t,\tau)\) & Estimated \(H\)-step return combining predicted rewards and the terminal value estimate.\\
\(K,k\) & Number of proposal modes and mode index, respectively.\\
\(N\) & Number of elite action chunks retained per mode in k-MPPI.\\
\(\mathcal E_k\) & Batch of elite action chunks retained in mode \(k\).\\
\(\mathcal E\) & Concatenation of all retained batches \(\mathcal E_k\).\\
\(w^P(\tau)\) & Candidate weight normalized over all retained action chunks.\\
\(\alpha_k^P\) & Planner probability of mode \(k\), obtained by summing candidate weights within \(\mathcal E_k\).\\
\(\pi_P\) & Empirical planner proposal over complete action chunks.\\
\(\pi_{P,0}\) & First-action marginal of the planner proposal \(\pi_P\).\\
\(\pi_{P,0}^{\mathrm G}\) & Gaussian approximation of the first-action marginal \(\pi_{P,0}\).\\

\midrule
\multicolumn{2}{@{}l}{\textbf{MFA representation}}\\*
\(\pi_\theta\) & MFA action-chunk policy.\\
\(\theta\) & Parameters of the MFA action-chunk policy.\\
\(e_k\) & Learnable embedding for mode \(k\).\\
\(r\) & Rank parameter of the factor covariance, with \(r\ll D\).\\
\(\alpha_k^\theta\) & Probability of component \(k\) predicted by the policy.\\
\(\mu_k^\theta\) & Mean of policy component \(k\).\\
\(U_k^\theta\) & Factor loading matrix of policy component \(k\), with dimensions \(D\times r\).\\
\(\sigma_k^\theta\) & Residual standard deviations of policy component \(k\).\\
\(\Sigma_k^\theta\) & Covariance of policy component \(k\).\\
\(\mu_k^P\) & Mean of planner component \(k\) after MFA projection.\\
\(U_k^P\) & Factor loading matrix of planner component \(k\) after MFA projection.\\
\(\sigma_k^P\) & Residual standard deviations of planner component \(k\) after MFA projection.\\
\(\Sigma_k^P\) & Covariance of planner component \(k\) after MFA projection.\\
\(\widetilde\pi_P\) & Continuous MFA approximation of the planner proposal \(\pi_P\).\\

\midrule
\multicolumn{2}{@{}l}{\textbf{APO and theoretical analysis}}\\*
\(A(\tau)\) & Advantage of chunk \(\tau\) relative to the
expected return under \(\pi_P\).\\
\(\bar A(\tau)\) & Advantage normalized by the discounted
horizon and the weighted standard deviation,
\(\bar A(\tau)=A(\tau)/(s_A\sum_{h=0}^{H-1}\gamma^h)\).\\
\(s_A\) & Weighted standard deviation of the horizon-normalized
advantage under \(\pi_P\) and floored at \(10^{-6}\).\\
\(\epsilon_{\mathrm{MD}}\) & KL radius for the proposal improvement step.\\
\(\lambda\) & Lagrange multiplier for the proposal improvement constraint.\\
\(\mathcal L(\pi,\lambda)\) & Lagrangian for advantage-guided proposal improvement.\\
\(\pi^*\) & Optimal target of the KL-regularized proposal improvement objective.\\
\(\widetilde\pi^*\) & MFA approximation of the optimal target \(\pi^*\).\\
\(\epsilon\) & Budget for the average planner alignment KL.\\
\(\beta\) & Lagrange multiplier for the planner alignment constraint.\\
\(\mathcal L_{\mathrm{APO}}(\theta,\beta)\) & Lagrangian for policy learning under the planner alignment constraint.\\
\(\eta_\beta\) & Step size of the normalized dual update for \(\beta\).\\
\([\cdot]_{\mathrm{EMA}}\) & Exponential moving average used to set the alignment budget.\\
\([\cdot]_+\) & Projection onto nonnegative values in the dual update.\\
\(\widehat J_H(\pi)\) & Expected \(H\)-step return under \(\pi\) with the fixed latent world model at \(z_t\).\\
\(\pi_{\mathrm{IL}}\) & Ideal planner imitation policy, defined by \(\pi_{\mathrm{IL}}=\widetilde\pi_P\).\\
\(A_{\max}\) & Finite bound on the absolute estimated return centered at \(\widehat J_H(\widetilde\pi_P)\).\\
\(\mathrm{KL}\) & Kullback--Leibler divergence, retaining branch labels for MFA distributions.\\

\midrule
\multicolumn{2}{@{}l}{\textbf{Evaluation}}\\*
\(T_{80}\) & Environment steps until the mean evaluation curve first reaches \(80\%\) of SWAN's final performance. All methods share a threshold for each task or suite average.\\
\(T_{\mathrm{method}},T_{\mathrm{SWAN}}\) & Wall-clock evaluation times of a compared method and SWAN, respectively.\\

\end{longtable}
\endgroup

\section{Details of k-MPPI and MFA}
\label{app:proposal_representation}
\label{app:mfa}

We provide the k-MPPI procedure and the MFA operations used
for proposal representation and policy learning.

\subsection{k-MPPI}
\label{app:kmppi}

We describe how k-MPPI constructs the empirical planner proposal.
Unlike the single Gaussian update in Eq.~\ref{eq:mppi_update},
k-MPPI maintains local Gaussians initialized from the policy
components and a global Gaussian for exploration.
After \(M\) refinement iterations, the local and global search
results form a proposal with \(K\) modes, each represented by
\(N\) elite action chunks in \(\mathcal E_k\).
The branch labels retain mode assignments for MFA projection
and policy learning.
Algorithm~\ref{alg:kmppi} gives the refinement and final
replacement procedure.

The sampling budget is shared across all branches, and batch
concatenation preserves source labels and repeated entries.
Actions obey the task bounds and masks, and search standard
deviations are clipped to \([\sigma_{\min},\sigma_{\max}]\).

In Algorithm~\ref{alg:kmppi}, \(k=1,\ldots,K\) indexes
local search branches and \(g\) denotes the global branch.
For \(b\in\{1,\ldots,K,g\}\), \(\mu_b^{(m)}\) and
\(\sigma_b^{(m)}\) are the mean and standard deviations
of the Gaussian search distribution at iteration \(m\).
\begin{algorithm}[htb]
\caption{k-MPPI}
\label{alg:kmppi}
\algrenewcommand{\algorithmicrequire}{\textbf{Input}}
\algrenewcommand{\alglinenumber}[1]{\scriptsize #1}
\begin{algorithmic}[1]
\Require \(z_t,\pi_\theta,K,M,N\), previous global mean,
 sampling allocation, and \(\sigma_{\min},\sigma_{\max}\)
\State Initialize \(\mu_k^{(1)}\) and \(\sigma_k^{(1)}\) from the
 mean and marginal standard deviations of policy component \(k\)
 \State Draw policy samples \(\mathcal S_k\) from each
 component using Eq.~\ref{eq:mfa_sampling} and reuse them
 across refinement iterations
\State Shift the previous global mean by one action and append
 zero to obtain \(\mu_g^{(1)}\), using zero at episode start,
 and set \(\sigma_g^{(1)}=\sigma_{\max}\mathbf 1\)
\For{\(m=1,\ldots,M\)}
    \State Form local batches \(\mathcal T_k^{(m)}\) from
    \(\mathcal S_k\) and fresh local Gaussian samples,
    and draw a global batch \(\mathcal T_g^{(m)}\)
    \State Concatenate the batches into \(\mathcal T^{(m)}\)
    and evaluate \(\widehat G_H\) using Eq.~\ref{eq:h_step_return}
    \State Retain \(N\) elites from each local batch as
    \(\mathcal E_k^{(m)}\) and \(N\) elites from
    \(\mathcal T^{(m)}\) as \(\mathcal E_g^{(m)}\)
    \State Update each Gaussian using Eq.~\ref{eq:mppi_update}
    with weights normalized over the corresponding elite batch
\EndFor
\State Compute final candidate weights
 \(w^{(M)}(\tau)\gets
 \operatorname{softmax}_{\mathcal T^{(M)}}
 (\widehat G_H(z_t,\tau)/\eta_{\mathrm{MPPI}})\)
\State Sum the weights by source branch
 \(p_b\gets\sum_{\tau\in\mathcal T_b^{(M)}}w^{(M)}(\tau)\)
 for \(b\in\{1,\ldots,K,g\}\)
\State Select the nearest local branch 
 \(k^\star\gets\arg\min_k
 \sum_{i=1}^{D}
 \dfrac{(\mu_{g,i}^{(M+1)}-\mu_{k,i}^{(M+1)})^2}
 {(\sigma_{k,i}^{(M+1)})^2+10^{-6}}\)
\If{\(p_g>p_{k^\star}\)}
    \State Replace \(\mathcal E_{k^\star}^{(M)}\) and
    the associated scores with the global elites and scores
\EndIf
\State Set \(\mathcal E_k=\mathcal E_k^{(M)}\) and compute
 \(w^P,\alpha^P\) using Eq.~\ref{eq:kmppi_weights}
\State Select \(\tau\) with probability \(w^P(\tau)\),
 or the highest-return chunk for greedy evaluation
\State Save \(\mu_g^{(M+1)}\) for the next warm start
\If{evaluation}
    \State \Return the first action of \(\tau\)
\EndIf
\State Construct \(\widetilde\pi_P\) using
 Appendix~\ref{app:mfa_projection} and add execution noise
 to \(\tau\) from the selected mode
\State \Return the bounded first action, retained candidates
 and scores, \(w^P\), and \(\widetilde\pi_P\)
\end{algorithmic}
\end{algorithm}

Execution adds independent Gaussian noise using the marginal
standard deviations of \(\Sigma_k^P\) for the selected mode,
clipped to \([\sigma_{\min},\sigma_{\max}]\).
Proposal statistics are computed before execution noise.

\subsection{MFA Projection}
\label{app:mfa_projection}


We construct the MFA approximations \(\widetilde\pi_P\) and
\(\widetilde\pi^*\) from weighted candidates for policy learning.
Since k-MPPI supplies branch assignments, each component is
fitted separately using weights normalized within its branch.
For nonnegative candidate weights \(\omega(\tau)\) with positive
total weight in each branch, the conditional weights are
\(\bar\omega_k(\tau)=\omega(\tau)/
\sum_{\tau'\in\mathcal E_k}\omega(\tau')\).
An ideal component fit minimizes the weighted negative
log-likelihood
\begin{equation}
\min_{\mu,U,\sigma}\quad
-\sum_{\tau\in\mathcal E_k}\bar\omega_k(\tau)
\log\mathcal N\!\left(
\tau\mid\mu,UU^\top+\operatorname{Diag}(\sigma^2)
\right),
\label{eq:app_mfa_projection}
\end{equation}
where \(U\in\mathbb R^{D\times r}\) and residual standard
deviations are bounded away from zero.
In implementation, we approximate the component distributions from
their weighted moments. 
The component parameters are combined with mode probabilities
obtained by summing the globally normalized input weights within
each branch. Weights $\omega=w^P$ therefore give
$\widetilde{\pi}_P$ with $\alpha_k^P$, whereas
$\omega(\tau)=\pi^*(\tau\mid z_t)$ gives
$\widetilde{\pi}^*$ with $\alpha_k^*$, as detailed in Appendix C.2.

\subsection{Direct Sampling}
\label{app:mfa_sampling}

For direct action-chunk generation, we select component \(k\)
with probability \(\alpha_k^\theta\).
Independent Gaussian noise \(\xi\sim\mathcal N(0,I_r)\) and
\(\zeta\sim\mathcal N(0,I_D)\) then gives
\begin{equation}
\tau=\mu_k^\theta+U_k^\theta\xi+\sigma_k^\theta\odot\zeta.
\label{eq:mfa_sampling}
\end{equation}
k-MPPI initialization applies the same expression separately
to each component.

\subsection{MFA KL Divergence}
\label{app:mfa_kl}
\label{app:apo_learning}

We compute the MFA KL used in policy learning and
Theorem~\ref{thm:apo_policy_gain} with branch labels retained.
The labels identify corresponding Gaussian components, giving
the joint distributions \(q(k,\tau)=\alpha_k^q q_k(\tau)\)
and \(p(k,\tau)=\alpha_k^p p_k(\tau)\), where \(q_k,p_k\)
are Gaussian components.
The joint KL therefore decomposes as
\begin{equation}
\mathrm{KL}(q\|p)
=
\sum_{k=1}^{K}\alpha_k^q\log\frac{\alpha_k^q}{\alpha_k^p}
+
\sum_{k=1}^{K}\alpha_k^q\mathrm{KL}(q_k\|p_k).
\label{eq:app_mfa_kl}
\end{equation}
For each pair of Gaussian components, omitting parameter subscripts,
\begin{equation}
\mathrm{KL}(q_k\|p_k)
=
\frac12\left[
\operatorname{tr}((\Sigma^p)^{-1}\Sigma^q)
+(\mu^p-\mu^q)^\top(\Sigma^p)^{-1}(\mu^p-\mu^q)
-D+\log\frac{|\Sigma^p|}{|\Sigma^q|}
\right].
\label{eq:app_gaussian_kl}
\end{equation}
The inverse and determinant are evaluated using the factor
covariance with the Woodbury and determinant identities.
Writing \(q(\tau)=\sum_k q(k,\tau)\) and
\(p(\tau)=\sum_k p(k,\tau)\), marginalization gives
\(\mathrm{KL}(q(\tau)\|p(\tau))\leq\mathrm{KL}(q\|p)\).
The joint KL therefore bounds the marginal action-chunk
divergence used in the policy improvement proof.

\section{Details of APO}
\label{app:apo}

\subsection{Algorithm}
\label{app:apo_algorithm}
SWAN follows the BMPC training loop and lazy reanalyze scheme
~\citep{wang2025bootstrapped}, retaining its
model-training objectives for \(E_\psi,F_\psi,R_\psi\) while replacing the planner and policy
with k-MPPI and the MFA action-chunk policy.
Replay buffer stores the parameters of \(\widetilde\pi_P\) and
\(\widetilde\pi^*\) alongside environment transitions, so
policy learning uses Eq.~\ref{eq:apo_loss} without repeating
planning at every update.
For value learning, a chunk is sampled from \(\pi_\theta\)
and the first action is used at each policy call in the TD rollout.
Algorithm~\ref{alg:swan} summarizes the changes to the inherited
training procedure.

\begin{algorithm}[htb]
\caption{SWAN training}
\label{alg:swan}
\algrenewcommand{\algorithmicrequire}{\textbf{Input}}
\algrenewcommand{\alglinenumber}[1]{\scriptsize #1}
\begin{algorithmic}[1]
\Require Latent world model \(E_\psi,F_\psi,R_\psi,V_\phi\),
 policy \(\pi_\theta\), replay buffer \(\mathcal B\),
 planning and learning settings
\State Initialize the networks, target value model, and
 \(\beta\geq0\)
\While{the training budget remains}
    \For{each environment interaction}
        \State Encode the state as \(z_t=E_\psi(s_t)\)
        \State Run Algorithm~\ref{alg:kmppi} to obtain the action,
        retained candidates, evaluations, and \(\widetilde\pi_P\)
        \State Construct \(\pi^*\) using
        Eq.~\ref{eq:apo_target} and project the target using
        Appendix~\ref{app:mfa_projection}
        \State Execute the action and observe \(r_t,s_{t+1}\)
        \State Store the transition and parameters of
        \(\widetilde\pi_P,\widetilde\pi^*\) in \(\mathcal B\)
    \EndFor
    \For{each learning update}
        \State Sample replay sequences with their planner
        and target parameters
        \State Update \(E_\psi,F_\psi,R_\psi\) using model-training objectives in BMPC~\citep{wang2025bootstrapped}
        \State Update the value model \(V_\phi\) with BMPC model-based TD
        learning, using the first action of a sampled MFA chunk
        at each policy call
        \State Update \(\theta\) using
        Eq.~\ref{eq:apo_loss} and the MFA KL in
        Eq.~\ref{eq:app_mfa_kl}
        \State Update the current budget \(\epsilon\) and
        multiplier \(\beta\) using Eq.~\ref{eq:apo_beta}
        \State Update the target value model following BMPC
        \If{a reanalysis update is scheduled}
            \State Run k-MPPI on the selected replay states
            and construct new targets using Eq.~\ref{eq:apo_target}
            \State Project the planner and target distributions
            and replace their stored parameters
        \EndIf
    \EndFor
\EndWhile
\end{algorithmic}
\end{algorithm}

Target construction uses the planner evaluations from the same
planning call.
The stored distribution parameters and current budget are fixed
during policy and dual updates.
The EMA coefficient and reanalysis schedule are training settings,
while the corresponding update rules remain those given in the
main text.

\subsection{Target Derivation}
\label{app:apo_derivation}
We derive the optimal target \(\pi^*\) in
Eq.~\ref{eq:apo_target}, starting with the advantage
normalization introduced in Sec.~\ref{sec:apo}.
To reduce sensitivity to reward scale and facilitate reuse
of \(\lambda\) across tasks, we divide the horizon-normalized
advantage by its weighted standard deviation under \(\pi_P\).
Since \(\pi_P\) assigns weight \(w^P(\tau)\) to each retained
entry in \(\mathcal E\) and \(\mathbb E_{\pi_P}[A]=0\),
we compute
\(
s_A=
\sqrt{
\sum_{\tau\in\mathcal E}w^P(\tau)
\left(
\frac{A(\tau)}{\sum_{h=0}^{H-1}\gamma^h}
\right)^2
},
\)
We floor \(s_A\) at \(10^{-6}\) before normalization and obtain
\(\bar A(\tau)=A(\tau)/(s_A\sum_{h=0}^{H-1}\gamma^h)\),
as defined in Sec.~\ref{sec:apo}.

Using \(\bar A\) as the improvement signal, we obtain
\(\pi^*\) by maximizing the Lagrangian in
Eq.~\ref{eq:apo_lagrangian} over \(\pi\), taking the current
planner proposal and trajectory evaluations at \(z_t\) as given.
For fixed \(\lambda>0\), the term
\(\lambda\epsilon_{\mathrm{MD}}\) is independent of \(\pi\),
so the maximization reduces to
\(\mathbb E_\pi[\bar A]
-\lambda\,\mathrm{KL}(\pi\|\pi_P)\),
which defines a mirror-descent update from \(\pi_P\).
Because finite KL excludes probability mass outside
\(\mathcal E\), we optimize the retained-candidate probabilities
subject to \(\sum_{\tau\in\mathcal E}\pi(\tau\mid z_t)=1\).
Introducing a multiplier \(\nu\) for this normalization
constraint and differentiating with respect to each
candidate probability gives
\begin{equation}
\bar A(\tau)
-\lambda\left[
\log\frac{\pi(\tau\mid z_t)}{\pi_P(\tau\mid z_t)}+1
\right]-\nu=0.
\label{eq:app_target_stationarity}
\end{equation}
Solving Eq.~\ref{eq:app_target_stationarity} for
\(\pi(\tau\mid z_t)\) gives
\(\pi(\tau\mid z_t)
=w^P(\tau)\exp(\bar A(\tau)/\lambda)\exp(-1-\nu/\lambda)\),
where \(\exp(-1-\nu/\lambda)\) is independent of \(\tau\).
Normalizing over \(\mathcal E\) therefore eliminates \(\nu\)
and recovers Eq.~\ref{eq:apo_target} as
\begin{equation}
\pi^*(\tau\mid z_t)
=
\frac{
w^P(\tau)\exp(\bar A(\tau)/\lambda)
}{
\sum_{\tau'\in\mathcal E}w^P(\tau')
\exp(\bar A(\tau')/\lambda)
},
\qquad \tau\in\mathcal E.
\label{eq:app_optimal_pi}
\end{equation}
Since the objective is strictly concave for \(\lambda>0\),
the normalized stationary solution is its unique maximizer.
Computing \(\pi^*\) only reweights the retained candidates
using their existing MPPI evaluations, so no additional
model rollouts are required.


\noindent\textbf{Induced KL radius.}
To relate the target derived above to the trust region in
Eq.~\ref{eq:apo_improvement}, define
\(\epsilon_{\mathrm{MD}}=\mathrm{KL}(\pi^*\|\pi_P)\).
For any \(\pi\) with
\(\mathrm{KL}(\pi\|\pi_P)\leq\epsilon_{\mathrm{MD}}\),
optimality of \(\pi^*\) in the regularized objective implies
\begin{equation}
\mathbb E_\pi[\bar A]-\mathbb E_{\pi^*}[\bar A]
\leq
\lambda\left[
\mathrm{KL}(\pi\|\pi_P)-\epsilon_{\mathrm{MD}}
\right]
\leq0.
\label{eq:app_induced_radius}
\end{equation}
Hence, \(\pi^*\) also solves Eq.~\ref{eq:apo_improvement}
with the induced radius.
The radius is nonnegative and becomes zero if the advantage
is constant on the retained support.
A fixed \(\lambda\) therefore specifies a regularized update
rather than a common prescribed radius across states.

\noindent\textbf{Projection of the target.}
For policy learning, we approximate the empirical target
\(\pi^*\) with an MFA distribution \(\widetilde\pi^*\).
The mixture coefficients of \(\widetilde\pi^*\) preserve
the target probability mass in each branch
\begin{equation}
\alpha_k^*
=
\sum_{\tau\in\mathcal E_k}\pi^*(\tau\mid z_t).
\label{eq:app_target_projection_weights}
\end{equation}
Within branch \(k\), the normalized target weights
\(\pi^*(\tau\mid z_t)/\alpha_k^*\) provide the weights
for the component fitting objective in
Eq.~\ref{eq:app_mfa_projection}.
We obtain the component parameters using the approximation
described in Appendix~\ref{app:mfa_projection} and combine
them with \(\alpha_k^*\) to construct \(\widetilde\pi^*\).
The MFA approximation is used for policy learning, while
Theorem~\ref{thm:apo_target_gain} concerns the exact target
\(\pi^*\).

\subsection{Proofs}
\label{app:apo_proofs}

We restate and prove the two theorems in Sec.~\ref{sec:theory}.
Throughout, the latent state \(z_t\), latent world model, and
reference distributions are fixed.
KL divergences between MFA distributions retain branch labels
as in Appendix~\ref{app:mfa_kl}, while expected returns and
\(L^1\) norms use the corresponding action-chunk marginals.
We omit the common conditioning on \(z_t\).

\apotargetgain*

\begin{proof}
By definition, \(\mathbb E_{\pi_P}[A]=0\).
Normalization by the positive scale
\(s_A\sum_{h=0}^{H-1}\gamma^h\) therefore gives
\(\mathbb E_{\pi_P}[\bar A]=0\).
Since \(\mathrm{KL}(\pi_P\|\pi_P)=0\), comparing \(\pi^*\)
with \(\pi_P\) in the KL-regularized objective yields
\begin{equation}
\mathbb E_{\pi^*}[\bar A]
\geq
\lambda\,\mathrm{KL}(\pi^*\|\pi_P)
\geq0.
\label{eq:app_normalized_gain}
\end{equation}
Restoring the original return scale gives
\[
\mathbb E_{\pi^*}[A]
=
s_A\left(\sum_{h=0}^{H-1}\gamma^h\right)
\mathbb E_{\pi^*}[\bar A]
\geq0.
\]
The definition of \(A\) then establishes
\begin{equation*}
\widehat J_H(\pi^*)-\widehat J_H(\pi_P)
=
\mathbb E_{\tau\sim\pi^*}[A(\tau)]
\geq0.
\qedhere
\end{equation*}
\end{proof}

\apopolicygain*

\begin{proof}
We bound the return improvement over
\(\pi_{\mathrm{IL}}=\widetilde\pi_P\) using the policy-fitting
terms in Eq.~\ref{eq:apo_loss}.
For fixed \(\beta\geq0\), both forward KL terms fit
\(\pi_\theta\) to their respective target distributions.
Combining their contributions to the expected negative
log-likelihood gives the normalized reference
\begin{equation}
q=\frac{\widetilde\pi^*+\beta\widetilde\pi_P}{1+\beta},
\label{eq:app_fusion_dist}
\end{equation}
where \(q\) is a joint distribution over branch labels
and action chunks.
Expanding the KL divergences yields
\begin{equation}
\mathrm{KL}(\widetilde\pi^*\|\pi_\theta)
+\beta\,\mathrm{KL}(\widetilde\pi_P\|\pi_\theta)
=
(1+\beta)\,\mathrm{KL}(q\|\pi_\theta)
+\mathrm{KL}(\widetilde\pi^*\|q)
+\beta\,\mathrm{KL}(\widetilde\pi_P\|q).
\label{eq:app_object}
\end{equation}
Since the two divergences to \(q\) are nonnegative,
the policy-fitting terms bound
\begin{equation}
\mathrm{KL}(q\|\pi_\theta)
\leq
\frac{
\mathrm{KL}(\widetilde\pi^*\|\pi_\theta)
+\beta\,\mathrm{KL}(\widetilde\pi_P\|\pi_\theta)
}{1+\beta}.
\label{eq:app_auxiliary_kl}
\end{equation}
Marginalization and Pinsker's inequality then imply
\begin{equation}
\|\pi_\theta-q\|_1
\leq
\sqrt{2\,\mathrm{KL}(q\|\pi_\theta)}
\leq
\sqrt{\frac{
2[\mathrm{KL}(\widetilde\pi^*\|\pi_\theta)
+\beta\,\mathrm{KL}(\widetilde\pi_P\|\pi_\theta)]
}{1+\beta}}.
\label{eq:app_policy_distance}
\end{equation}

To relate this distributional bound to policy improvement,
we decompose the return difference as
\begin{equation}
\widehat J_H(\pi_\theta)-\widehat J_H(\widetilde\pi_P)
=
\widehat J_H(q)-\widehat J_H(\widetilde\pi_P)
+\widehat J_H(\pi_\theta)-\widehat J_H(q).
\label{eq:app_delta_j}
\end{equation}
The reference contribution to the centered return under \(q\)
therefore vanishes, and linearity of expectation yields
\begin{equation}
\widehat J_H(q)-\widehat J_H(\widetilde\pi_P)
=
\frac{\widehat J_H(\widetilde\pi^*)
-\widehat J_H(\widetilde\pi_P)}{1+\beta}.
\label{eq:app_auxiliary_gain}
\end{equation}
It remains to bound the return difference between
\(\pi_\theta\) and \(q\).
Since their marginal densities both integrate to one,
subtracting \(\widehat J_H(\widetilde\pi_P)\) from the return
does not change this difference.
The boundedness assumption therefore gives
\begin{equation}
\begin{aligned}
\widehat J_H(\pi_\theta)-\widehat J_H(q)
&=
\int
\bigl(\pi_\theta(\tau)-q(\tau)\bigr)
\widehat G_H(z_t,\tau)\,d\tau
\\
&=
\int
\bigl(\pi_\theta(\tau)-q(\tau)\bigr)
\bigl(\widehat G_H(z_t,\tau)
-\widehat J_H(\widetilde\pi_P)\bigr)\,d\tau
\\
&\geq
-A_{\max}
\int\bigl|\pi_\theta(\tau)-q(\tau)\bigr|\,d\tau
\\
&=
-A_{\max}\|\pi_\theta-q\|_1.
\end{aligned}
\label{eq:app_penelty}
\end{equation}
Adding
\(\widehat J_H(q)-\widehat J_H(\widetilde\pi_P)\)
to both sides yields
\begin{equation}
\widehat J_H(\pi_\theta)-\widehat J_H(\widetilde\pi_P)
\geq
\widehat J_H(q)-\widehat J_H(\widetilde\pi_P)
-A_{\max}\|\pi_\theta-q\|_1.
\label{eq:app_policy_score}
\end{equation}
Substituting Eqs.~\ref{eq:app_policy_distance}
and~\ref{eq:app_auxiliary_gain}, and using
\(\pi_{\mathrm{IL}}=\widetilde\pi_P\), gives
\begin{equation*}
\widehat J_H(\pi_\theta)-\widehat J_H(\pi_{\mathrm{IL}})
\geq
\frac{
\widehat J_H(\widetilde\pi^*)
-\widehat J_H(\widetilde\pi_P)
}{1+\beta}
-
A_{\max}
\sqrt{\frac{
2\left[
\mathrm{KL}(\widetilde\pi^*\|\pi_\theta)
+\beta\,\mathrm{KL}(\widetilde\pi_P\|\pi_\theta)
\right]
}{1+\beta}}.
\qedhere
\end{equation*}
\end{proof}

\subsection{Unified View}
\label{app:unified_view}

In this section, we derive the policy learning objectives of prior methods and
APO from a common distribution improvement problem.
The derivation fixes the reference and evaluation signal for each
update.
Outer state expectations and auxiliary regularization are omitted
unless needed to specify the learning distribution.

\noindent\textbf{Distribution improvement.}
Let \(x\) denote an action or action chunk, \(\rho\) a reference
distribution, and \(f(x)\) a value or advantage estimate.
Starting from \(\max_\pi\mathbb E_\pi[f]\), a KL constraint
limits departure from \(\rho\).
The Lagrangian in Eq.~\ref{eq:apo_lagrangian} gives the
regularized objective and optimal distribution
\begin{equation}
q^*
=
\arg\max_\pi\left\{
\mathbb E_\pi[f]-\lambda\,\mathrm{KL}(\pi\|\rho)
\right\},
\qquad
q^*(x)=\frac{\rho(x)e^{f(x)/\lambda}}{Z},
\label{eq:app_unified_target}
\end{equation}
where \(\lambda>0\) and
\(Z=\mathbb E_\rho[e^{f/\lambda}]<\infty\).
The reference determines where improvement starts, and the signal
changes the relative probability of the action chunks.

\noindent\textbf{Direct policy optimization.}
Substituting Eq.~\ref{eq:app_unified_target} into KL gives
\begin{equation}
\mathbb E_\pi[f]-\lambda\,\mathrm{KL}(\pi\|\rho)
=
\lambda\log Z-\lambda\,\mathrm{KL}(\pi\|q^*).
\label{eq:app_unified_reverse}
\end{equation}
Since \(\log Z\) is fixed, restricting \(\pi\) to a
parameterized policy family recovers minimization of reverse KL
toward \(q^*\).
Two choices recover the critic-guided objectives below.

For TD-MPC2~\citep{hansen2024td}, let \(x=a\),
\(f(a)=Q_\phi(z_t,a)\), and let \(\rho\) be uniform on the
bounded action space \(\mathcal A\).
The identity
\(\mathrm{KL}(\pi\|\rho)=-\mathcal H(\pi)+\log|\mathcal A|\)
then gives the maximum-entropy policy objective
\begin{equation}
\max_{\theta_a}\quad
\mathbb E_{a\sim\pi_{\theta_a}}[Q_\phi(z_t,a)]
+\lambda\,\mathcal H(\pi_{\theta_a}),
\label{eq:app_unified_tdmpc}
\end{equation}
where \(\mathcal H\) is differential entropy.
Value scaling is absorbed into the relative coefficient.

For PO-MPC~\citep{serra-gomez2026a}, replace the uniform reference
with the adaptive prior \(\pi_p\) learned from the planner's
first-action distribution and set \(f=Q_{\mathrm{reg}}\),
the regularized action-value function.
Equation~\ref{eq:app_unified_reverse} then recovers
\begin{equation}
\max_{\theta_a}\quad
\mathbb E_{a\sim\pi_{\theta_a}}[Q_{\mathrm{reg}}(z_t,a)]
-\lambda\,\mathrm{KL}(\pi_{\theta_a}\|\pi_p).
\label{eq:app_unified_pompc}
\end{equation}
Both updates optimize an action policy, with planner information
entering through the reference in Eq.~\ref{eq:app_unified_pompc}.

\noindent\textbf{Policy learning from the improved target.}
A second route constructs \(q^*\) first and learns a policy by
minimizing \(\mathrm{KL}(q^*\|\pi)\).
For compatible densities with finite terms, the target entropy
is constant during the policy update, giving
\begin{equation}
\arg\min_\pi\mathrm{KL}(q^*\|\pi)
=
\arg\min_\pi
\left\{-\mathbb E_{x\sim\rho}
\left[\frac{e^{f(x)/\lambda}}{Z}\log\pi(x)\right]\right\}.
\label{eq:app_unified_forward}
\end{equation}
The forward and reverse objectives use the same improved
distribution but define different updates in a restricted
policy family.

BMPC~\citep{wang2025bootstrapped} follows by setting
\(\rho=\pi_{P,0}^{\mathrm G}\) and \(f=0\).
Then \(q^*=\rho\), and Eq.~\ref{eq:app_unified_forward} becomes
\begin{equation}
\min_{\theta_a}\quad
\mathrm{KL}(\pi_{P,0}^{\mathrm G}\|\pi_{\theta_a}).
\label{eq:app_unified_bmpc}
\end{equation}
The resulting objective is the planner imitation described in
Sec.~\ref{sec:prelim}.
The zero signal denotes no additional proposal improvement.

For BOOM~\citep{zhan2026bootstrap}, apply
Eq.~\ref{eq:app_unified_target} to a uniform reference over a
replay batch \(\mathcal C\) of latent-state and action pairs
\((z,a)\), using \(f(z,a)=Q_\phi(z,a)\).
The resulting entry weights yield the alignment objective
\begin{equation}
\omega(z,a)=\operatorname{softmax}_{\mathcal C}
\!\left(Q_\phi(z,a)/\lambda\right),
\qquad
\min_{\theta_a}\quad
-\sum_{(z,a)\in\mathcal C}
\omega(z,a)\log\pi_{\theta_a}(a\mid z).
\label{eq:app_unified_boom}
\end{equation}
BOOM combines the weighted alignment term with a separate
critic-guided objective.
The weights are normalized across replay entries rather than
a complete proposal at one state.

\noindent\textbf{APO.}
Setting \(x=\tau\), \(\rho=\pi_P\), and \(f=\bar A\) in
Eq.~\ref{eq:app_unified_target} gives the optimal target
\(q^*=\pi^*\) from Eq.~\ref{eq:apo_target}.
The reference is the current complete planner proposal, and the
signal comes from the same MPPI trajectory evaluations.
MFA projection gives the approximation \(\widetilde\pi^*\)
for policy learning, while the separate alignment constraint
preserves alignment with \(\widetilde\pi_P\).
For fixed \(\beta\), Eq.~\ref{eq:apo_loss} therefore reduces
to
\begin{equation}
\min_\theta\quad
\mathrm{KL}(\widetilde\pi^*\|\pi_\theta)
+\beta\,\mathrm{KL}(\widetilde\pi_P\|\pi_\theta),
\label{eq:app_unified_apo}
\end{equation}
after omitting the constant budget term.

With no additional advantage improvement, \(f=0\) gives
\(\pi^*=\pi_P\) and
\(\widetilde\pi^*=\widetilde\pi_P\).
Equation~\ref{eq:app_unified_apo} becomes
\begin{equation}
\min_\theta\quad
(1+\beta)\,\mathrm{KL}(\widetilde\pi_P\|\pi_\theta),
\label{eq:app_unified_imitation}
\end{equation}
which has the same minimizers as planner imitation of the complete proposal.
APO thus incorporates additional proposal improvement into
policy learning while preserving planner alignment.
Theorems~\ref{thm:apo_target_gain} and
\ref{thm:apo_policy_gain} establish nonnegative gain for the exact target and a sufficient condition for improvement over ideal
planner imitation after projection and policy learning.
The common formulation separates the choices of reference,
evaluation signal, and policy update, providing a basis for
further developments in learning from planning.

\newpage
\section{Benchmark Task Details and Visualizations}
\label{app:task_details}
\raggedbottom

\begin{figure}[H]
\centering
\includegraphics[width=\linewidth]{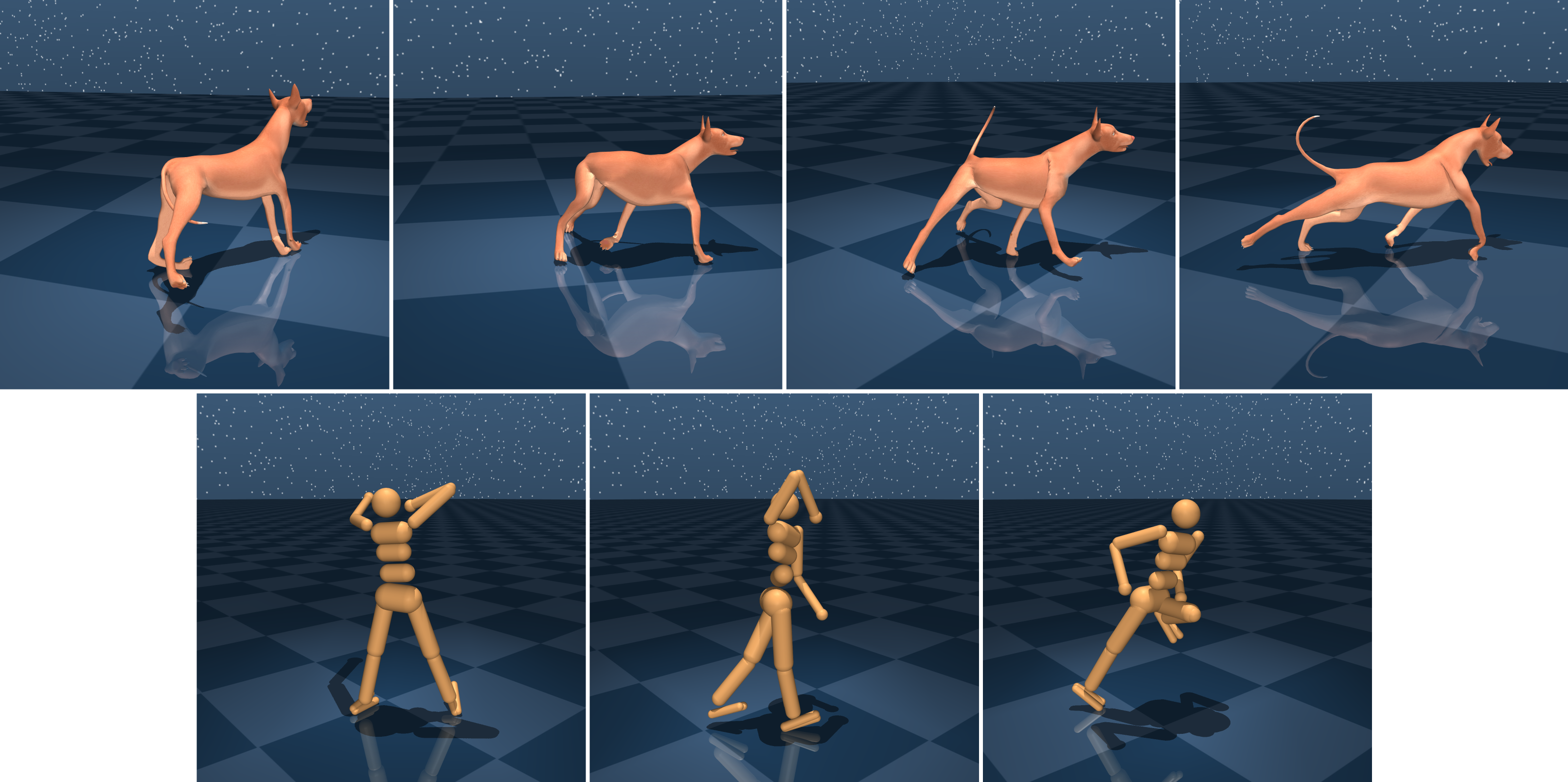}
\caption{\textbf{DMControl task visualization.}
The seven Dog and Humanoid locomotion tasks used in our experiments span standing, walking, trotting, and running.}
\label{fig:task_vis_dmc}
\end{figure}

\begin{figure}[H]
\centering
\includegraphics[width=\linewidth]{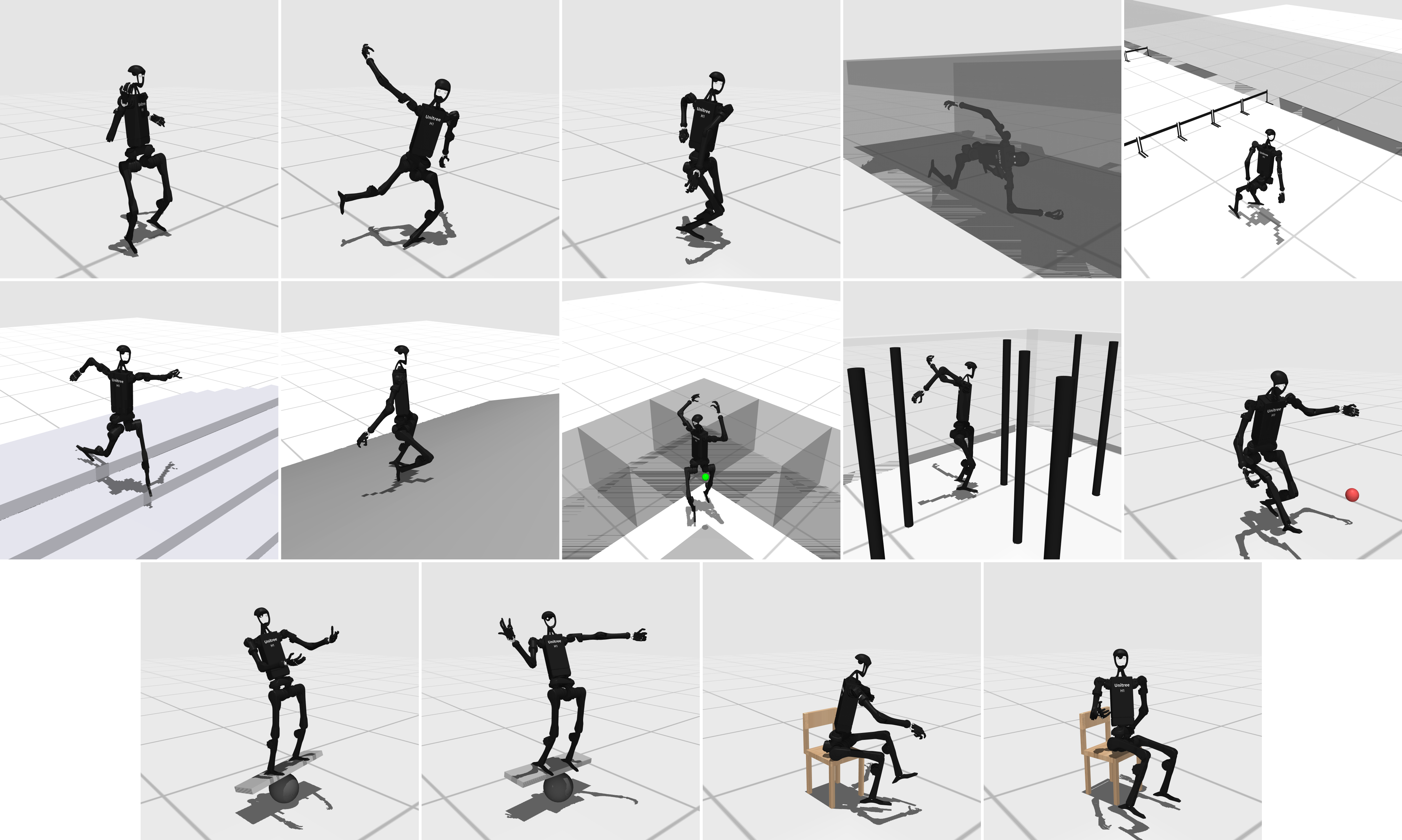}
\caption{\textbf{HumanoidBench task visualization.}
The 14 whole-body control tasks used in our experiments cover locomotion, obstacle traversal, navigation,
balancing, sitting, and reaching.}
\label{fig:task_vis_humanoidbench}
\end{figure}

\begin{figure}[H]

\centering
\includegraphics[width=\linewidth]{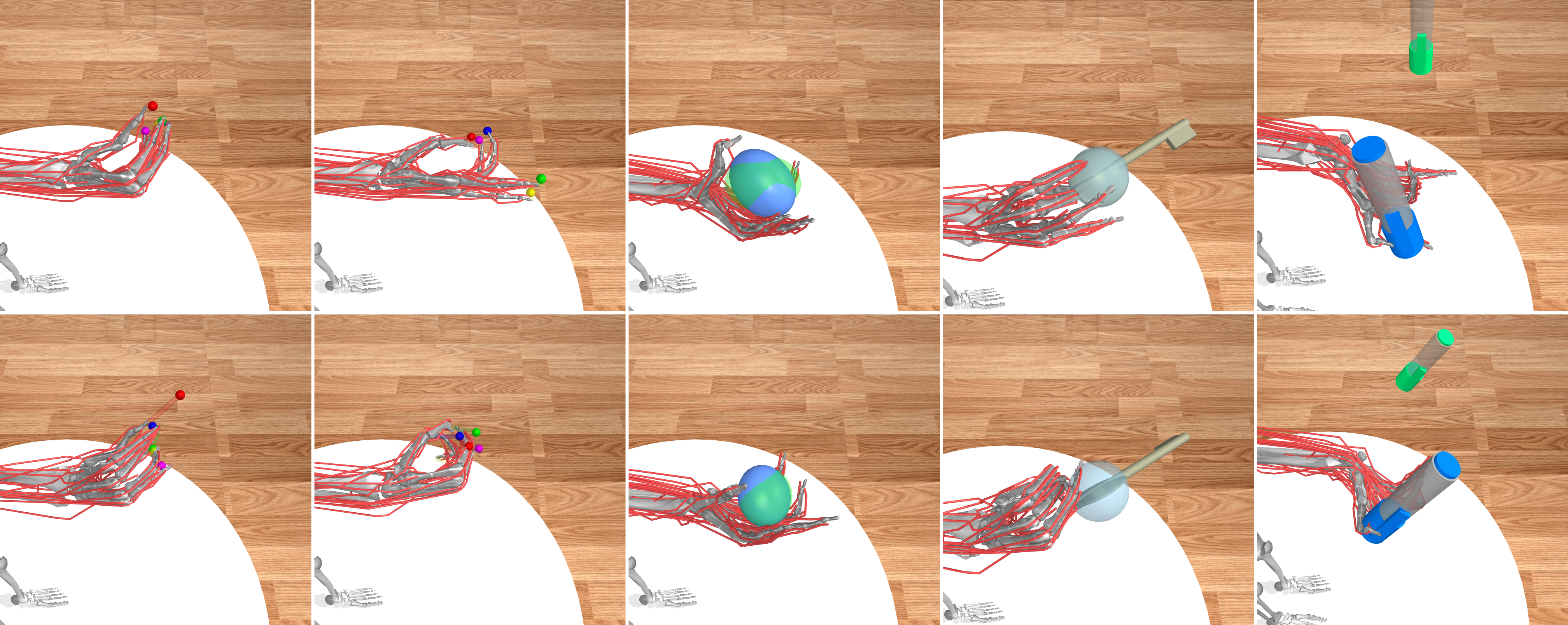}
\caption{\textbf{MyoSuite task visualization.} Ten musculoskeletal hand-control tasks spanning
reaching, posing, object holding, key turning, and pen twirling, each with standard
and harder randomized variants.}
\label{fig:task_vis_myo}
\end{figure}

\begin{figure}[H]
\centering
\includegraphics[width=\linewidth]{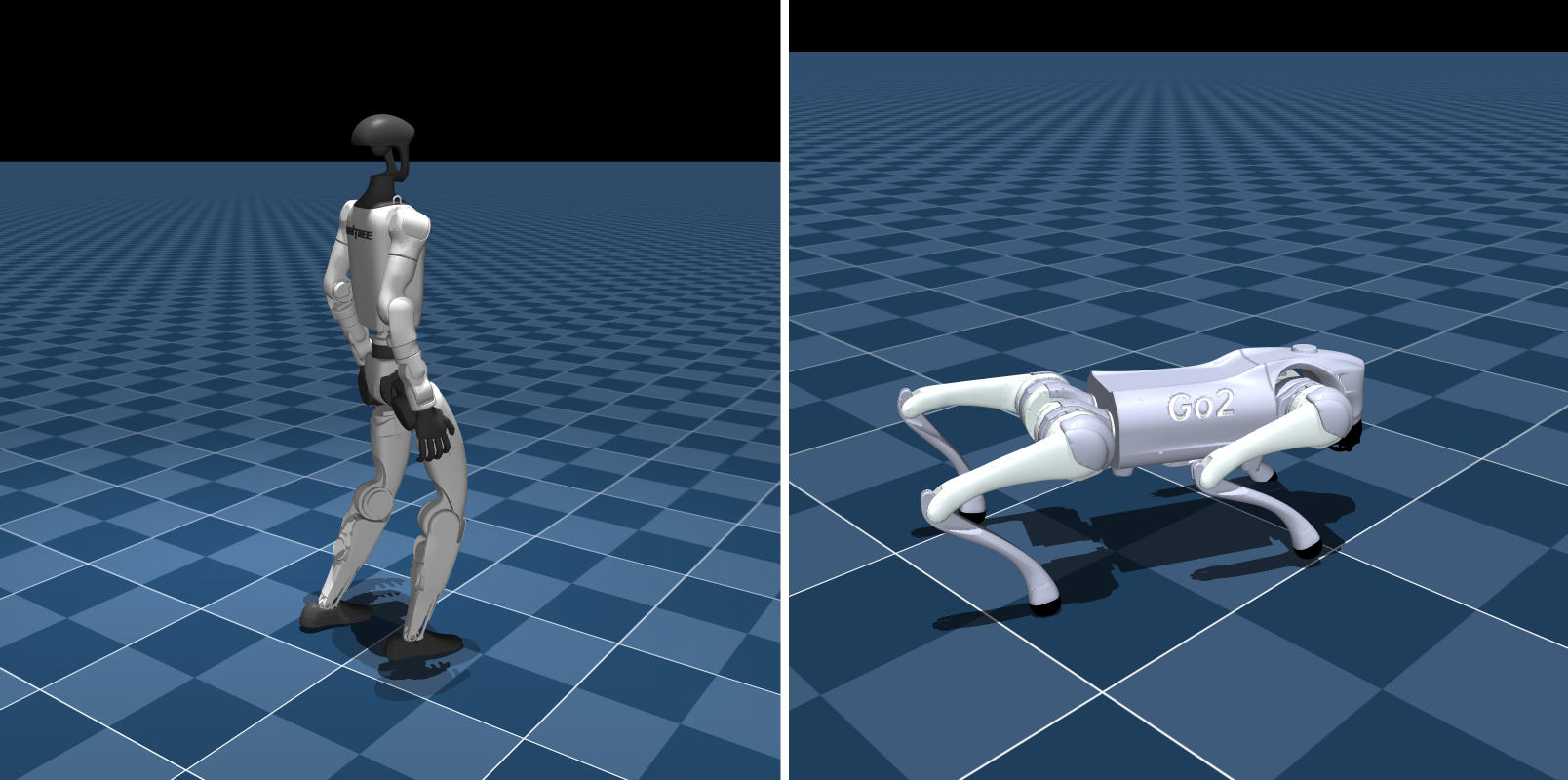}
\caption{\textbf{Unitree locomotion visualization.}
Unitree G1 and Go2 velocity-tracking tasks.}
\label{fig:task_vis_unitree}
\end{figure}

We report details for the 31 benchmark tasks from DMControl,
HumanoidBench, and MyoSuite, together with the two Unitree locomotion
tasks used in our robot experiments. Table~\ref{tab:task_details}
summarizes the task dimensions and objectives, and
Figs.~\ref{fig:task_vis_dmc}--\ref{fig:task_vis_unitree}
show representative environments.


\begin{table*}[h]
\centering
\small
\setlength{\tabcolsep}{4pt}
\renewcommand{\arraystretch}{1.05}
\caption{Task details used in this paper. The state and action dimensions and task descriptions are listed below.}
\label{tab:task_details}
\begin{tabularx}{\textwidth}{
@{}
>{\raggedright\arraybackslash}p{0.39\textwidth}
cc
>{\raggedright\arraybackslash}X
@{}}
\toprule
\textbf{Task} & \textbf{State} & \textbf{Action} & \textbf{Task descriptions} \\
\midrule

\multicolumn{4}{@{}l}{\textbf{DMControl~\citep{tunyasuvunakool2020}}} \\
dog-stand       & 223 & 38 & Stand stably. \\
dog-walk        & 223 & 38 & Walk forward. \\
dog-trot        & 223 & 38 & Trot forward. \\
dog-run         & 223 & 38 & Run forward. \\
humanoid-stand  & 67  & 21 & Stand upright. \\
humanoid-walk   & 67  & 21 & Walk forward. \\
humanoid-run    & 67  & 21 & Run forward. \\

\addlinespace[2pt]
\multicolumn{4}{@{}l}{\textbf{HumanoidBench~\citep{sferrazza2024humanoidbench}}} \\
humanoid\_h1hand-stand-v0
    & 151 & 61 & Stand stably. \\
humanoid\_h1hand-walk-v0
    & 151 & 61 & Track walking velocity. \\
humanoid\_h1hand-run-v0
    & 151 & 61 & Track running velocity. \\
humanoid\_h1hand-crawl-v0
    & 151 & 61 & Crawl through a tunnel. \\
humanoid\_h1hand-hurdle-v0
    & 151 & 61 & Cross hurdles. \\
humanoid\_h1hand-stair-v0
    & 151 & 61 & Traverse stairs. \\
humanoid\_h1hand-slide-v0
    & 151 & 61 & Traverse inclined slides. \\
humanoid\_h1hand-pole-v0
    & 151 & 61 & Avoid poles while moving forward. \\
humanoid\_h1hand-sit\_simple-v0
    & 151 & 61 & Sit on a fixed chair. \\
humanoid\_h1hand-sit\_hard-v0
    & 164 & 61 & Sit under randomized configurations. \\
humanoid\_h1hand-balance\_simple-v0
    & 164 & 61 & Balance on a fixed-pivot board. \\
humanoid\_h1hand-balance\_hard-v0
    & 177 & 61 & Balance on a moving-pivot board. \\
humanoid\_h1hand-maze-v0
    & 151 & 61 & Navigate through a maze. \\
humanoid\_h1hand-reach-v0
    & 157 & 61 & Reach a 3-D target. \\

\addlinespace[2pt]
\multicolumn{4}{@{}l}{\textbf{MyoSuite~\citep{caggiano2022myosuite}}} \\
myo-reach
    & 115 & 39 & Reach fixed fingertip targets. \\
myo-reach-hard
    & 115 & 39 & Reach randomized fingertip targets. \\
myo-pose
    & 108 & 39 & Match a fixed hand pose. \\
myo-pose-hard
    & 108 & 39 & Match randomized hand poses. \\
myo-obj-hold
    & 91 & 39 & Hold an object. \\
myo-obj-hold-hard
    & 91 & 39 & Hold an object from randomized states. \\
myo-key-turn
    & 93 & 39 & Turn a key to a fixed target. \\
myo-key-turn-hard
    & 93 & 39 & Turn a key to randomized targets. \\
myo-pen-twirl
    & 83 & 39 & Reorient a pen to a fixed target. \\
myo-pen-twirl-hard
    & 83 & 39 & Reorient a pen to randomized targets. \\

\addlinespace[2pt]
\multicolumn{4}{@{}l}{\textbf{Unitree RL MjLab~\citep{zakka2026mjlab}}} \\
Unitree-G1-Flat
    & 98 & 29 & Track commanded planar velocity. \\
Unitree-Go2-Flat
    & 47 & 12 & Track commanded planar velocity. \\

\bottomrule
\end{tabularx}
\end{table*}


\section{Detailed Experimental Settings}
\label{sec:appendix_training_setting}

This section provides additional experimental details for the results
reported in Sec.~\ref{sec:experiment}.
We first describe the common \textbf{training and evaluation protocol},
followed by the \textbf{implementation settings of the baselines}.
We then provide the \textbf{hyperparameters} specific to SWAN and finally describe the \textbf{benchmark rewards}.


\subsection{Training and Evaluation Protocol}
\label{app:training_evaluation}

\paragraph{Training details.}
All benchmark experiments are conducted in an online reinforcement
learning setting, where agents continuously collect environment
transitions during training.
For fair comparison, we match the total number of environment steps
across methods on each task.
We define environment steps as the number of agent decision steps
multiplied by the environment action repeat.
The action repeat is \(2\) for DMControl and \(1\) for
HumanoidBench, MyoSuite, Unitree G1, and Unitree Go2.
We train agents for \(4\)M environment steps on DMControl,
\(2\)M steps on HumanoidBench and MyoSuite,
\(2\)M steps on Unitree G1, and \(1\)M steps on Unitree Go2.
All simulation experiments are conducted with
three independent random seeds on single NVIDIA RTX A6000 GPU.

For MPPI-based methods, we consider two planning-budget settings.
In the \textbf{full-budget} comparison, BMPC, BOOM, PO-MPC, and
BMPC-DP use \(M=8\) MPPI refinement iterations, whereas SWAN uses
\(M=4\).
In the \textbf{half-budget} comparison, all MPPI-based methods use
\(M=4\).
The number of candidate trajectories per refinement iteration is
kept unchanged when varying \(M\), such that reducing the number of
refinement iterations directly
reduces the number of model-rollout evaluations during planning.

\paragraph{Evaluation protocol.}
We evaluate the current controller every \(50{,}000\) agent
steps, corresponding to \(100{,}000\) environment steps
on DMControl, and using \(10\) evaluation episodes per task. MPPI-based methods retain online planning during evaluation using the refinement budget associated with the corresponding experimental setting. For DMControl, HumanoidBench, Unitree Go2, and Unitree G1, we report the mean episode return, whereas for MyoSuite we report the task success rate.


\subsection{Baseline Implementation Details}
\label{app:baseline_implementation}

We compare SWAN with the planning-based MBRL methods
BMPC~\citep{wang2025bootstrapped}, BOOM~\citep{zhan2026bootstrap}, and PO-MPC~\citep{serra-gomez2026a}, the imagination-based MBRL method
DreamerV3~\citep{hafner2023mastering}, and the model-free methods
FPO~\citep{mcallister2026flow} and SAC~\citep{haarnoja2018soft}.
For BMPC, BOOM, and PO-MPC, we follow algorithm-specific 
hyperparameters reported in their original implementations,
and use the same model architectures and optimization settings 
across these methods for fair comparison. 
We detail specific and common hyperparameters of these planning-based 
methods in~\ref{app:swan_hyperparameters}.
The only modification in the half-budget setting is reducing the number of MPPI refinement iterations to \(M=4\), while all other settings are kept unchanged. For DreamerV3, FPO, and SAC, we use their default configurations for all tasks across all benchmark suites and match the environment-interaction budget to that of the other methods.

We additionally construct BMPC-DP by replacing BMPC's one-step
policy with a diffusion policy that generates complete \(H\)-step
action chunks, while retaining the original MPPI planner and
planner-imitation objective. The diffusion policy is parameterized by a conditional 1D residual CNN operating over the action sequence, with four residual blocks, a hidden width of \(512\), and kernel size \(3\).
During training, we use a \texttt{DDPMScheduler} with \(1{,}000\) diffusion timesteps. During inference, action chunks are generated using a
\texttt{DDIMScheduler} with \(20\) denoising steps.


\subsection{Hyperparameter Settings}
\label{app:swan_hyperparameters}
In this section, we detail specific and common hyperparameters of all
planning-based methods. As shown in Tab~\ref{tab:shared_hyperparameters}, we summarize
the shared architecture and optimization settings for BMPC, BOOM,
PO-MPC, and SWAN.
Since the methods differ in their planners, policies,
and value models, we list their specific settings separately in 
Tab.~\ref{tab:method_hyperparameters}.

In the full- and half-budget comparisons, the MPPI-based
baselines use \(M=8\) and \(M=4\), respectively, while
SWAN uses \(M=4\) in both.
\subsection{Hyperparameter Selection for SWAN}
\label{app:hyperparameter_selection}
\begin{figure}[htbp]
    \centering
    \includegraphics[width=\linewidth]
    {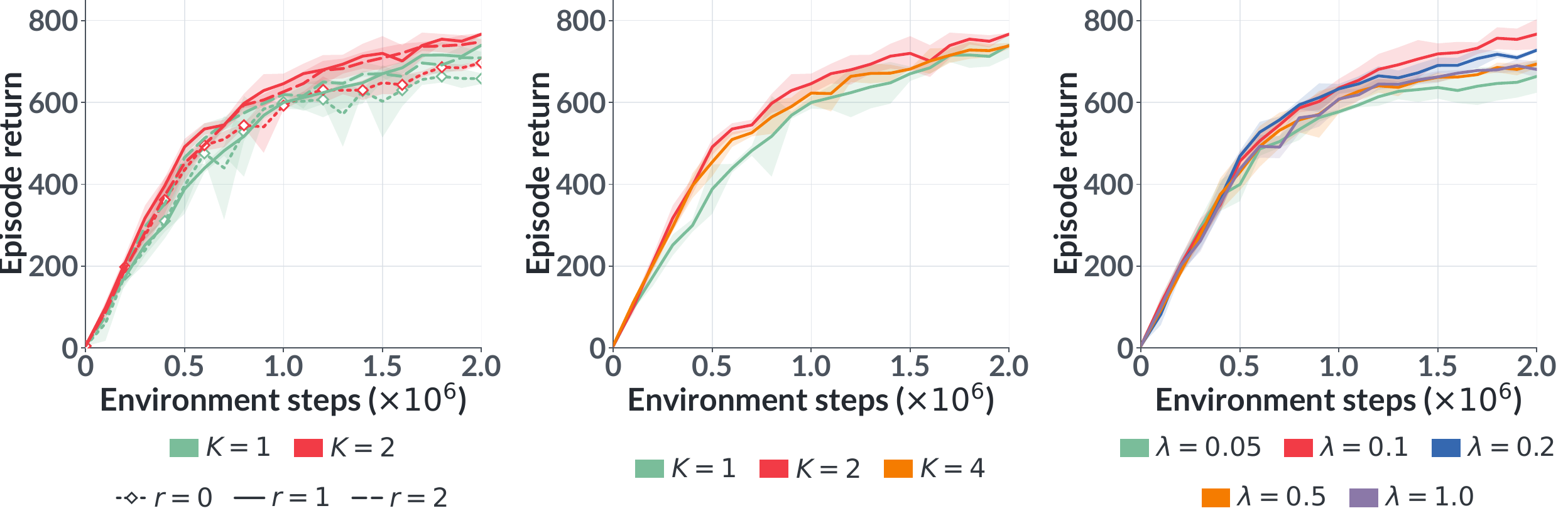}
    \caption{
    \textbf{Hyperparameter selection on \textit{Dog-Run}.}
    \textbf{Left.} Comparison of \(r\in\{0,1,2\}\) for
    \(K\in\{1,2\}\).
    \textbf{Middle.} Comparison of \(K\in\{1,2,4\}\) with \(r=1\).
    \textbf{Right.} Comparison of APO target temperatures
    \(\lambda\in\{0.05,0.1,0.2,0.5,1.0\}\).
    All panels show episode return over \(2\)M environment steps.
    }
    \label{fig:dog_run_parameter_selection}
\end{figure}

Further, we show the hyperparameter selection of proposal modes \(K\), 
the MFA rank parameter \(r\), and the APO target temperature \(\lambda\)
through comparisons on \textit{Dog-Run}.
As shown in Fig.~\ref{fig:dog_run_parameter_selection} (left),
based on BMPC-MFA, increasing \(r\) from \(0\) to \(1\) provides consistent significant return gain for both \(K=1\) and \(K=2\), supporting the importance of modeling correlations across time and action dimensions. However,
increasing \(r\) from \(1\) to \(2\) provides no consistent
return gain for either \(K=1\) or \(K=2\), supporting
\(r=1\) with fewer covariance parameters.
The middle panel further shows that \(K=2\) improves returns
over \(K=1\), while \(K=4\) provides no further gain.
For APO, \(\lambda=0.1\) achieves the highest final return
among the five tested values in the right panel.
We therefore use \(K=2\), \(r=1\), and \(\lambda=0.1\)
as the default SWAN settings.
\begin{table}[htbp]
    \centering
    \small
    \setlength{\tabcolsep}{6pt}
    \renewcommand{\arraystretch}{1.00}
    \caption{Planning and method-specific settings for the full- and half-budget benchmark comparisons.}
    \label{tab:method_hyperparameters}
    \begin{tabularx}{\linewidth}{@{}X r@{}}
        \toprule
        \textbf{Hyperparameter} & \textbf{Value} \\
        \midrule
        \multicolumn{2}{@{}l}{\textbf{Shared planning settings}} \\
        Planning horizon \(H\) & \(3\) \\
        Total candidates per refinement iteration & \(512\) \\
        MPPI temperature \(\eta_{\mathrm{MPPI}}\) & \(2\) \\
        Search standard deviation bounds \(\sigma^{(m)}\) & \([0.05,\ 2]\) \\
        \midrule
        \multicolumn{2}{@{}l}{\textbf{BMPC, BOOM, and PO-MPC}} \\
        Refinement iterations \(M\) in the full-budget comparison & \(8\) \\
        Refinement iterations \(M\) in the half-budget comparison & \(4\) \\
        Policy trajectories & \(24\) \\
        Elite count & \(64\) \\
        One-step policy \(\pi_{\theta_a}\) hidden layers / width & \(2\times512\) \\
        Entropy coefficient & \(10^{-4}\) \\
        \midrule
        \multicolumn{2}{@{}l}{\textbf{BMPC}} \\
        Policy log standard deviation bounds & \([-3,\ 1]\) \\
        \midrule
        \multicolumn{2}{@{}l}{\textbf{BOOM}} \\
        Number of \(Q\) networks & \(5\) \\
        Policy log standard deviation bounds & \([-10,\ 2]\) \\
        Alignment coefficient / \(d_a\) for \(24\leq d_a\leq48\) & \(0.001\) \\
        Alignment coefficient / \(d_a\) otherwise & \(0.02\) \\
        \midrule
        \multicolumn{2}{@{}l}{\textbf{PO-MPC}} \\
        Number of \(Q\) networks / \(Q^\lambda\) networks & \(5\,/\,5\) \\
        Policy and prior log standard deviation bounds & \([-2,\ 2]\) \\
        Learned prior hidden layers / width & \(2\times512\) \\
        KL-regularized value loss coefficient & \(0.1\) \\
        Policy / value KL coefficient & \(1\) \\
        Prior learning loss & Reverse KL \\
        \midrule
        \multicolumn{2}{@{}l}{\textbf{Value models for BMPC and SWAN}} \\
        Number of value models \(V_\phi\) & \(2\) \\
        \midrule
        \multicolumn{2}{@{}l}{\textbf{Reanalyze for BMPC, PO-MPC, and SWAN}} \\
        Interval / batch size / horizon & \(10\,/\,20\,/\,3\) \\
        \midrule
        \multicolumn{2}{@{}l}{\textbf{SWAN k-MPPI}} \\
        Refinement iterations \(M\) in both comparisons & \(4\) \\
        Local modes \(K\) & \(2\) \\
        Policy samples per mode & \(24\) \\
        Exploration samples per mode & \(104\) \\
        Global proposal samples & \(256\) \\
        Elites per mode \(N\) & \(64\) \\
        \midrule
        \multicolumn{2}{@{}l}{\textbf{SWAN MFA action-chunk policy \(\pi_\theta\)}} \\
        Rank parameter \(r\) & \(1\) \\
        Mode embedding dimension \(\dim(e_k)\) & \(16\) \\
        Hidden layers / width & \(2\times512\) \\
        Marginal log standard deviation bounds & \([-3,\ 1]\) \\
        \midrule
        \multicolumn{2}{@{}l}{\textbf{SWAN APO}} \\
        Target temperature \(\lambda\) & \(0.1\) \\
        Initial alignment multiplier \(\beta_0\) & \(1.0\) \\
        Dual step size \(\eta_\beta\) & \(0.001\) \\
        Alignment-budget EMA coefficient & \(0.001\) \\
        Entropy coefficient & \(0\) \\
        \bottomrule
    \end{tabularx}
\end{table}
\begin{table}[htbp]
    \centering
    \small
    \setlength{\tabcolsep}{6pt}
    \renewcommand{\arraystretch}{1.10}
    \caption{Shared architecture and optimization settings for BMPC, BOOM, PO-MPC, and SWAN.}
    \label{tab:shared_hyperparameters}
    \begin{tabularx}{\linewidth}{@{}X r@{}}
        \toprule
        \textbf{Hyperparameter} & \textbf{Value} \\
        \midrule
        \multicolumn{2}{@{}l}{\textbf{General}} \\
        Training steps & \(2{,}000{,}000\) \\
        Replay buffer capacity & \(1{,}000{,}000\) \\
        Random seeds & \(1,\ 10,\ 100\) \\
        \midrule
        \multicolumn{2}{@{}l}{\textbf{Network architecture}} \\
        Latent dimension \(d_z\) & \(512\) \\
        SimNorm group size & \(8\) \\
        Number of bins & \(101\) \\
        Symlog minimum / maximum & \(-10\,/\,10\) \\
        Encoder \(E_\psi\) width & \(256\) \\
        Number of encoder layers & \(2\) \\
        Dynamics model \(F_\psi\) hidden layers / width & \(2\times512\) \\
        Reward model \(R_\psi\) hidden layers / width & \(2\times512\) \\
        Critic hidden layers / width & \(2\times512\) \\
        \midrule
        \multicolumn{2}{@{}l}{\textbf{Optimization}} \\
        Learning rate & \(3\times10^{-4}\) \\
        Encoder learning rate & \(9\times10^{-5}\) \\
        Batch size & \(256\) \\
        Maximum gradient norm & \(20\) \\
        Adam betas & \((0.9,\ 0.999)\) \\
        Adam epsilon for world model / policy & \(10^{-8}\,/\,10^{-5}\) \\
        Target-network EMA coefficient & \(0.01\) \\
        \midrule
        \multicolumn{2}{@{}l}{\textbf{Model learning}} \\
        Discount \(\gamma\) for DMControl / HumanoidBench / MyoSuite & \(0.99\,/\,0.995\,/\,0.95\) \\
        Model time discount \(\rho\) & \(0.5\) \\
        Consistency coefficient & \(20\) \\
        Reward loss coefficient & \(0.1\) \\
        Value loss coefficient & \(0.1\) \\
        \bottomrule
    \end{tabularx}
\end{table}


\subsection{Benchmark Rewards}
\label{app:reward_details}
For each task, all methods use the same reward definition to ensure a fair comparison. Unless otherwise specified below, we use the reward functions provided by the corresponding benchmark environments.

\paragraph{DMControl, HumanoidBench, and MyoSuite.}
For DMControl~\citep{tunyasuvunakool2020},
HumanoidBench~\citep{sferrazza2024humanoidbench}, and
MyoSuite~\citep{caggiano2022myosuite}, we use the reward functions provided by the original benchmark environments.

\paragraph{Unitree Go2.}
The Go2 velocity-tracking task is based on the locomotion reward of
DMO~\citep{amigo2025first}, reimplemented in MJLab.
For the simulation experiments, we retain the original DMO reward
terms and coefficients without modification.
For the refinement stage used prior to real-world deployment, we
modify only two reward components to improve tracking precision and
standing behavior.

First, we refine the linear- and yaw-velocity tracking rewards using
mixtures of wide and narrow exponential kernels:
\begin{align}
r_t^{\mathrm{lin}}
={}&
0.5\exp\!\left(
-\frac{\|\mathbf c_t^{xy}-\mathbf v_t^{xy}\|_2^2}{0.5^2}
\right)
+
0.5\exp\!\left(
-\frac{\|\mathbf c_t^{xy}-\mathbf v_t^{xy}\|_2^2}{0.15^2}
\right),
\\
r_t^{\mathrm{yaw}}
={}&
0.75\exp\!\left(
-\frac{(c_t^\omega-\omega_t^z)^2}{0.5^2}
\right)
+
0.25\exp\!\left(
-\frac{(c_t^\omega-\omega_t^z)^2}{0.10^2}
\right),
\end{align}
where \(\mathbf c_t^{xy}\) and \(c_t^\omega\) denote the commanded
planar and yaw velocities, and
\(\mathbf v_t^{xy}\) and \(\omega_t^z\) are the corresponding base
velocities.
The narrower kernels provide stronger incentives for precise
velocity tracking.

Second, for near-zero commands,
\(\|\mathbf c_t\|_2<0.1\), we disable the swing-contact and
swing-clearance penalties and instead encourage the nominal standing
pose using
\begin{equation}
\ell_t^{\mathrm{stand}}
=
\|\mathbf q_t-\mathbf q_0\|_2^2,
\end{equation}
where \(\mathbf q_t\) is the current joint configuration and
\(\mathbf q_0\) is the nominal standing pose.
All other DMO reward terms and coefficients remain unchanged during
this refinement stage.

\paragraph{Unitree G1.}
The G1 task is based on the official
\texttt{Unitree-G1-Flat} environment in
MJLab~\citep{zakka2026mjlab}.
Unless specified below, we retain its original reward terms,
coefficients, and control-timestep scaling.

We replace the native foot-gait reward with our defined
gait reward,
\begin{equation}
r_t^{\mathrm{gait}}
=
-0.8 H_t
-0.8 H_t^{\mathrm{sym}}
+0.2 C_t
+0.5 P_t
+0.5 G_t ,
\label{eq:g1_gait_reward}
\end{equation}
where \(H_t\) penalizes insufficient swing-foot clearance,
\(H_t^{\mathrm{sym}}\) penalizes left--right asymmetry in swing
clearance, \(C_t\) rewards transferring contact load from the
swing foot to the stance foot, \(P_t\) rewards airborne foot
progress along the commanded motion, and \(G_t\) rewards agreement
between the planned gait phase and the observed contact pattern.
The clearance target increases from \(0.05\,\mathrm{m}\) to
\(0.10\,\mathrm{m}\) with commanded foot speed, and the symmetry
coefficient is \(0.25\).
The gait reward and phase observation are disabled when
\(\|\mathbf c_t\|_2<0.1\), consistent with the standing branch of
the nominal-pose reward.

We additionally replace the native torso-orientation penalty with
\begin{equation}
-3\max
\left(
o_t^{\mathrm{torso}},
o_t^{\mathrm{pelvis}}
\right),
\end{equation}
where each term measures the squared horizontal component of gravity
projected into the corresponding body frame.
We also terminate an episode when the root height falls below
\(0.2\,\mathrm{m}\) or when a non-foot ground collision exceeds
\(10\,\mathrm{N}\); both events receive the same \(-200\)
non-time-limit termination penalty.
As in the original MJLab implementation, the complete weighted reward
is multiplied by the \(0.02\,\mathrm{s}\) control timestep.
\clearpage
\section{Full Benchmark Learning Curves}
\label{app:all_task_curves}

\begin{figure}[H]
    \centering
    \includegraphics[width=0.90\textwidth]
    {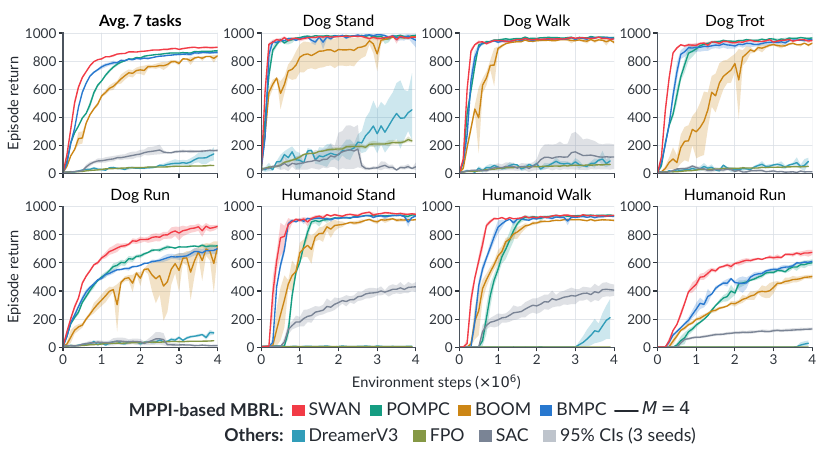}
    \caption{
    \textbf{Full learning curves on all \(7\) DMControl tasks under
    the half-budget setting.}
    All MPPI-based methods use \(M=4\) refinement iterations.
    }
    \label{fig:all_tasks_dmc_iter4}
\end{figure}

\begin{figure}[H]
    \centering
    \includegraphics[width=0.90\textwidth]
    {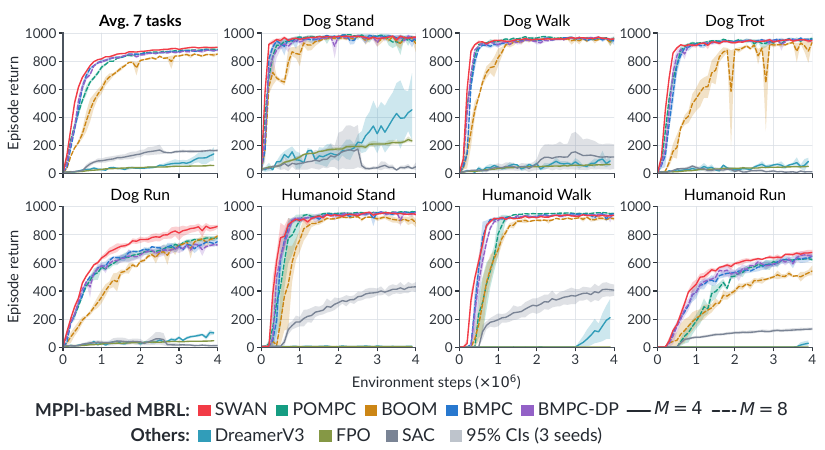}
    \caption{
    \textbf{Full learning curves on all \(7\) DMControl tasks under
    the full-budget comparison.}
    SWAN uses \(M=4\), while the MPPI-based baselines use \(M=8\).
    }
    \label{fig:all_tasks_dmc_iter8}
\end{figure}

\begin{figure}[H]
    \centering
    \includegraphics[width=\textwidth]
    {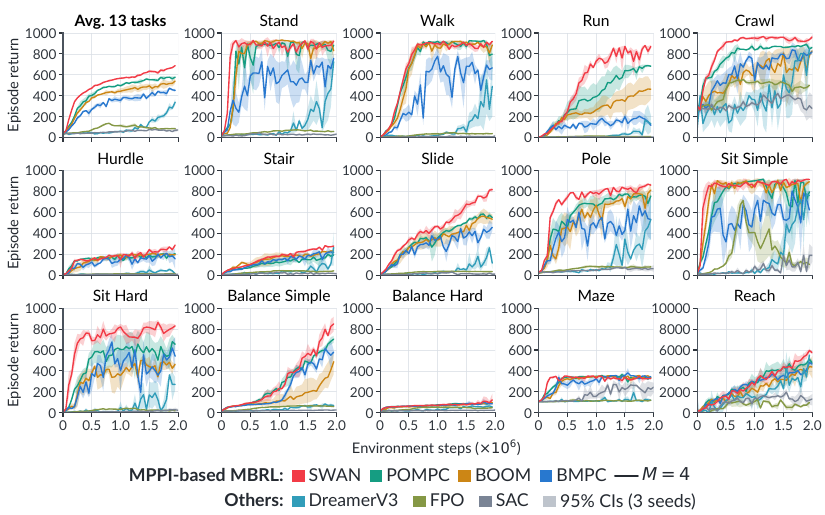}
    \caption{
    \textbf{Full learning curves on all \(14\) HumanoidBench tasks
    under the half-budget setting.}
    All MPPI-based methods use \(M=4\) refinement iterations.
    }
    \label{fig:all_tasks_humanoidbench_iter4}
\end{figure}

\begin{figure}[H]
    \centering
    \includegraphics[width=\textwidth]
    {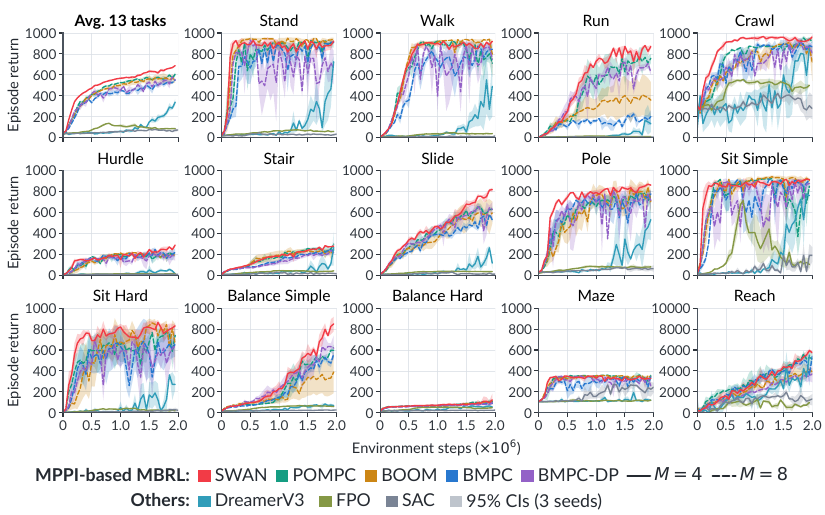}
    \caption{
    \textbf{Full learning curves on all \(14\) HumanoidBench tasks
    under the full-budget comparison.}
    SWAN uses \(M=4\), while the MPPI-based baselines use \(M=8\).
    }
    \label{fig:all_tasks_humanoidbench_iter8}
\end{figure}

\begin{figure}[H]
    \centering
    \includegraphics[width=0.90\textwidth]
    {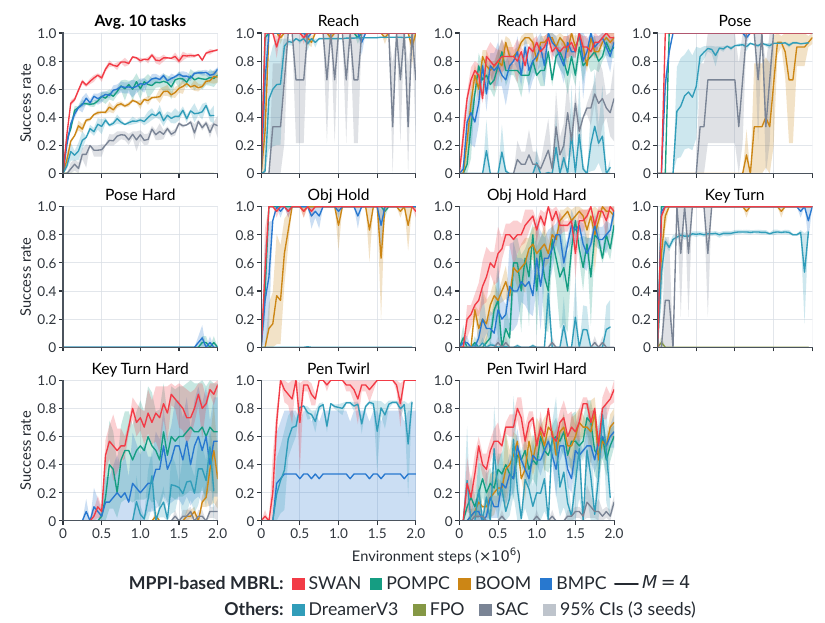}
    \caption{
    \textbf{Full learning curves on all \(10\) MyoSuite tasks under
    the half-budget setting.}
    All MPPI-based methods use \(M=4\) refinement iterations.
    }
    \label{fig:all_tasks_myosuite_iter4}
\end{figure}

\begin{figure}[H]
    \centering
    \includegraphics[width=0.90\textwidth]
    {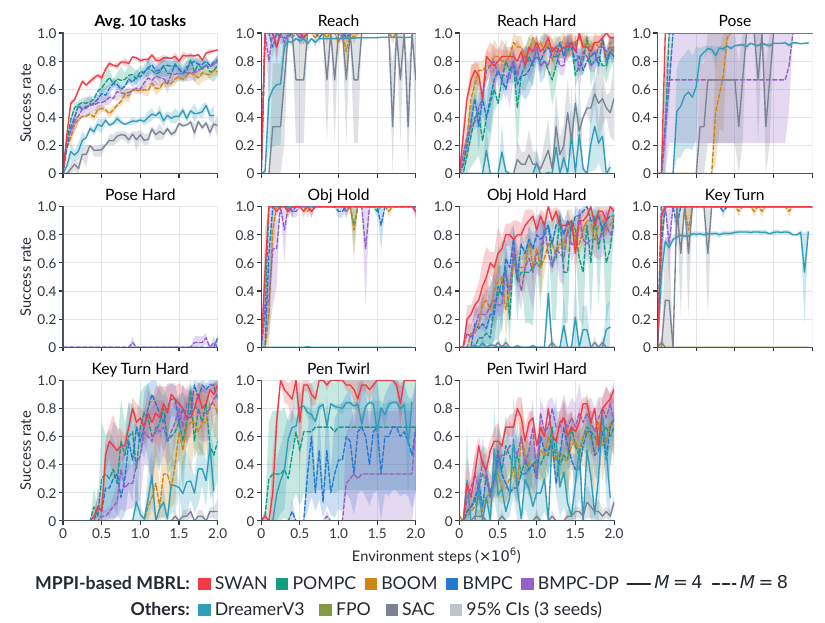}
    \caption{
    \textbf{Full learning curves on all \(10\) MyoSuite tasks under
    the full-budget comparison.}
    SWAN uses \(M=4\), while the MPPI-based baselines use \(M=8\).
    }
    \label{fig:all_tasks_myosuite_iter8}
\end{figure}

We provide the complete learning curves for all \(31\) benchmark
tasks evaluated in Sec.~\ref{subsec:main_result}.
Figs.~\ref{fig:all_tasks_dmc_iter4}--\ref{fig:all_tasks_myosuite_iter8}
report all \(7\) DMControl~\citep{tunyasuvunakool2020}, \(14\) HumanoidBench~\citep{sferrazza2024humanoidbench}, and \(10\) MyoSuite~\citep{caggiano2022myosuite}
tasks under the half- and full-budget comparisons.
SWAN uses \(M=4\) MPPI refinement iterations throughout.
The MPPI-based baselines use \(M=4\) in the half-budget setting and
\(M=8\) in the full-budget comparison.
Curves and shaded regions show means and \(95\%\) CIs over three seeds.

Across the complete task set, SWAN consistently achieves strong control performance relative to the competing baselines. Despite using only \(M=4\) refinement iterations, SWAN remains competitive with or outperforms the baselines on many tasks across all three benchmark suites, with larger gains under the matched refinement budget. These results further demonstrate the effectiveness of SWAN across diverse locomotion and manipulation tasks.

\setlength{\intextsep}{6pt}
\captionsetup{skip=3pt}

\clearpage

\clearpage

\clearpage
\section{Final Performance and Sample Efficiency}
\label{app:sample_efficiency}
\raggedbottom

We provide complete final-performance and sample-efficiency
statistics for Sec.~\ref{subsec:main_result}, reporting
full- and half-budget comparisons separately.
SWAN uses \(M=4\) MPPI refinement iterations in both settings,
while the MPPI-based baselines use \(M=8\) and \(M=4\),
respectively.

For each method, task-level evaluation curves are averaged
over three random seeds.
Suite-average curves use evaluation steps shared by all
included tasks and seeds, averaging first across tasks within
each seed and then across seeds.
The HumanoidBench suite average includes 13 tasks, excluding
Reach.
For each task or suite average, \textit{Final} is the last
observed value of the corresponding mean evaluation curve,
without smoothing or averaging over a terminal window.
Because individual task curves may extend beyond the last
shared evaluation step, suite-level \textit{Final} need not
equal the average of task-level \textit{Final} values.

We measure sample efficiency by \(T_{80}\), the first observed
environment step at which a method's mean evaluation curve
reaches \(80\%\) of SWAN's \textit{Final} for the corresponding
task or suite average.
All methods therefore share the same reference threshold.
We report \(T_{80}\) in millions of environment steps, with
lower values indicating better sample efficiency.
The symbol ``--'' indicates that the threshold is not reached
within the recorded evaluations.

\noindent\textbf{Analysis.}
Under the full-budget comparison in
Tables~\ref{tab:dmc_full_budget}, \ref{tab:humanoid_full_budget},
and \ref{tab:myosuite_full_budget}, SWAN achieves the highest
suite-average performance across all three benchmarks despite using
only \(M=4\) refinement iterations.
On DMControl, SWAN obtains an average return of \(898.8\) with
\(T_{80}=0.60\)M, compared with \(883.7\) and \(1.00\)M for
PO-MPC with \(M=8\).
On \textit{Dog-Run}, SWAN improves the final return over BOOM by
\(67.2\), while reducing \(T_{80}\) from \(2.80\)M to \(1.30\)M;
on \textit{Humanoid-Run}, it improves over BMPC by \(25.8\) with
\(T_{80}\) reduced from \(2.10\)M to \(1.40\)M.
On HumanoidBench, SWAN reaches an average return of \(688.0\) with
\(T_{80}=0.90\)M, compared with \(604.0\) and \(1.25\)M for
PO-MPC.
The gains are particularly pronounced on \textit{Run} and
\textit{Sit Hard}, where SWAN improves the final return over PO-MPC
by \(112.2\) and \(92.0\), while reducing \(T_{80}\) by
\(37.9\%\) and \(62.5\%\), respectively.
On MyoSuite, SWAN achieves \(88.0\%\) average success with
\(T_{80}=0.55\)M, compared with \(81.0\%\) and \(1.05\)M for
BMPC.
On \textit{Pen Twirl}, SWAN reaches \(100\%\) final success with
\(T_{80}=0.25\)M, whereas the MPPI-based baselines reach at most
\(66.7\%\) and do not reach the shared threshold.

Under the matched \(M=4\) half-budget setting in
Tables~\ref{tab:dmc_half_budget}, \ref{tab:humanoid_half_budget},
and \ref{tab:myosuite_half_budget}, the gaps generally increase.
SWAN improves the suite-average return over PO-MPC by \(25.3\) on
DMControl and \(110.5\) on HumanoidBench, while reducing
\(T_{80}\) by \(50.0\%\) and \(37.9\%\), respectively.
On MyoSuite, SWAN improves the average success rate over BMPC by
\(13.7\) percentage points and reduces \(T_{80}\) by \(63.3\%\).
The wider gaps therefore indicate that the competing MPPI-based
methods benefit more from additional refinement, whereas SWAN
maintains strong control performance and sample efficiency
under the reduced planning budget. Moreover, BMPC and BOOM no longer reach the 
HumanoidBench suite-average threshold with four iterations, while PO-MPC
and BOOM no longer reach the MyoSuite threshold.

\begin{table}[htb]
\centering
\small
\setlength{\tabcolsep}{2pt}
\renewcommand{\arraystretch}{1.00}
\captionsetup{font=small,skip=5pt,position=top}
\caption{\textbf{DMControl full-budget comparison.}
SWAN uses \(M=4\) and MPPI-based baselines use \(M=8\).
Paired rows report final episode return \(\uparrow\) and \(T_{80}\) \(\downarrow\).}
\label{tab:dmc_full_budget}
\begin{tabular*}{\linewidth}{@{\extracolsep{\fill}}llrrrrrrrr@{}}
\toprule
&&\multicolumn{5}{c}{MPPI-based methods}&\multicolumn{3}{c}{Other baselines}\\
\cmidrule(lr){3-7}\cmidrule(l){8-10}
Task & Metric & \textbf{SWAN} & PO-MPC & BMPC & BOOM & BMPC-DP & DreamerV3 & FPO & SAC \\
 & & \(M=4\) & \(M=8\) & \(M=8\) & \(M=8\) & \(M=8\) & & & \\
\midrule
\multirow{2}{*}{\textbf{Avg. 7 tasks}} & Final & \textbf{898.8}\phantom{0} & 883.7\phantom{0} & 881.5\phantom{0} & 850.1\phantom{0} & 876.0\phantom{0} & 140.0\phantom{0} & 57.6\phantom{0} & 165.4\phantom{0} \\
 & \(T_{80}\) & \textbf{0.60} & 1.00 & 0.70 & 1.40 & 0.80 & -- & -- & -- \\
\midrule
\multirow{2}{*}{Dog Stand} & Final & \textbf{970.5}\phantom{0} & 947.5\phantom{0} & 965.7\phantom{0} & 924.3\phantom{0} & 958.1\phantom{0} & 454.0\phantom{0} & 232.5\phantom{0} & 48.9\phantom{0} \\
 & \(T_{80}\) & \textbf{0.20} & 0.30 & \textbf{0.20} & 0.70 & 0.30 & -- & -- & -- \\
\addlinespace[1.3pt]
\multirow{2}{*}{Dog Walk} & Final & 957.5\phantom{0} & \textbf{964.1}\phantom{0} & 943.8\phantom{0} & 960.5\phantom{0} & 957.0\phantom{0} & 90.1\phantom{0} & 64.6\phantom{0} & 118.3\phantom{0} \\
 & \(T_{80}\) & \textbf{0.30} & 0.40 & 0.40 & 1.00 & 0.40 & -- & -- & -- \\
\addlinespace[1.3pt]
\multirow{2}{*}{Dog Trot} & Final & 946.6\phantom{0} & \textbf{960.4}\phantom{0} & 954.5\phantom{0} & 934.2\phantom{0} & 936.6\phantom{0} & 86.7\phantom{0} & 51.6\phantom{0} & 13.3\phantom{0} \\
 & \(T_{80}\) & \textbf{0.40} & 0.50 & 0.50 & 1.60 & 0.50 & -- & -- & -- \\
\addlinespace[1.3pt]
\multirow{2}{*}{Dog Run} & Final & \textbf{859.4}\phantom{0} & 777.0\phantom{0} & 751.5\phantom{0} & 792.2\phantom{0} & 727.2\phantom{0} & 101.8\phantom{0} & 46.4\phantom{0} & 11.8\phantom{0} \\
 & \(T_{80}\) & \textbf{1.30} & 2.50 & 2.10 & 2.80 & 2.30 & -- & -- & -- \\
\addlinespace[1.3pt]
\multirow{2}{*}{Humanoid Stand} & Final & 945.1\phantom{0} & \textbf{962.5}\phantom{0} & 957.0\phantom{0} & 883.9\phantom{0} & 960.0\phantom{0} & 7.5\phantom{0} & 5.5\phantom{0} & 429.8\phantom{0} \\
 & \(T_{80}\) & \textbf{0.50} & 0.70 & 0.60 & 1.00 & 0.70 & -- & -- & -- \\
\addlinespace[1.3pt]
\multirow{2}{*}{Humanoid Walk} & Final & 937.8\phantom{0} & \textbf{952.2}\phantom{0} & 949.3\phantom{0} & 913.6\phantom{0} & 948.9\phantom{0} & 211.1\phantom{0} & 1.6\phantom{0} & 406.2\phantom{0} \\
 & \(T_{80}\) & 0.70 & 1.00 & \textbf{0.60} & 1.10 & 0.80 & -- & -- & -- \\
\addlinespace[1.3pt]
\multirow{2}{*}{Humanoid Run} & Final & \textbf{674.7}\phantom{0} & 622.5\phantom{0} & 648.9\phantom{0} & 542.2\phantom{0} & 644.2\phantom{0} & 28.5\phantom{0} & 1.1\phantom{0} & 129.8\phantom{0} \\
 & \(T_{80}\) & \textbf{1.40} & 2.30 & 2.10 & 4.00 & 1.60 & -- & -- & -- \\
\bottomrule
\end{tabular*}

\par\vspace{12pt}

\caption{\textbf{DMControl half-budget comparison.}
All methods use \(M=4\). Each method reports final episode return \(\uparrow\) and \(T_{80}\) \(\downarrow\).}
\label{tab:dmc_half_budget}
\begin{tabular*}{\linewidth}{@{\extracolsep{\fill}}lrrrrrrrr@{}}
\toprule
Task & \multicolumn{2}{c}{\textbf{SWAN}} & \multicolumn{2}{c}{PO-MPC} & \multicolumn{2}{c}{BMPC} & \multicolumn{2}{c}{BOOM} \\
\cmidrule(lr){2-3}\cmidrule(lr){4-5}\cmidrule(lr){6-7}\cmidrule(l){8-9}
& Final \(\uparrow\) & \(T_{80}\) \(\downarrow\) & Final \(\uparrow\) & \(T_{80}\) \(\downarrow\) & Final \(\uparrow\) & \(T_{80}\) \(\downarrow\) & Final \(\uparrow\) & \(T_{80}\) \(\downarrow\) \\
\midrule
\textbf{Avg. 7 tasks} & \textbf{898.8} & \textbf{0.60} & 873.5 & 1.20 & 862.4 & 0.80 & 837.7 & 1.90 \\
\midrule
Dog Stand & 970.5 & \textbf{0.20} & \textbf{983.0} & 0.30 & 948.4 & \textbf{0.20} & 972.0 & 1.00 \\
Dog Walk & 957.5 & \textbf{0.30} & \textbf{968.5} & 0.50 & 960.3 & 0.40 & 933.9 & 0.80 \\
Dog Trot & 946.6 & \textbf{0.40} & \textbf{964.6} & 0.70 & 952.8 & 0.60 & 927.7 & 2.00 \\
Dog Run & \textbf{859.4} & \textbf{1.30} & 721.2 & 2.30 & 701.2 & 3.80 & 718.1 & 3.70 \\
Humanoid Stand & \textbf{945.1} & \textbf{0.50} & 941.3 & 1.20 & 934.4 & 0.70 & 907.4 & 1.40 \\
Humanoid Walk & \textbf{937.8} & \textbf{0.70} & 934.9 & 1.30 & 929.6 & 0.90 & 901.8 & 1.20 \\
Humanoid Run & \textbf{674.7} & \textbf{1.40} & 600.7 & 3.10 & 610.4 & 2.80 & 502.9 & -- \\
\bottomrule
\end{tabular*}

\par\vspace{5pt}
\begin{minipage}{\linewidth}
\small
\textit{Note.} \(T_{80}\) is in millions of environment steps.
\end{minipage}
\end{table}

\clearpage
\begin{table}[htb]
\centering
\small
\setlength{\tabcolsep}{2pt}
\renewcommand{\arraystretch}{1.00}
\captionsetup{font=small,skip=5pt,position=top}
\caption{\textbf{HumanoidBench full-budget comparison.}
SWAN uses \(M=4\) and MPPI-based baselines use \(M=8\).
Paired rows report final episode return \(\uparrow\) and \(T_{80}\) \(\downarrow\).}
\label{tab:humanoid_full_budget}
\begin{tabular*}{\linewidth}{@{\extracolsep{\fill}}llrrrrrrrr@{}}
\toprule
&&\multicolumn{5}{c}{MPPI-based methods}&\multicolumn{3}{c}{Other baselines}\\
\cmidrule(lr){3-7}\cmidrule(l){8-10}
Task & Metric & \textbf{SWAN} & PO-MPC & BMPC & BOOM & BMPC-DP & DreamerV3 & FPO & SAC \\
 & & \(M=4\) & \(M=8\) & \(M=8\) & \(M=8\) & \(M=8\) & & & \\
\midrule
\multirow{2}{*}{\textbf{Avg. 13 tasks}} & Final & \textbf{688.0}\phantom{0} & 604.0\phantom{0} & 551.1\phantom{0} & 545.9\phantom{0} & 548.3\phantom{0} & 338.7\phantom{0} & 86.5\phantom{0} & 72.0\phantom{0} \\
 & \(T_{80}\) & \textbf{0.90} & 1.25 & 1.95 & 1.45 & 1.80 & -- & -- & -- \\
\midrule
\multirow{2}{*}{Stand} & Final & 922.6\phantom{0} & 896.3\phantom{0} & 910.3\phantom{0} & \textbf{943.4}\phantom{0} & 663.3\phantom{0} & 725.5\phantom{0} & 59.0\phantom{0} & 33.1\phantom{0} \\
 & \(T_{80}\) & \textbf{0.15} & 0.30 & 0.40 & 0.20 & 0.25 & -- & -- & -- \\
\addlinespace[1.3pt]
\multirow{2}{*}{Walk} & Final & \textbf{920.8}\phantom{0} & 711.6\phantom{0} & 844.0\phantom{0} & 796.1\phantom{0} & 749.3\phantom{0} & 489.8\phantom{0} & 37.6\phantom{0} & 15.6\phantom{0} \\
 & \(T_{80}\) & 0.55 & 0.60 & 0.80 & \textbf{0.50} & 0.55 & -- & -- & -- \\
\addlinespace[1.3pt]
\multirow{2}{*}{Run} & Final & \textbf{870.4}\phantom{0} & 758.2\phantom{0} & 200.1\phantom{0} & 358.3\phantom{0} & 625.9\phantom{0} & 133.1\phantom{0} & 10.1\phantom{0} & 6.8\phantom{0} \\
 & \(T_{80}\) & \textbf{0.90} & 1.45 & -- & -- & 1.45 & -- & -- & -- \\
\addlinespace[1.3pt]
\multirow{2}{*}{Crawl} & Final & \textbf{961.9}\phantom{0} & 919.6\phantom{0} & 868.9\phantom{0} & 721.6\phantom{0} & 809.4\phantom{0} & 852.3\phantom{0} & 499.3\phantom{0} & 277.7\phantom{0} \\
 & \(T_{80}\) & \textbf{0.35} & 0.50 & 0.75 & 0.95 & 0.65 & 1.80 & -- & -- \\
\addlinespace[1.3pt]
\multirow{2}{*}{Hurdle} & Final & \textbf{287.8}\phantom{0} & 208.7\phantom{0} & 219.7\phantom{0} & 193.0\phantom{0} & 180.1\phantom{0} & 13.7\phantom{0} & 14.3\phantom{0} & 8.1\phantom{0} \\
 & \(T_{80}\) & \textbf{1.50} & -- & 1.60 & -- & -- & -- & -- & -- \\
\addlinespace[1.3pt]
\multirow{2}{*}{Stair} & Final & 278.4\phantom{0} & \textbf{296.9}\phantom{0} & 238.1\phantom{0} & 245.2\phantom{0} & 213.1\phantom{0} & 264.0\phantom{0} & 41.0\phantom{0} & 12.7\phantom{0} \\
 & \(T_{80}\) & 1.35 & 1.25 & 1.85 & \textbf{1.15} & 1.90 & 1.95 & -- & -- \\
\addlinespace[1.3pt]
\multirow{2}{*}{Slide} & Final & \textbf{818.5}\phantom{0} & 630.1\phantom{0} & 529.7\phantom{0} & 596.2\phantom{0} & 628.6\phantom{0} & 115.5\phantom{0} & 34.7\phantom{0} & 10.0\phantom{0} \\
 & \(T_{80}\) & \textbf{1.50} & -- & -- & -- & 1.70 & -- & -- & -- \\
\addlinespace[1.3pt]
\multirow{2}{*}{Pole} & Final & \textbf{860.8}\phantom{0} & 755.6\phantom{0} & 776.3\phantom{0} & 702.3\phantom{0} & 715.9\phantom{0} & 531.4\phantom{0} & 79.4\phantom{0} & 61.2\phantom{0} \\
 & \(T_{80}\) & \textbf{0.35} & 0.90 & 0.65 & 0.90 & 0.70 & -- & -- & -- \\
\addlinespace[1.3pt]
\multirow{2}{*}{Sit Simple} & Final & \textbf{915.1}\phantom{0} & 850.8\phantom{0} & 913.4\phantom{0} & 903.6\phantom{0} & 875.3\phantom{0} & 760.7\phantom{0} & 103.1\phantom{0} & 189.8\phantom{0} \\
 & \(T_{80}\) & \textbf{0.15} & 0.30 & 0.30 & 0.30 & 0.45 & 1.90 & -- & -- \\
\addlinespace[1.3pt]
\multirow{2}{*}{Sit Hard} & Final & \textbf{831.3}\phantom{0} & 739.3\phantom{0} & 657.4\phantom{0} & 659.8\phantom{0} & 642.2\phantom{0} & 271.6\phantom{0} & 18.9\phantom{0} & 23.6\phantom{0} \\
 & \(T_{80}\) & \textbf{0.30} & 0.80 & 0.95 & 0.65 & 0.55 & -- & -- & -- \\
\addlinespace[1.3pt]
\multirow{2}{*}{Balance Simple} & Final & \textbf{849.8}\phantom{0} & 598.5\phantom{0} & 561.2\phantom{0} & 389.9\phantom{0} & 618.4\phantom{0} & 69.5\phantom{0} & 59.1\phantom{0} & 26.5\phantom{0} \\
 & \(T_{80}\) & \textbf{1.75} & -- & -- & -- & -- & -- & -- & -- \\
\addlinespace[1.3pt]
\multirow{2}{*}{Balance Hard} & Final & \textbf{123.6}\phantom{0} & 108.8\phantom{0} & 89.6\phantom{0} & 101.0\phantom{0} & 97.4\phantom{0} & 55.9\phantom{0} & 47.9\phantom{0} & 26.1\phantom{0} \\
 & \(T_{80}\) & 1.80 & \textbf{1.60} & -- & 1.80 & -- & -- & -- & -- \\
\addlinespace[1.3pt]
\multirow{2}{*}{Maze} & Final & 328.4\phantom{0} & \textbf{354.4}\phantom{0} & 345.6\phantom{0} & 343.7\phantom{0} & 259.3\phantom{0} & 119.6\phantom{0} & 120.2\phantom{0} & 245.4\phantom{0} \\
 & \(T_{80}\) & \textbf{0.15} & 0.25 & 0.30 & 0.20 & 0.25 & -- & -- & 1.20 \\
\addlinespace[1.3pt]
\multirow{2}{*}{Reach} & Final & \textbf{5750.5}\phantom{0} & 5173.3\phantom{0} & 5387.1\phantom{0} & 3893.4\phantom{0} & 3712.8\phantom{0} & 5517.0\phantom{0} & 977.6\phantom{0} & 1317.2\phantom{0} \\
 & \(T_{80}\) & \textbf{1.70} & \textbf{1.70} & \textbf{1.70} & -- & -- & 1.95 & -- & -- \\
\bottomrule
\end{tabular*}

\par\vspace{12pt}

\caption{\textbf{HumanoidBench half-budget comparison.}
All methods use \(M=4\). Each method reports final episode return \(\uparrow\) and \(T_{80}\) \(\downarrow\).}
\label{tab:humanoid_half_budget}
\begin{tabular*}{\linewidth}{@{\extracolsep{\fill}}lrrrrrrrr@{}}
\toprule
Task & \multicolumn{2}{c}{\textbf{SWAN}} & \multicolumn{2}{c}{PO-MPC} & \multicolumn{2}{c}{BMPC} & \multicolumn{2}{c}{BOOM} \\
\cmidrule(lr){2-3}\cmidrule(lr){4-5}\cmidrule(lr){6-7}\cmidrule(l){8-9}
& Final \(\uparrow\) & \(T_{80}\) \(\downarrow\) & Final \(\uparrow\) & \(T_{80}\) \(\downarrow\) & Final \(\uparrow\) & \(T_{80}\) \(\downarrow\) & Final \(\uparrow\) & \(T_{80}\) \(\downarrow\) \\
\midrule
\textbf{Avg. 13 tasks} & \textbf{688.0} & \textbf{0.90} & 577.5 & 1.45 & 452.0 & -- & 540.3 & -- \\
\midrule
Stand & \textbf{922.6} & \textbf{0.15} & 884.0 & 0.25 & 756.2 & 1.55 & 830.8 & 0.30 \\
Walk & \textbf{920.8} & \textbf{0.55} & 798.0 & 0.60 & 666.6 & 1.00 & 884.2 & 0.60 \\
Run & \textbf{870.4} & \textbf{0.90} & 682.9 & -- & 117.7 & -- & 462.7 & -- \\
Crawl & \textbf{961.9} & \textbf{0.35} & 862.5 & 0.55 & 819.9 & 0.90 & 828.3 & 1.50 \\
Hurdle & \textbf{287.8} & \textbf{1.50} & 205.9 & -- & 155.9 & -- & 195.6 & -- \\
Stair & \textbf{278.4} & \textbf{1.35} & 193.1 & -- & 215.5 & 1.70 & 231.4 & 1.75 \\
Slide & \textbf{818.5} & \textbf{1.50} & 551.7 & -- & 457.8 & -- & 535.2 & -- \\
Pole & \textbf{860.8} & \textbf{0.35} & 756.9 & 1.15 & 532.6 & -- & 814.2 & 1.35 \\
Sit Simple & \textbf{915.1} & \textbf{0.15} & 796.6 & 0.40 & 624.0 & 1.65 & 893.5 & 0.25 \\
Sit Hard & \textbf{831.3} & \textbf{0.30} & 654.9 & 1.90 & 541.1 & 0.85 & 463.0 & -- \\
Balance Simple & \textbf{849.8} & \textbf{1.75} & 703.8 & 1.90 & 580.1 & -- & 488.5 & -- \\
Balance Hard & \textbf{123.6} & \textbf{1.80} & 94.0 & -- & 85.4 & \textbf{1.80} & 97.3 & -- \\
Maze & 328.4 & \textbf{0.15} & 337.4 & 0.30 & \textbf{344.4} & 0.35 & 335.7 & 0.40 \\
Reach & \textbf{5750.5} & 1.70 & 4755.5 & \textbf{1.60} & 4658.0 & 1.95 & 4370.1 & -- \\
\bottomrule
\end{tabular*}

\par\vspace{5pt}
\begin{minipage}{\linewidth}
\small
\textit{Note.} \(T_{80}\) is in millions of environment steps.
The suite average excludes Reach, which is reported separately.
\end{minipage}
\end{table}

\clearpage
\begin{table}[htb]
\centering
\small
\setlength{\tabcolsep}{2pt}
\renewcommand{\arraystretch}{1.00}
\captionsetup{font=small,skip=5pt,position=top}
\caption{\textbf{MyoSuite full-budget comparison.}
SWAN uses \(M=4\) and MPPI-based baselines use \(M=8\).
Paired rows report final success rate (\%) \(\uparrow\) and \(T_{80}\) \(\downarrow\).}
\label{tab:myosuite_full_budget}
\begin{tabular*}{\linewidth}{@{\extracolsep{\fill}}llrrrrrrrr@{}}
\toprule
&&\multicolumn{5}{c}{MPPI-based methods}&\multicolumn{3}{c}{Other baselines}\\
\cmidrule(lr){3-7}\cmidrule(l){8-10}
Task & Metric & \textbf{SWAN} & PO-MPC & BMPC & BOOM & BMPC-DP & DreamerV3 & FPO & SAC \\
 & & \(M=4\) & \(M=8\) & \(M=8\) & \(M=8\) & \(M=8\) & & & \\
\midrule
\multirow{2}{*}{\textbf{Avg. 10 tasks}} & Final & \textbf{88.0}\phantom{0} & 77.3\phantom{0} & 81.0\phantom{0} & 72.3\phantom{0} & 79.7\phantom{0} & 41.3\phantom{0} & 0.0\phantom{0} & 34.0\phantom{0} \\
 & \(T_{80}\) & \textbf{0.55} & 1.00 & 1.05 & 1.65 & 1.40 & -- & -- & -- \\
\midrule
\multirow{2}{*}{Reach} & Final & \textbf{100.0}\phantom{0} & \textbf{100.0}\phantom{0} & \textbf{100.0}\phantom{0} & \textbf{100.0}\phantom{0} & \textbf{100.0}\phantom{0} & 97.0\phantom{0} & 0.0\phantom{0} & 66.7\phantom{0} \\
 & \(T_{80}\) & \textbf{0.05} & \textbf{0.05} & \textbf{0.05} & \textbf{0.05} & \textbf{0.05} & 0.30 & -- & 0.35 \\
\addlinespace[1.3pt]
\multirow{2}{*}{Reach Hard} & Final & \textbf{96.7}\phantom{0} & 90.0\phantom{0} & 83.3\phantom{0} & 86.7\phantom{0} & 86.7\phantom{0} & 4.3\phantom{0} & 0.0\phantom{0} & 53.3\phantom{0} \\
 & \(T_{80}\) & \textbf{0.35} & 0.60 & 0.70 & 0.45 & 0.45 & -- & -- & -- \\
\addlinespace[1.3pt]
\multirow{2}{*}{Pose} & Final & \textbf{100.0}\phantom{0} & \textbf{100.0}\phantom{0} & \textbf{100.0}\phantom{0} & \textbf{100.0}\phantom{0} & \textbf{100.0}\phantom{0} & 93.0\phantom{0} & 0.0\phantom{0} & \textbf{100.0}\phantom{0} \\
 & \(T_{80}\) & \textbf{0.10} & 0.15 & 0.20 & 0.90 & 1.75 & 0.55 & -- & 1.15 \\
\addlinespace[1.3pt]
\multirow{2}{*}{Pose Hard} & Final & 0.0\phantom{0} & 0.0\phantom{0} & \textbf{6.7}\phantom{0} & 0.0\phantom{0} & 0.0\phantom{0} & 0.0\phantom{0} & 0.0\phantom{0} & 0.0\phantom{0} \\
 & \(T_{80}\) & -- & -- & -- & -- & -- & -- & -- & -- \\
\addlinespace[1.3pt]
\multirow{2}{*}{Obj Hold} & Final & 96.7\phantom{0} & \textbf{100.0}\phantom{0} & \textbf{100.0}\phantom{0} & \textbf{100.0}\phantom{0} & 96.7\phantom{0} & 0.0\phantom{0} & 0.0\phantom{0} & 0.0\phantom{0} \\
 & \(T_{80}\) & \textbf{0.10} & \textbf{0.10} & 0.15 & 0.15 & 0.25 & -- & -- & -- \\
\addlinespace[1.3pt]
\multirow{2}{*}{Obj Hold Hard} & Final & \textbf{96.7}\phantom{0} & 93.3\phantom{0} & 93.3\phantom{0} & 93.3\phantom{0} & 86.7\phantom{0} & 14.3\phantom{0} & 0.0\phantom{0} & 0.0\phantom{0} \\
 & \(T_{80}\) & \textbf{0.60} & 1.10 & 1.00 & 1.40 & 1.10 & -- & -- & -- \\
\addlinespace[1.3pt]
\multirow{2}{*}{Key Turn} & Final & \textbf{100.0}\phantom{0} & \textbf{100.0}\phantom{0} & \textbf{100.0}\phantom{0} & \textbf{100.0}\phantom{0} & \textbf{100.0}\phantom{0} & 82.0\phantom{0} & 0.0\phantom{0} & \textbf{100.0}\phantom{0} \\
 & \(T_{80}\) & \textbf{0.05} & \textbf{0.05} & 0.10 & \textbf{0.05} & \textbf{0.05} & 0.45 & -- & 0.25 \\
\addlinespace[1.3pt]
\multirow{2}{*}{Key Turn Hard} & Final & \textbf{96.7}\phantom{0} & 56.7\phantom{0} & 86.7\phantom{0} & 76.7\phantom{0} & 90.0\phantom{0} & 21.3\phantom{0} & 0.0\phantom{0} & 6.7\phantom{0} \\
 & \(T_{80}\) & \textbf{0.90} & 1.30 & 1.10 & 1.90 & 1.50 & -- & -- & -- \\
\addlinespace[1.3pt]
\multirow{2}{*}{Pen Twirl} & Final & \textbf{100.0}\phantom{0} & 66.7\phantom{0} & 66.7\phantom{0} & 0.0\phantom{0} & 66.7\phantom{0} & 84.3\phantom{0} & 0.0\phantom{0} & 0.0\phantom{0} \\
 & \(T_{80}\) & \textbf{0.25} & -- & -- & -- & -- & 0.50 & -- & -- \\
\addlinespace[1.3pt]
\multirow{2}{*}{Pen Twirl Hard} & Final & \textbf{93.3}\phantom{0} & 66.7\phantom{0} & 73.3\phantom{0} & 66.7\phantom{0} & 70.0\phantom{0} & 16.7\phantom{0} & 0.0\phantom{0} & 13.3\phantom{0} \\
 & \(T_{80}\) & \textbf{0.75} & -- & -- & -- & 1.05 & 1.85 & -- & -- \\
\bottomrule
\end{tabular*}

\par\vspace{12pt}

\caption{\textbf{MyoSuite half-budget comparison.}
All methods use \(M=4\). Each method reports final success rate (\%) \(\uparrow\) and \(T_{80}\) \(\downarrow\).}
\label{tab:myosuite_half_budget}
\begin{tabular*}{\linewidth}{@{\extracolsep{\fill}}lrrrrrrrr@{}}
\toprule
Task & \multicolumn{2}{c}{\textbf{SWAN}} & \multicolumn{2}{c}{PO-MPC} & \multicolumn{2}{c}{BMPC} & \multicolumn{2}{c}{BOOM} \\
\cmidrule(lr){2-3}\cmidrule(lr){4-5}\cmidrule(lr){6-7}\cmidrule(l){8-9}
& Final \(\uparrow\) & \(T_{80}\) \(\downarrow\) & Final \(\uparrow\) & \(T_{80}\) \(\downarrow\) & Final \(\uparrow\) & \(T_{80}\) \(\downarrow\) & Final \(\uparrow\) & \(T_{80}\) \(\downarrow\) \\
\midrule
\textbf{Avg. 10 tasks} & \textbf{88.0} & \textbf{0.55} & 70.0 & -- & 74.3 & 1.50 & 68.7 & -- \\
\midrule
Reach & \textbf{100.0} & \textbf{0.05} & \textbf{100.0} & \textbf{0.05} & \textbf{100.0} & 0.10 & \textbf{100.0} & 0.10 \\
Reach Hard & \textbf{96.7} & 0.35 & 90.0 & 0.55 & 93.3 & 0.70 & \textbf{96.7} & \textbf{0.30} \\
Pose & \textbf{100.0} & \textbf{0.10} & \textbf{100.0} & 0.15 & \textbf{100.0} & 0.15 & 96.7 & 1.55 \\
Pose Hard & \textbf{0.0} & -- & \textbf{0.0} & -- & \textbf{0.0} & -- & \textbf{0.0} & -- \\
Obj Hold & 96.7 & \textbf{0.10} & \textbf{100.0} & \textbf{0.10} & \textbf{100.0} & 0.15 & \textbf{100.0} & 0.35 \\
Obj Hold Hard & \textbf{96.7} & \textbf{0.60} & 86.7 & 0.95 & \textbf{96.7} & 1.25 & 93.3 & 1.20 \\
Key Turn & \textbf{100.0} & \textbf{0.05} & \textbf{100.0} & \textbf{0.05} & \textbf{100.0} & \textbf{0.05} & \textbf{100.0} & 0.10 \\
Key Turn Hard & \textbf{96.7} & \textbf{0.90} & 63.3 & -- & 56.7 & -- & 30.0 & -- \\
Pen Twirl & \textbf{100.0} & \textbf{0.25} & 0.0 & -- & 33.3 & -- & 0.0 & -- \\
Pen Twirl Hard & \textbf{93.3} & \textbf{0.75} & 60.0 & -- & 63.3 & -- & 70.0 & 1.75 \\
\bottomrule
\end{tabular*}

\par\vspace{5pt}
\begin{minipage}{\linewidth}
\small
\textit{Note.} \(T_{80}\) is in millions of environment steps.
For Pose Hard, SWAN's final success rate and the reference
threshold are zero. Its unreported \(T_{80}\) entries should not be interpreted
as failure to reach a positive threshold.
\end{minipage}
\end{table}

\section{Unitree Sim-to-Real Transfer}
\label{app:unitree_results}
\begin{wrapfigure}[15]{r}{0.5\columnwidth}
  \centering
  \begin{subfigure}[t]{\linewidth}
    \centering
    \includegraphics[width=\linewidth]{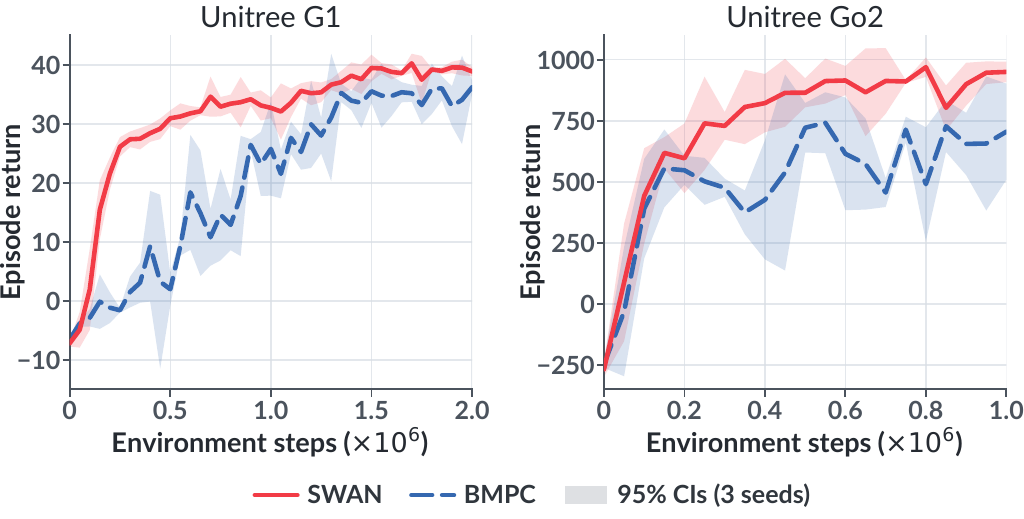}
  \end{subfigure}
  \caption{
  \textbf{Unitree G1 and Go2 locomotion in simulation.}
    Comparison between SWAN (solid) and BMPC (dashed), where SWAN uses half as many MPPI refinement iterations.
    Curves and shading show means and 95\% CIs over 3 seeds.
  }
  \label{fig:unitree_simulation}
\end{wrapfigure}

\noindent\textbf{Unitree G1 and Go2 Locomotion.}
In this section, we evaluate on velocity tracking tasks for Unitree 
G1 and Go2 in MJLab~\citep{zakka2026mjlab} to assess whether SWAN's
performance and sample-efficiency gains extend to robot locomotion.
As shown in Fig.~\ref{fig:unitree_simulation}, SWAN outperforms
BMPC in final return and sample efficiency while using half as
many MPPI refinement iterations on both tasks.
Specifically, BMPC uses \(M=8\) and \(M=6\) on G1 and Go2,
respectively, whereas SWAN uses \(M=4\) and \(M=3\).
The sample-efficiency advantage is particularly pronounced
on G1, demonstrating effective learning on the higher-dimensional
task despite the smaller refinement budget.


\noindent\textbf{Unitree Go2 Sim-to-Real Transfer.} 
Beyond simulation, we evaluate SWAN's ability to
zero-shot sim-to-real transfer by deploying the trained 
SWAN controller on a physical Unitree Go2.
To facilitate sim-to-real transfer, we apply 
domain randomization alongside the terrain curriculum in policy 
learning.
After training on the simulation, SWAN demonstrates zero-shot sim-to-real transfer on a physical Unitree Go2 with single NVIDIA GeForce RTX 5090.
Under a limited online planning budget, the robot tracks commanded forward, backward, and lateral
velocities, turns in both directions, and follows figure-eight
trajectories that combine translation with changes in turning
direction.

\begin{figure}[t]
    \vspace{-0.1in}
    \centering
    \begin{subfigure}[t]{0.96\linewidth}
        \centering
        \includegraphics[width=\linewidth]{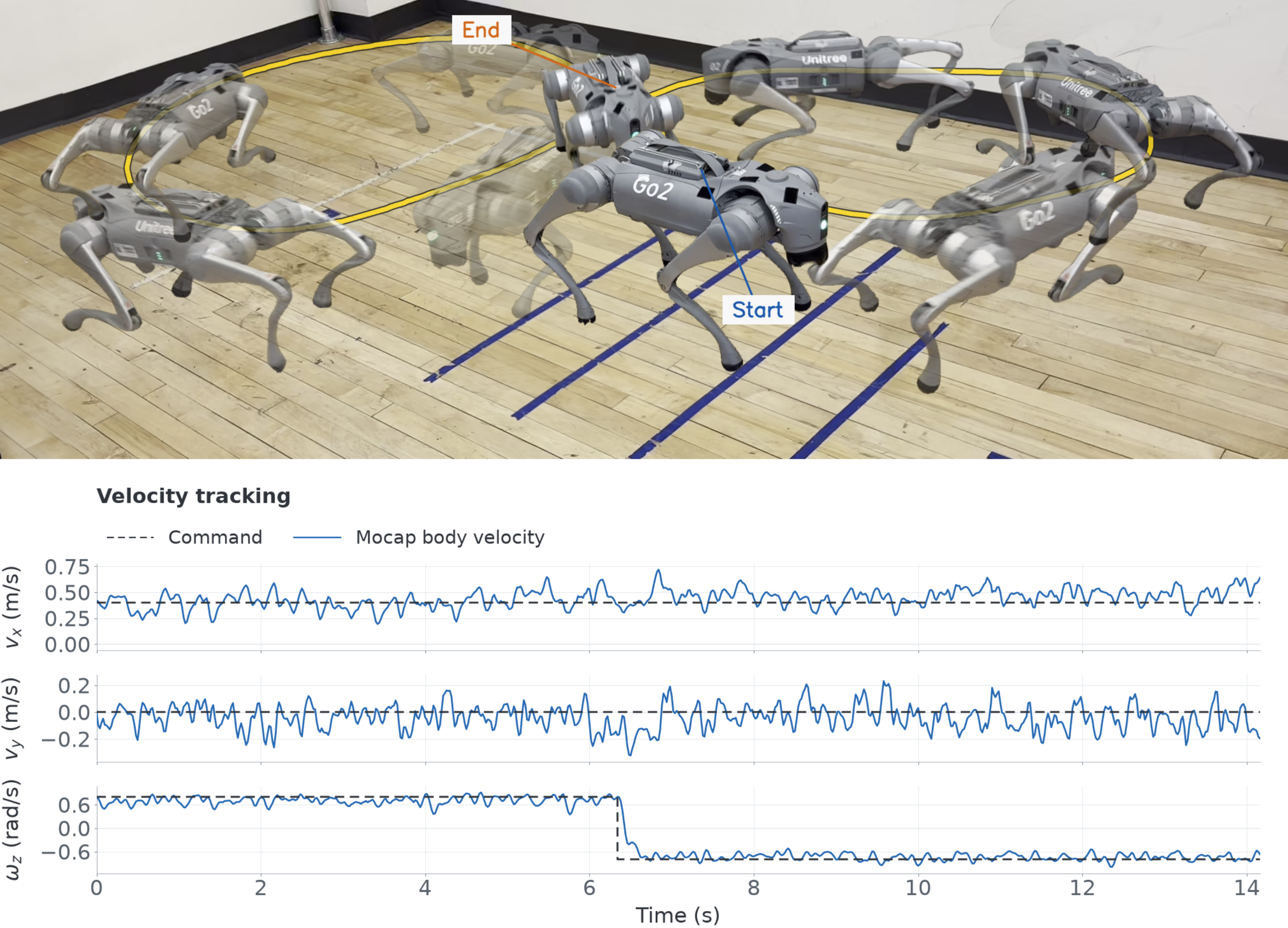}
        \caption{\textbf{Figure-eight walking.}}
        \label{fig:go2_figure8_tracking}
    \end{subfigure}

    \vspace{0.3em}

    \begin{subfigure}[t]{0.96\linewidth}
        \centering
        \includegraphics[width=\linewidth]{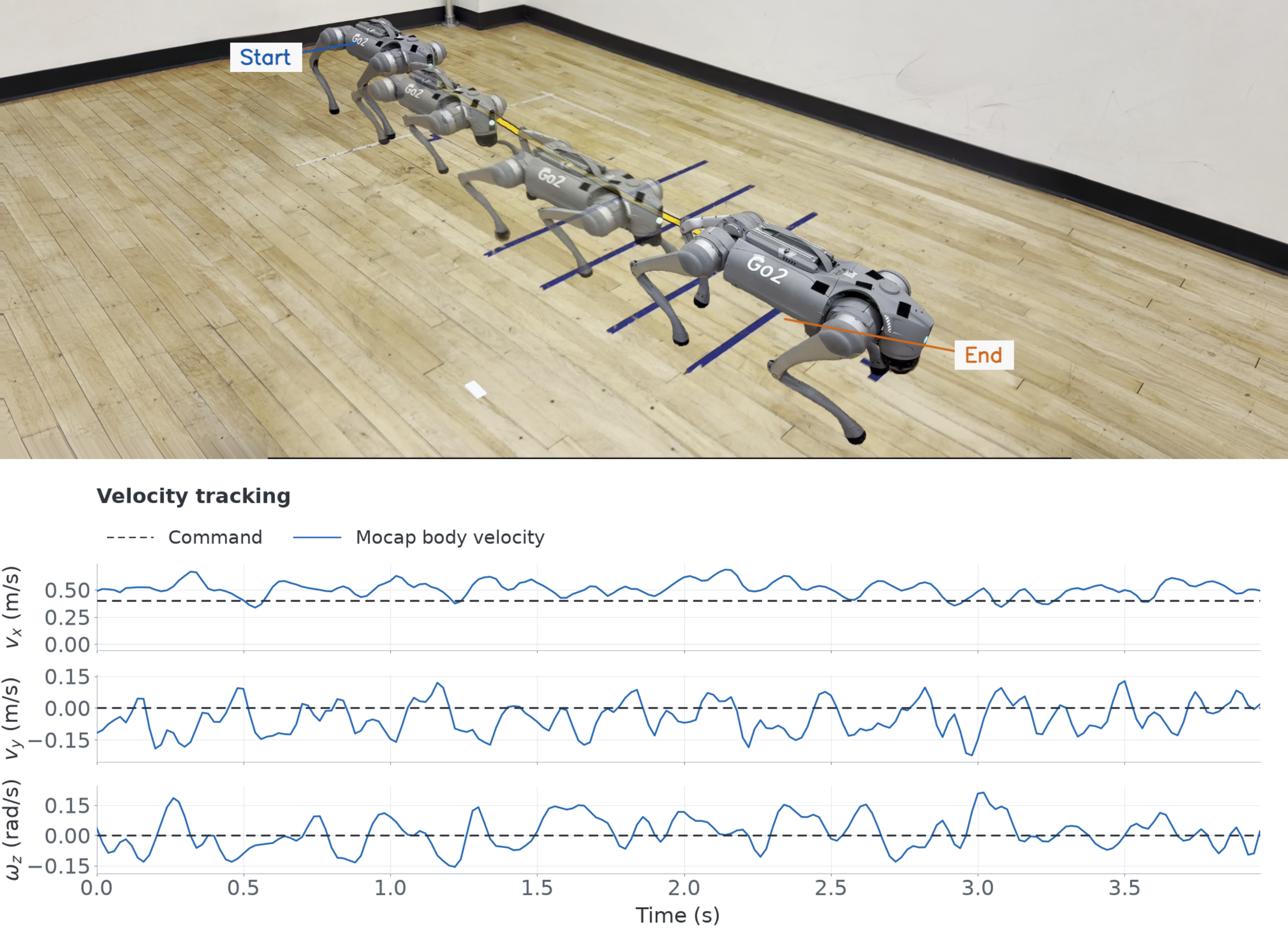}
        \caption{\textbf{Straight-line walking.}}
        \label{fig:go2_straight_tracking}
    \end{subfigure}
    \caption{\textbf{Zero-shot sim-to-real velocity tracking on a physical Unitree Go2.}
    The two panels show figure-eight and straight-line walking after training in simulation.
    In each panel, the montage visualizes the executed motion, while the curves compare velocity commands (dashed) with body-frame mocap measurements (solid).
    Tracking errors are summarized in the text using MAE and RMSE over the plotted post-startup intervals.}
    \label{fig:go2_tracking}
\end{figure}

To quantify tracking during figure-eight and straight-line
walking, we compare velocity commands with body-frame velocities
from motion capture (mocap), as shown in Fig.~\ref{fig:go2_tracking}.
Using aligned measurements before visualization smoothing,
we compute the planar velocity error at each sample as the
Euclidean norm of the forward and lateral velocity errors.
For yaw angular velocity, we use the absolute difference
between measured and commanded values.
We then summarize tracking over each plotted interval using
mean absolute error (MAE) and root mean square error (RMSE),
computed as the mean error magnitude and the square root
of the mean squared magnitude, respectively, with equal
weight per sample.

For figure-eight and straight-line walking, the planar velocity
MAEs are \(0.126\) and \(0.149\,\mathrm{m/s}\), respectively,
with RMSEs of \(0.144\) and \(0.161\,\mathrm{m/s}\).
The corresponding yaw angular velocity MAEs are \(0.112\)
and \(0.068\,\mathrm{rad/s}\), with RMSEs of \(0.177\)
and \(0.084\,\mathrm{rad/s}\).

We further assess locomotion stability by applying external
pushes during motion.
The robot maintains stable locomotion without falling.
With only \(M=3\) MPPI refinement iterations per control step,
the controller operates at an average frequency of
\(308.16\,\mathrm{Hz}\), demonstrating high-frequency control
under a limited online planning budget.
Real-world demonstrations are available at
\url{https://swanfastworldplanner.github.io/}.




\end{document}